\def\cvprPaperID{13815}
\def\confName{CVPR}
\def\confYear{2025}

\def\paperTitle{Global Average Feature Augmentation for Robust Semantic Segmentation with Transformers} 

\def\authorBlock{
Alberto G. Rodriguez Salgado$^{1}$ \quad Maying Shen$^{2}$\quad 
 Philipp Harzig$^{1}$ \quad Peter Mayer $^{1}$  \quad Jose M. Alvarez$^{2}$\\
$^{1}$Volkswagen AG \qquad 
$^{2}$NVIDIA \\
}

\newif\ifreview 
\newif\ifarxiv \newcommand{\arxiv}{\arxivtrue}
\newif\ifcamera 
\newif\ifrebuttal

\arxiv
\pdfoutput=1
\documentclass[10pt,twocolumn,letterpaper]{article}
\ifreview \usepackage[review]{cvpr} \fi
\ifarxiv \usepackage[pagenumbers]{cvpr} \fi
\ifrebuttal \usepackage[rebuttal]{cvpr} \fi
\ifcamera \usepackage{cvpr} \fi

\usepackage{graphicx}
\usepackage{amsmath}
\usepackage{amssymb}
\usepackage{booktabs}

\usepackage{times}
\usepackage{microtype}
\usepackage{epsfig}
\usepackage{pifont}
\usepackage[table,xcdraw]{xcolor}
\usepackage{caption}
\usepackage{float}
\usepackage{placeins}
\usepackage{amssymb}
\usepackage{color, colortbl}
\usepackage{stfloats}
\usepackage{enumitem}
\usepackage{tabularx}
\usepackage{xstring}
\usepackage{multirow}
\usepackage{xspace}
\usepackage{url}
\usepackage{subcaption}
\usepackage{xcolor}
\usepackage{cuted}
\definecolor{Gray}{gray}{0.5}
\usepackage[hang,flushmargin]{footmisc}
\usepackage{algorithm}
\usepackage{algpseudocode}
\ifcamera \usepackage[accsupp]{axessibility} \fi

\usepackage{xcolor}
\usepackage{xparse}
\definecolor{R1color}{rgb}{1, 0, 0}      
\definecolor{R2color}{rgb}{0, 0.5, 0}    
\definecolor{R3color}{rgb}{0, 0, 1}      

% ...

\ifarxiv  \fi

\newcommand{\R}[1]{{%
    \textbf{%
        \ifstrequal{#1}{1}{\textcolor{red}{R#1}}{%
        \ifstrequal{#1}{2}{\textcolor{blue}{R#1}}{%
        \ifstrequal{#1}{3}{\textcolor{magenta}{R#1}}{%
        \ifstrequal{#1}{4}{\textcolor{teal}{R#1}}{%
                           \textcolor{cyan}{R#1}%
        }}}}%
    }%
}}

\usepackage{xr-hyper}

\makeatletter
\newcommand*{\addFileDependency}[1]{
  \typeout{(#1)}
  \@addtofilelist{#1}
  \IfFileExists{#1}{}{\typeout{No file #1.}}
}

\makeatother

\usepackage[pagebackref,breaklinks,colorlinks]{hyperref}
\usepackage[capitalize]{cleveref}
\crefname{section}{Sec.}{Secs.}
\crefname{table}{Table}{Tables}
\crefname{figure}{Fig.}{Figs.}

\begin{document}
\title{\paperTitle}
\author{\authorBlock}
\maketitle
\begin{strip}
  \vspace{-0.7cm}
  \centering
\begin{tabular}{cc}
\includegraphics[width=0.57\linewidth]{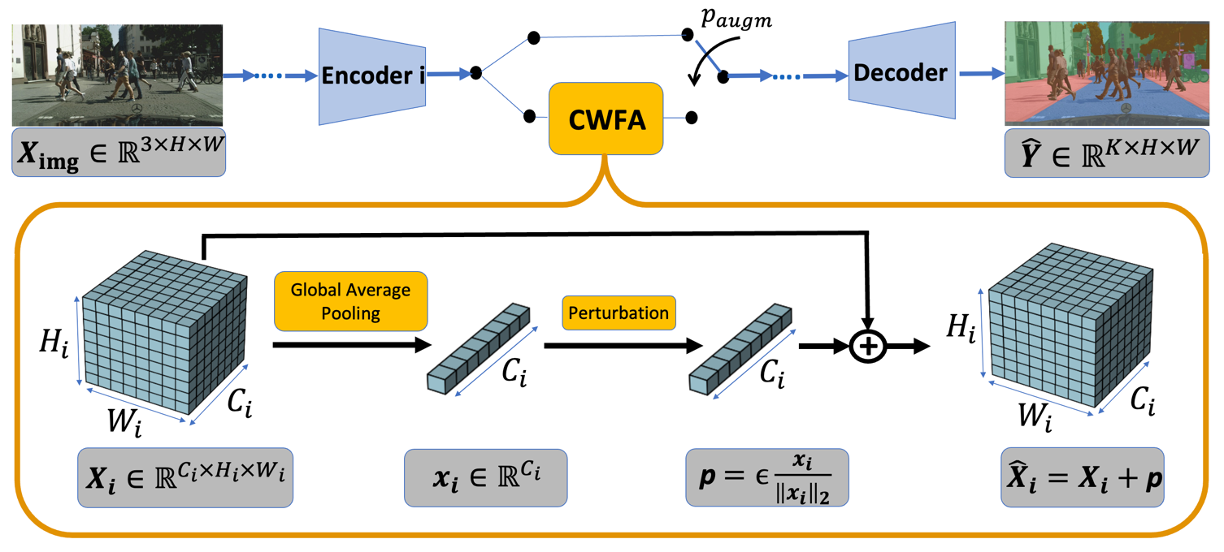}&
\includegraphics[width=0.43\linewidth]{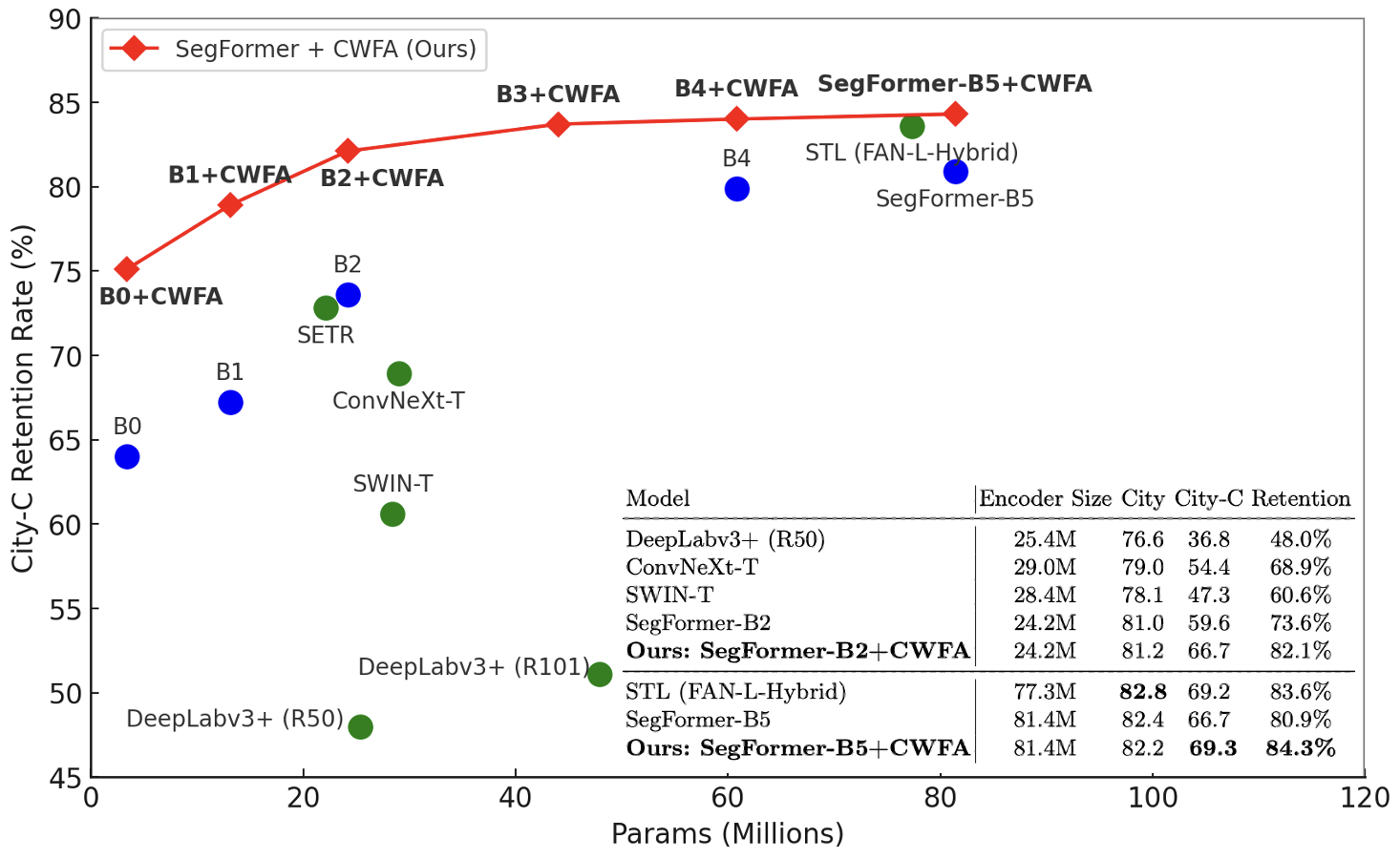}
\end{tabular}
  \vspace{-0.5cm}
   \captionof{figure}{\textbf{Left:} We propose CWFA, a feature augmentation module for Vision Transformers. CWFA computes a feature perturbation based on a global average feature rather than independent perturbations for each feature. \textbf{Right:} Compared to the baseline SegFormer and other CNNs and Transformer models, our approach consistently outperforms them independently of the model size and yields up to 27\% improvements . Our results with large models (SegFormer-B5+CWFA) set a new state-of-the-art retention ratio for semantic segmentation.} 
  \vspace{-5pt}
  \label{fig:teaser}
\end{strip}

\begin{abstract}

Robustness to out-of-distribution data is crucial for deploying modern neural networks. Recently, Vision Transformers, such as SegFormer for semantic segmentation, have shown impressive robustness to visual corruptions like blur or noise affecting the acquisition device. In this paper, we propose Channel Wise Feature Augmentation (CWFA), a simple yet efficient feature augmentation technique to improve the robustness of Vision Transformers for semantic segmentation. CWFA applies a globally estimated perturbation per encoder with minimal compute overhead during training. Extensive evaluations on Cityscapes and ADE20K, with three state-of-the-art Vision Transformer architectures—SegFormer, Swin Transformer, and Twins—demonstrate that CWFA-enhanced models significantly improve robustness without affecting clean data performance. For instance, on Cityscapes, a CWFA-augmented SegFormer-B1 model yields up to $27.7\%$ mIoU robustness gain on impulse noise compared to the non-augmented SegFormer-B1. Furthermore, CWFA-augmented SegFormer-B5 achieves a new state-of-the-art $84.3\%$ retention rate, a $0.7\%$ improvement over the recently published FAN+STL.
\end{abstract}

\section{Introduction}
\label{sec:intro}
Vision Transformers (ViTs) have emerged as predominant state-of-the-art architectures in most visual recognition tasks such as semantic segmentation or object detection. Besides their impressive performance, an essential property of Vision Transformers is their robustness against various corruptions~\cite{zhou2022understanding}. After SegFormer~\cite{xie2021segformer}, the seminal work on robust Vision Transformers for semantic segmentation, other works have analyzed and proposed variations in Vision Transformer architectures to improve their robustness~\cite{zhou2022understanding,fan2,liu2022convnet, liu2021swin, mao2022towards}. Instead of proposing a novel architecture, in this paper, we focus on data augmentation at training time to enhance their robustness to natural corruptions.

Data augmentation has been widely used to improve the overall accuracy and robustness of neural networks. For robustness, most methods focus on image augmentation~\cite{hendrycks2019augmix,wang2021augmax,hendrycks2022pixmix}. While effective, these approaches incur a significant training compute cost increase. For instance, AugMix~\cite{hendrycks2019augmix} requires $35 \%$ more training time. Fewer works have explored robustness through augmentations in the feature space. For instance, Stochastic Feature Augmentation (SFA)~\cite{li2021simple}  perturbs every feature via independent and class-specific Gaussian noise. The latter requires computing a covariance matrix for every class, making it impractical for large datasets. Further, SFA computes the loss for both original and perturbed features, further impacting efficiency. 

This paper introduces Channel-Wise Feature Augmentation (CWFA), a feature augmentation method for Vision Transformers that significantly improves robustness without sacrificing efficiency. Intuitively, image corruptions such as noise should affect all features within a channel block regardless of spatial location. CWFA perturbs channel-wise features, thus making the perturbation independent of the spatial feature dimension. Specifically, our method does not compute a random perturbation per feature nor requires covariance matrices, which is costly and inefficient. Instead, we first obtain a representative feature by averaging all features and normalizing it to generate the perturbation. This perturbation is then applied to every feature in the block. Our perturbation can be computed efficiently using a global average pooling operation and therefore incurs minimum training compute overhead.

We demonstrate the efficacy of CWFA on two datasets, Cityscapes and ADE20K, and three different Transformer architectures, SegFormer, Swin Transformer~\cite{liu2021swin} and Twins~\cite{chu2021twins}. On Cityscapes, CWFA improves the robustness of SegFormer-B1 by $27.7\%$ and $21.0\%$ for impulse noise and snow corruptions, respectively. Compared to SFA, our approach increases the retention rate of SegFormer-B1 by $9.7\%$. The largest SegFormer-B5 model trained using our proposed CWFA method establishes a new state-of-the-art in robustness on the Cityscapes-C benchmark, surpassing the recently published FAN-STL by $0.7\%$. As experiments will show, our results also generalize over other architectures and datasets. Importantly, our approach does not increase inference costs and only slightly increases the training costs by $2\%$.

In summary, our main contributions are:
\begin{enumerate}[nolistsep]
    \item We propose CWFA, a novel, simple and effective feature augmentation method to robustify Vision Transformers for semantic segmentation based on a channel-wise global average feature perturbation.
    \item Our method is a plug-in module that can be used with different architectures independently of the size of the model.
    \item Our method generalizes over datasets and different Transformers architectures, setting a new state of the art for robustness in semantic segmentation.
\end{enumerate} 

\section{Related Work}
\label{sec:related}
\label{sec:related}
\textbf{Robustness of Vision Transformers} Transformer architectures such as SETR \cite{zheng2021rethinking}, Swin Transformer, SegFormer, PVT \cite{wang2021pyramid}, and Twins have been extensively explored for segmentation tasks. Among these, SegFormer has demonstrated the robustness advantages of Vision Transformers over CNN-based methods like DeepLabv3+ \cite{chen2018encoder}, yet still faced performance drops on corrupted images. Recent works such as FAN \cite{zhou2022understanding} and FAN+STL \cite{fan2} further underscore the superiority of Transformer-based approaches over traditional CNNs in terms of robustness and generalization.

\textbf{Data Augmentation in Image Space} A variety of augmentation techniques have been developed to enhance model robustness. Methods like Cutout \cite{devries2017improved}, Mixup \cite{zhang2017mixup}, and CutMix \cite{yun2019cutmix} apply intense augmentations but with inconsistent robustness improvements \cite{hendrycks2019augmix}. Advanced techniques, such as AugMix , AugMax \cite{wang2021augmax}, and PixMix \cite{hendrycks2022pixmix}, further refine these approaches, though with challenges like computational overhead. Notably, \cite{mintun2021interaction} revealed that some augmentations might not generalize well across diverse corruptions. Other strategies involve noise-based augmentations \cite{erichson2022noisymix, rusak2020simple} or diverse transformations, like in PRIME \cite{modas2022prime}. A comprehensive review of image augmentation can be found in \cite{dataaugmentationsurvey}.

\textbf{Data Augmentation in Feature Space} Deep features typically show linear characteristics \cite{upchurch2017deep}, making them suitable for manipulation through vector interpolation \cite{gardner2015deep}. Notable methods like Noisy Feature Mixup \cite{lim2021noisy} and Stability Training \cite{zheng2016improving} have emerged in this domain. The Frequency Preference Control Module \cite{bu2023towards} adapts the frequency components of features to improve the robustness against adversarial attacks. Stochastic Feature Augmentation (SFA), the most similar technique to CWFA, focuses on domain generalization via noise in the feature space as a feature perturbation. SFA introduces an added loss term and needs to compute full covariance matrices for every class, a demanding task for datasets like ADE20K \cite{zhou2017scene} with up to 150 classes. There are two key differences between SFA and CWFA. First, SFA uses Gaussian noise as the perturbation while CWFA uses a normalized version of the feature itself. Second, SFA perturbs every feature of a feature map ($H \times W \times C$) with independent Gaussian noise, while in contrast CWFA applies a uniform perturbation across each channel dimension ($H \times W$) of the feature map. We emphasize that our experiments \ref{sec:experiments} show stronger robustness gains when using CWFA. 


\section{Method}
\label{sec:method}
In this section, we introduce CWFA, our proposal to improve robustness of Vision Transformers segmentation models. 
\subsection{Notation}
We refer to an encoder-decoder segmentation model as $f$, divided into n stacked encoders  $\textrm{enc}_i$ with $i \in [1...n]$ and a decoder head $h$. The model takes as an input images $\boldsymbol{X}_{\text{img}} \in \mathbb{R}^{3 \times H \times W}$. We also refer as $\Tilde{\mathcal{X}}_{\text{img}}$ to a set of corrupted images. Given an input image, each encoder $\textrm{enc}_i$ computes a feature space representation $ \boldsymbol{X_i} \in \mathbb{R}^{C_i \times H_i \times W_i} $ from the feature space $\mathcal{X}_i$. We note that the encoders correspond to module blocks at each spatial resolution in the backbone. The decoder head $h$ takes encoder features as input and combines them to a final segmentation prediction $ \boldsymbol{\hat{Y}} \in \mathbb{R}^{H \times W \times K}$ where $K$ refers to the number of semantic classes. We define the set of ground truths as $\mathcal{Y}$.

During training, the model $f$ has only access to training set $\{ \mathcal{X}_{\text{train}}  , \mathcal{Y}_{\text{train}} \}$ which only includes clean data. In the validation setting, we run inference on the clean and corrupted validation sets $\{ \mathcal{X}_{\text{val}}, \mathcal{Y}_{\text{val}} \}$ and $\{ \Tilde{\mathcal{X}}_{\text{val}} , \mathcal{Y}_{\text{val}}\}$. Ideally, a robust model should have a similar performance on both validation sets. 

\subsection{Channel-wise Feature Augmentation}
\begin{figure*}[!h]
    \centering
    \begin{minipage}[b]{0.47\textwidth}
        \begin{algorithm}[H]
            \caption{Global Average Feature Augmentation}
            \label{alg:cwfa}
            \begin{algorithmic}[1] 
                \State \textbf{Input} $\epsilon$, $p_{\text{augm}}$, $\boldsymbol{X_i}$
                \State \textbf{Output} $\boldsymbol{\hat{X_i}}$
                \State $\Tilde{p} \gets \text{Bernoulli}(p_{\text{augm}})$
                \If{$\Tilde{p} = 0$}
                    \State $\boldsymbol{\hat{X_i}} \gets \boldsymbol{X_i}$
                \Else
                    \State $\boldsymbol{x_i} \gets \text{GlobalAvgPool}(\boldsymbol{X_i})$
                    \State $\boldsymbol{p} \gets \epsilon \cdot \frac{\boldsymbol{x_i}}{\|\boldsymbol{x_i}\|_2}$
                    \State $\boldsymbol{\hat{X_i}} \gets \boldsymbol{X_i} + \boldsymbol{p}$
                \EndIf
            \end{algorithmic}
        \end{algorithm}
    \end{minipage}
    \hfill
\begin{minipage}[b]{0.45\textwidth}
        \centering
        \begin{subfigure}[b]{0.49\columnwidth}
            \includegraphics[width=\textwidth]{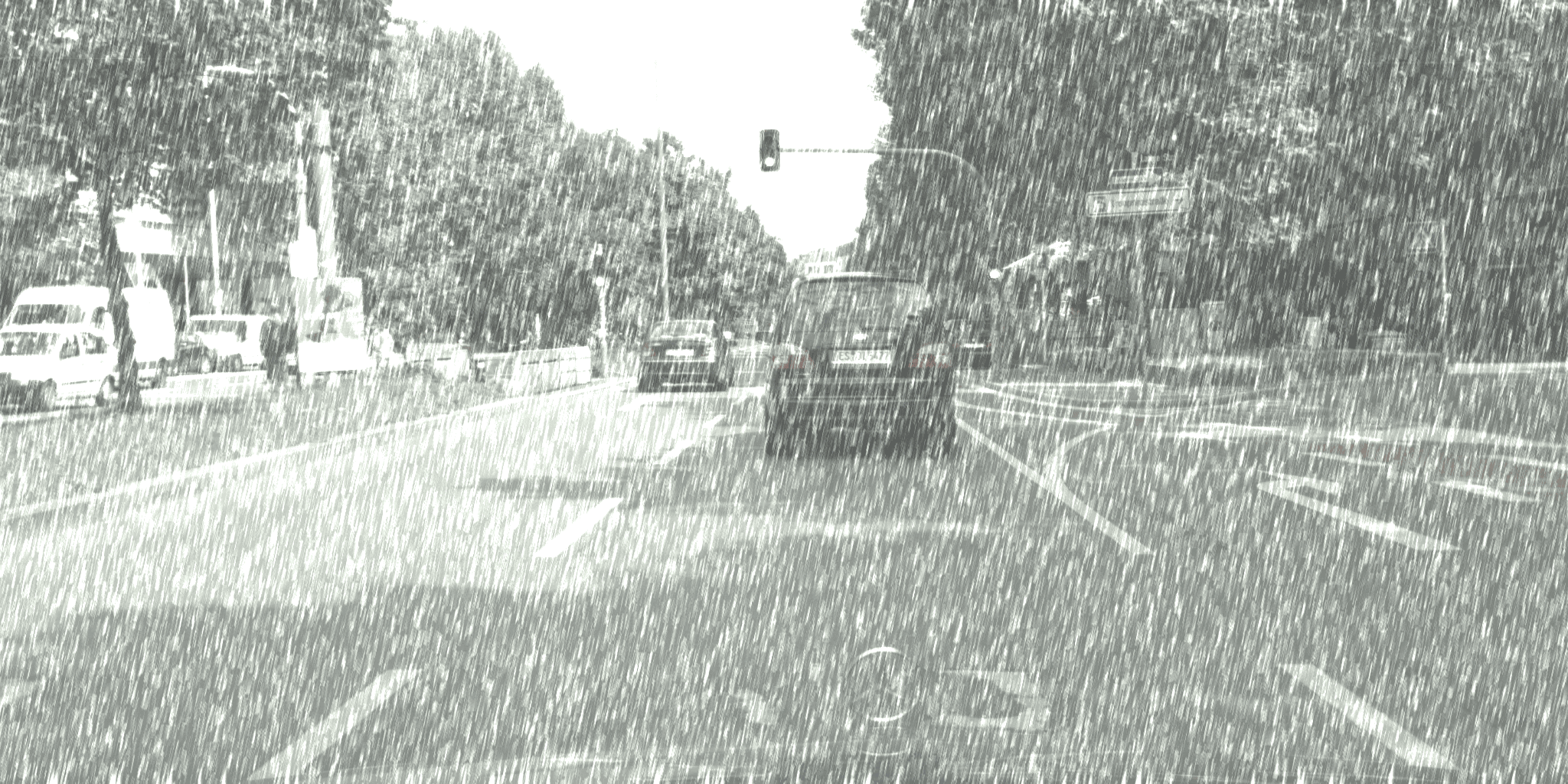}            
        \end{subfigure}    \\
        \begin{subfigure}[b]{0.49\columnwidth}
            \includegraphics[width=\textwidth]{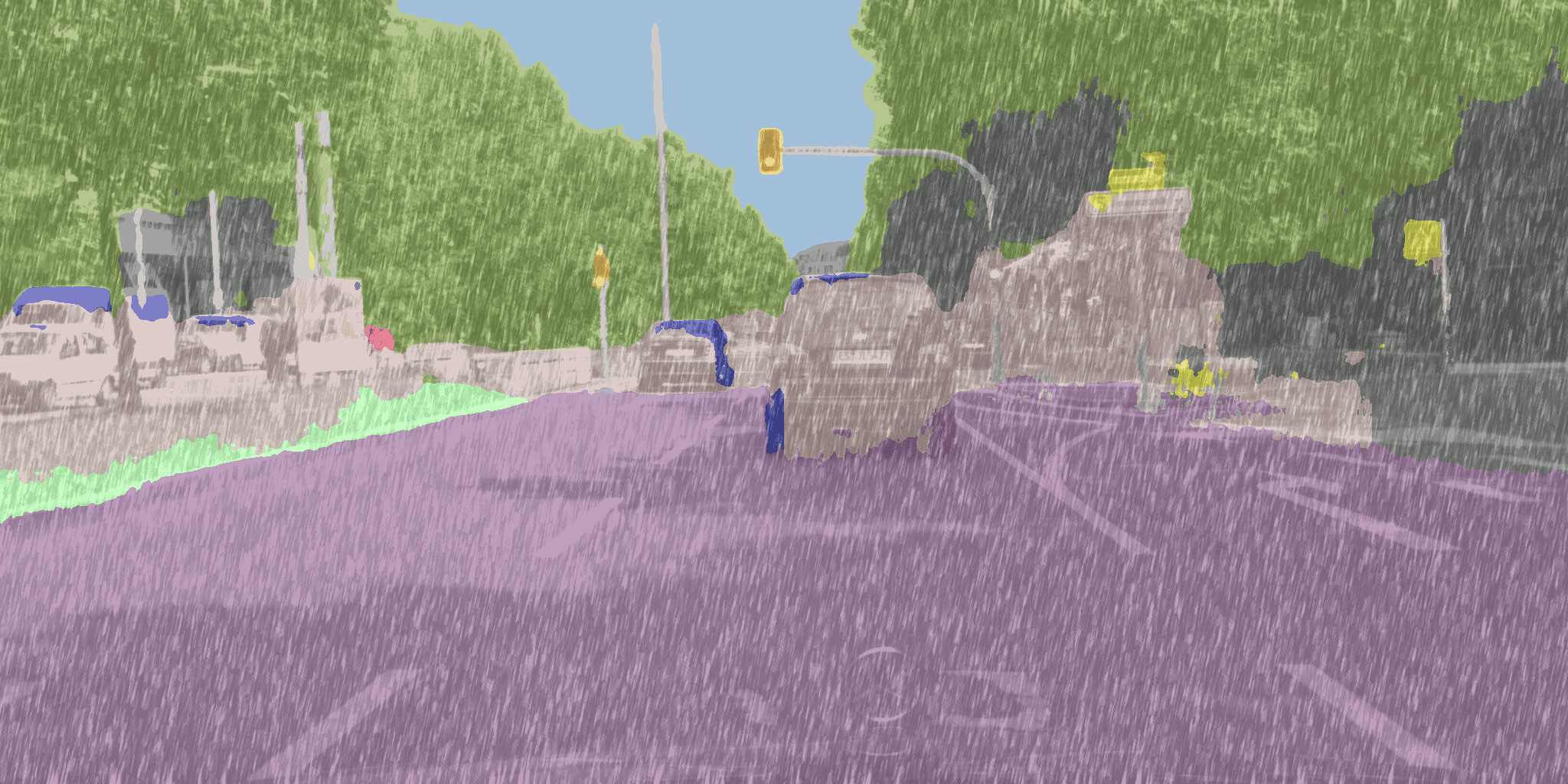}            
        \end{subfigure} 
        \begin{subfigure}[b]{0.49\columnwidth}
            \includegraphics[width=\textwidth]{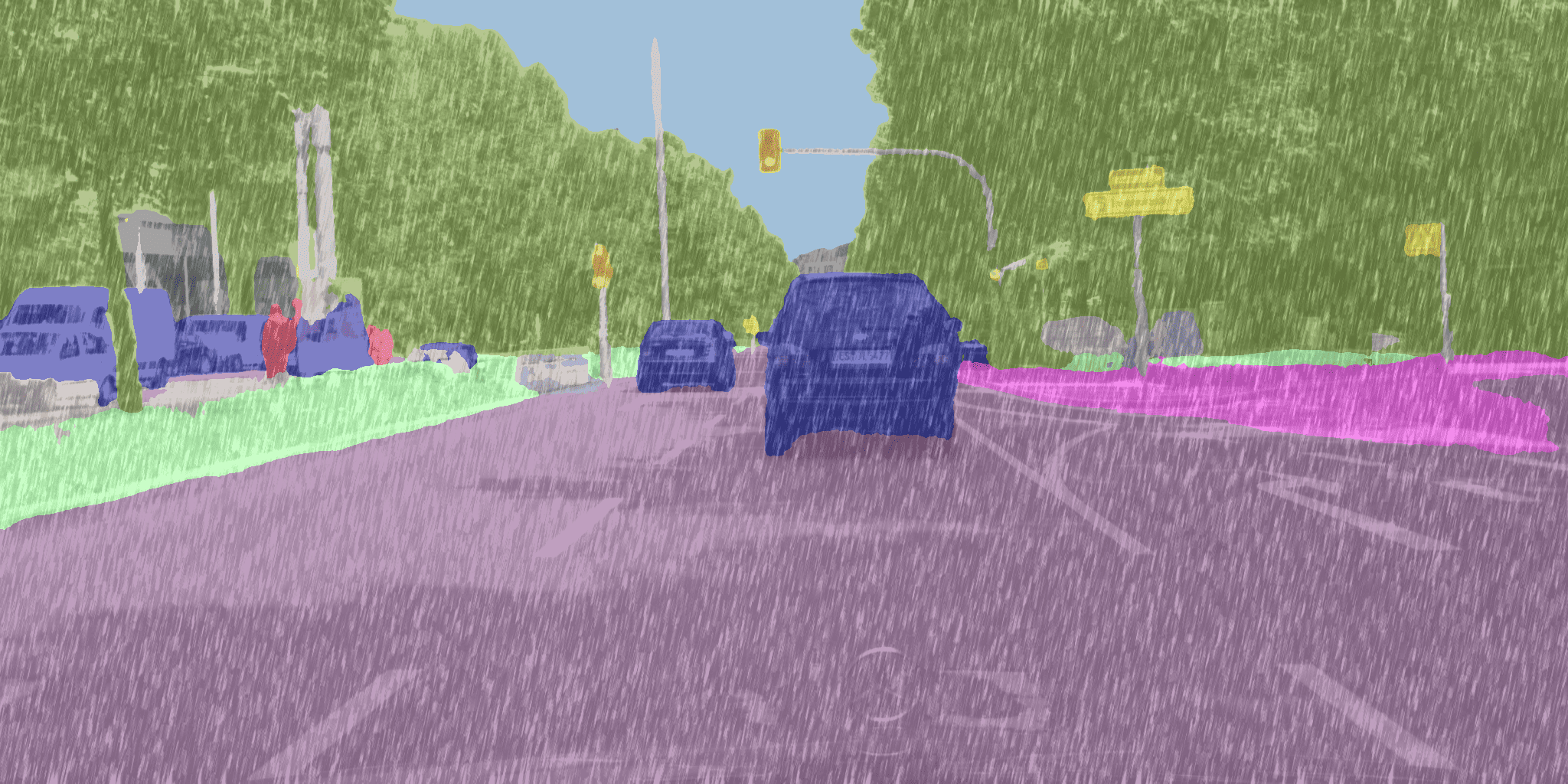}            
        \end{subfigure} 
        \caption{ {Qualitative results of CWFA on SegFormer-B5.} Top: Input image corrupted with snow severity 5; Bottom-left: Baseline. off-the shelf SegFormer-B5; Bottom-right: \textbf{Ours.} SegFormer-B5 trained using CWFA. }
        \label{fig:visual comparison cwfa B5}
    \end{minipage}        
    
\end{figure*}

Figure \ref{fig:teaser} illustrates the overall pipeline and the pseudo code of the algorithm is listed in Algorithm \ref{alg:cwfa}. Given the feature representation \(\boldsymbol{X}_i \in \mathbb{R}^{C_i \times H_i \times W_i}\) of an image \(\boldsymbol{X}_{img}\) after the i-th encoder \(\textrm{enc}_i\), CWFA augments the feature representation in a channel-wise manner as:

\begin{equation}\label{cwfa: equation 1}
    \boldsymbol{\hat{X}}_{i,(c,j,m)} =  \boldsymbol{X}_{i,(c,j,m)}  + \boldsymbol{p}_c
\end{equation}

where \(\boldsymbol{p} \in \mathbb{R}^{C_i}\) is the channel-wise feature perturbation vector, \(\boldsymbol{\hat{X}}_i\) is the augmented feature, \(c\) refers to the channel index, and \(j\) and \(m\) refer to the spatial dimensions (height and width respectively) of the feature map.

To compute the channel-wise perturbation \(\boldsymbol{p}\), we first obtain the global channel average feature \(\boldsymbol{x}_i \in \mathbb{R}^{C_i}\) of the clean feature map as:

\begin{align}
    \boldsymbol{x}_{i,c} = \frac{1}{H_i W_i} \sum_{j=1}^{H_i} \sum_{m=1}^{W_i} \boldsymbol{X}_{i,(c,j,m)}
\end{align}

Here, \(\boldsymbol{x}_{i,c}\) represents the average value of the \(c\)-th channel over all spatial positions \((j,m)\).

Next, we compute the feature perturbation vector \(\boldsymbol{p}\) as:

\begin{align}\label{cwfa: equation for perturbation p}
    \boldsymbol{p} = \epsilon \dfrac{\boldsymbol{x}_i}{\lvert \lvert \boldsymbol{x}_i \rvert \rvert_2}
\end{align}

where \(\epsilon\) is the perturbation strength. The larger the value of \(\epsilon\), the more significant the perturbation applied to the features.

In practice, the global average pooling can be efficiently performed using a global average pooling operation over the spatial dimensions. During training, CWFA is applied with a probability \(p_{\text{augm}}\), similar to common augmentation techniques.

The motivation for using Vision Transformers lies in their global attention mechanism, which enables the computation of global perturbations through the average pooling of the feature space. Unlike traditional CNNs, which are limited by local receptive fields, Vision Transformers leverage global attention to create features that encapsulate information from the entire image. This ability to efficiently capture and aggregate global information is crucial for identifying the common perturbation, as shown in equation \ref{cwfa: equation for perturbation p}.

\section{Experiments}
\label{sec:experiments}
\textbf{Datasets:} In our experiments, we train our models on two datasets Cityscapes \cite{cordts2016cityscapes} and ADE20K ~\cite{zhou2017scene}. The Cityscapes dataset consists of 2975 training and 500 validation images with pixel annotations of 19 categories. The ADE20K dataset is a scene parsing dataset consisting of 20210 training and 2000 validation images with pixel annotations of 150 categories. ADE20K is one of the most challenging segmentation benchmarks, where state-of-the-art models such as InternImage with 1300 M parameters achieve 62.9 mIoU ~\cite{wang2023internimage}. For evaluation, we follow other works suggesting that synthetic corruptions are a good indicator of the model's robustness against real-world natural corruptions without requiring the labeling of new data~\cite{hendrycks2021many, michaelis2019benchmarking}. We report the performance of models for the (original) validation sets and for Cityscapes-C (City-C), Cityscapes-$\Bar{\text{C}}$ (City-$\Bar{\text{C}}$), and ADE20K-C (ADE-C). The $C$ refers to a synthetic version of the validation set corrupted with 16 natural corruptions (each with 5 levels of severity, 5 being the most severe one), subdivided into the blur, noise, weather and digital~\cite{hendrycks2018benchmarking}. Cityscapes-$\Bar{\text{C}}$ is the validation set of Cityscapes corrupted with additional set of corruptions dissimilar from Cityscapes-C~\cite{mintun2021interaction}. The motivation for incorporating this version is to further evaluate the zero-shot robustness of the methods.  

We report standard evaluation metrics: the mIoU score for each dataset. For the corrupted versions, we report the averaged mIoU over severity levels 1 to 5 for each corruption type except for noise, where we average over the first 3 levels as this is standard practice on Cityscapes-C \cite{xie2021segformer,kamann2020benchmarking}. In addition, we report the average mIoU over all corruptions to summarize the performance on the corrupted datasets. We also report the retention rate (Retention $R$) to reflect the ability of a model to maintain its performance despite the corruption. This ratio is defined as  $R= \frac{\text{Robust mIoU}}{\text{Clean mIoU}} = \frac{\text{City-C}}{\text{City}}$. 

We follow the evaluation settings used in \cite{xie2021segformer}. We evaluate on the original resolution $2048\times1024$ Cityscapes images using a sliding window inference with window size $1024\times1024$. On ADE20K, we evaluate on $512\times512$ images without a sliding window. 
 
\textbf{Models:} We test our proposal on three Vision Transformer models for semantic segmentation: SegFormer, Twins, and Swin Transformer. We use SegFormer for a comprehensive evaluation on all its variants (SegFormer-B0 to B5) as it is one of the most robust models in the literature. For Twins, we use the Twins-PCPVT and SVT backbones with corresponding sizes S and for Swin Transformer, we use the Tiny version. Therefore, we evaluate on a larger variety of model sizes with emphasis on compact models, where the robustness tends to drop significantly. 

\textbf{Implementation details:} For our main experiments, we train for 160k iterations all SegFormer versions with CWFA using the same training regime as those used to train the original models from scratch for fair comparison. We use standard image augmentations such as random crop, random flip as well as random resizing with ratio $0.5$ to $2.0$, plus CWFA when applicable. For CWFA, we set the probability to $p_{\text{augm}}=0.3$, $\epsilon=9$ for the smaller models (SegFormer-B0 to B2) and $\epsilon=15$ for larger models (SegFormer-B3 to B5). Similarly, we set $\epsilon=9$ for both Twins PCPVT and SVT models, and $\epsilon=15$ for Swin-T. We provide further details on this choice as well as the selection of $\epsilon$ in our ablation studies in Section \ref{sec:ablation studies}, and the remaining implementation details in the supplementary material. Unless otherwise specified, we apply CWFA to every encoder in the architecture. That is, four times for SegFormer, Twins and Swin Transformers.

\begin{table*}[!t]
\caption{\textbf{SegFormer on Cityscapes.} Our results compared to the SegFormer baseline. Our approach yields impressive improvements for most corruptions independently of the model size. We report the mIoU score}
\label{table: segformer on cityscapes-c 160k}
\scalebox{0.6}{
\begin{tabular}{c|c|c|c|c|c|c|c|c|c|c|c|c|c|c|c|c|c|c|c|c|} 
   \multicolumn{1}{c|}{Model /}& \multirow{2}{*}{Method} & City &\multicolumn{4}{|c|}{Blur}  & \multicolumn{4}{|c|}{Noise} & \multicolumn{4}{|c|}{Digital} & \multicolumn{4}{|c|}{Weather} & \multirow{2}{*}{Average} & \multirow{2}{*}{Reten.}  \\
  \cline{4-19}
    \multicolumn{1}{c|}{Params. [M.]} & &(Clean) & Motion & Defoc & Glass & Gauss & Gauss & Impul & Shot & Speck & Bright & Contr & Satur & JPEG & Snow & Spatt & Fog & Frost & & \\
     \hline
    \multirow{2}{*}{B0 / 3.4}  & Segformer & 76.3 & 59.8 & 59.1 & 51.0 & 59.2 & 25.3 & 27.0 & 30.7 & 51.5 & \textbf{73.4} & 66.5 & \textbf{72.0} & 38.3 & 22.4 & 53.5 & \textbf{65.6} & 31.6 & 49.2 & 64.5 \\
   
 & +CWFA (Ours) & \textbf{76.4} & \textbf{64.3} & \textbf{62.8} & \textbf{59.0} & \textbf{63.3} & \textbf{43.1} & \textbf{43.9} & \textbf{49.9} & \textbf{62.4} & 72.9 & \textbf{66.6} & 71.7 & \textbf{47.3} & \textbf{41.6} & \textbf{63.0} & 64.5 & \textbf{42.6} & \textbf{57.4} & \textbf{75.1}\\
 \hline
    \multirow{2}{*}{B1 / 13.1}  & Segformer & \textbf{78.5} & 63.9 & 63.7 & 52.2 & 63.7 & 30.2 & 23.7 & 35.8 & 56.3 & \textbf{76.4} & \textbf{71.0} & \textbf{74.9} & 36.7 & 28.3 & 60.4 & \textbf{70.8} & 36.7 & 52.8 & 67.2 \\

& +CWFA (Ours) & 78.2 & \textbf{66.9} & \textbf{65.4} & \textbf{60.9} & \textbf{65.7} & \textbf{50.1} & \textbf{51.4} & \textbf{56.0} & \textbf{67.1} & 76.1 & 70.6 & 74.8 & \textbf{47.0} & \textbf{49.3} & \textbf{68.2} & 70.0 & \textbf{47.2} & \textbf{61.7} & \textbf{78.9}\\
  \hline
    \multirow{2}{*}{B2 / 24.2}  & Segformer & 81.0 & 67.7 & 67.8 & 58.6 & 68.3 & 35.8 & 35.6 & 41.4 & 60.3 & 79.7 & 75.6 & 78.4 & 46.0 & 35.0 & 65.4 & 75.7 & 41.3 & 58.3 & 72.0 \\

&  +CWFA (Ours)  &\textbf{81.2} & \textbf{70.5} & \textbf{69.7} & \textbf{66.3} & \textbf{70.4} & \textbf{57.7} & \textbf{58.0} & \textbf{63.1} & \textbf{71.9} & \textbf{79.8} & \textbf{76.1} & 78.4 & \textbf{50.4} & \textbf{55.3} & \textbf{72.1} & 75.7 & \textbf{51.1} & \textbf{66.7} & \textbf{82.1}  \\
\hline
  \multirow{2}{*}{B3 / 44.0}  & Segformer  & 80.9 & 67.6 & 66.9 & 59.9 & 67.9 & 49.6 & 43.6 & 54.0 & 65.8 & 79.7 & 75.5 & 78.6 & 49.4 & 32.3 & 63.5 & 75.6 & 42.1 & 60.7 & 75.1  \\ 

&  +CWFA (Ours) & \textbf{81.5} & \textbf{70.9} & \textbf{70.5} & \textbf{66.9} & \textbf{71.3} & \textbf{62.5} & \textbf{63.0} & \textbf{67.1} & \textbf{74.0} & \textbf{80.2} & \textbf{77.0} & \textbf{79.1} & \textbf{52.2} & \textbf{54.5} & \textbf{73.6} & \textbf{76.9} & \textbf{51.7} & \textbf{68.2} & \textbf{83.7} \\
\hline
  \multirow{2}{*}{B4 / 60.8}  & Segformer & \textbf{82.6} & 68.7 & 68.6 & 62.4 & 69.4 & 57.9 & 57.1 & 62.5 & 71.3 & \textbf{81.3} & \textbf{77.6} &\textbf{80.5} & \textbf{58.5} & 36.5 & 68.0 & \textbf{78.2} & 48.5 & 65.4 & 79.2\\

& +CWFA (Ours) &  82.1 & \textbf{71.0} & \textbf{70.8} & \textbf{68.0} & \textbf{71.7} & \textbf{61.8} & \textbf{64.0} & \textbf{67.3} & \textbf{74.7} & 80.9 & 77.5 & 79.7 & 53.9 & \textbf{58.2} & \textbf{72.7} & 77.9 & \textbf{54.5} & \textbf{69.0} & \textbf{84.0} \\
\hline
  \multirow{2}{*}{B5 / 81.4}  & Segformer & \textbf{82.4} & \textbf{70.4} & 69.4 & 64.8 & 70.3 & 60.2 & 55.2 & 65.4 & 73.5 & \textbf{81.3} & \textbf{77.7} & \textbf{80.4} & \textbf{58.9} & 43.1 & 68.9 & \textbf{78.6} & 48.7 & 66.7 & 80.9 \\

&  +CWFA (Ours)  & 82.2 & 70.1 & \textbf{69.9} & \textbf{66.5} & \textbf{71.2} & \textbf{64.5} & \textbf{66.6} & \textbf{68.6} & \textbf{74.7} & 80.8 & 77.1 & 79.9 & 54.7 & \textbf{58.2} & \textbf{73.6} & 78.1 & \textbf{53.6} & \textbf{69.3} & \textbf{84.3} \\
\end{tabular}}
\end{table*}

\subsection{Results}\label{sec: results}
\textbf{SegFormer on Cityscapes.} In our first experiment, we focus on analyzing the benefits of incorporating our approach to training a top-performing model such as SegFormer on Cityscapes. Table~\ref{table: segformer on cityscapes-c 160k} summarizes our results for all the variants. As we can see, our approach clearly outperforms the robustness of the baseline (SegFormer) independently of the model size and independently of the corruption type, without sacrificing the original mIoU performance. We observe the largest gains for the small and mid-size models SegFormer-B0 to B3. The largest gain is for noise-related corruptions. For example, on impulse noise CWFA improves the performance by $16.9\%$, $27.7\%$ and $22,4\%$ mIoU for SegFormer-B0 to B2, respectively. As expected, the gain decrements for the largest models. Nevertheless, for SegFormer-B5, we still observe a significant $\sim3.5\%$ boost in the retention rate without affecting the overall clean mIoU.  Figure \ref{fig:visual comparison cwfa B5} shows improved results with the SegFormer-B5 model using CWFA compared to the baseline. More visual comparisons can be found in the supplementary materials.

To further verify the zero-shot robustness of our method, in Table~\ref{table: segformer on city-c and city-c-bar} we also report results for Cityscapes-$\bar{\text{C}}$. This test set consists of perturbations that are completely different to those in Cityscapes-C and generated from different sources. This is clearly reflected in the average mIoU of both models (with and without CFWA). As we can see, the performance is much lower in this new evaluation set. Nevertheless, by using CWFA we are also able to boost the robustness of the original models by a large margin. For instance, for compact models, there is a $\sim8\%$ retention improvement and, for larger models such as SegFormer-B5, the retention rate on  Cityscapes-$\bar{\text{C}}$ is improved by $6.5\%$, which is significantly larger than the improvement on SegFormer-B5 for Cityscapes-C. This is likely due to the lower overall baseline performance of the models on the corrupted data from Cityscapes-$\bar{\text{C}}$. 

\begin{figure*}[t]
    \centering
    \begin{subfigure}[b]{0.32\textwidth}
        \centering
        \includegraphics[width=\textwidth]{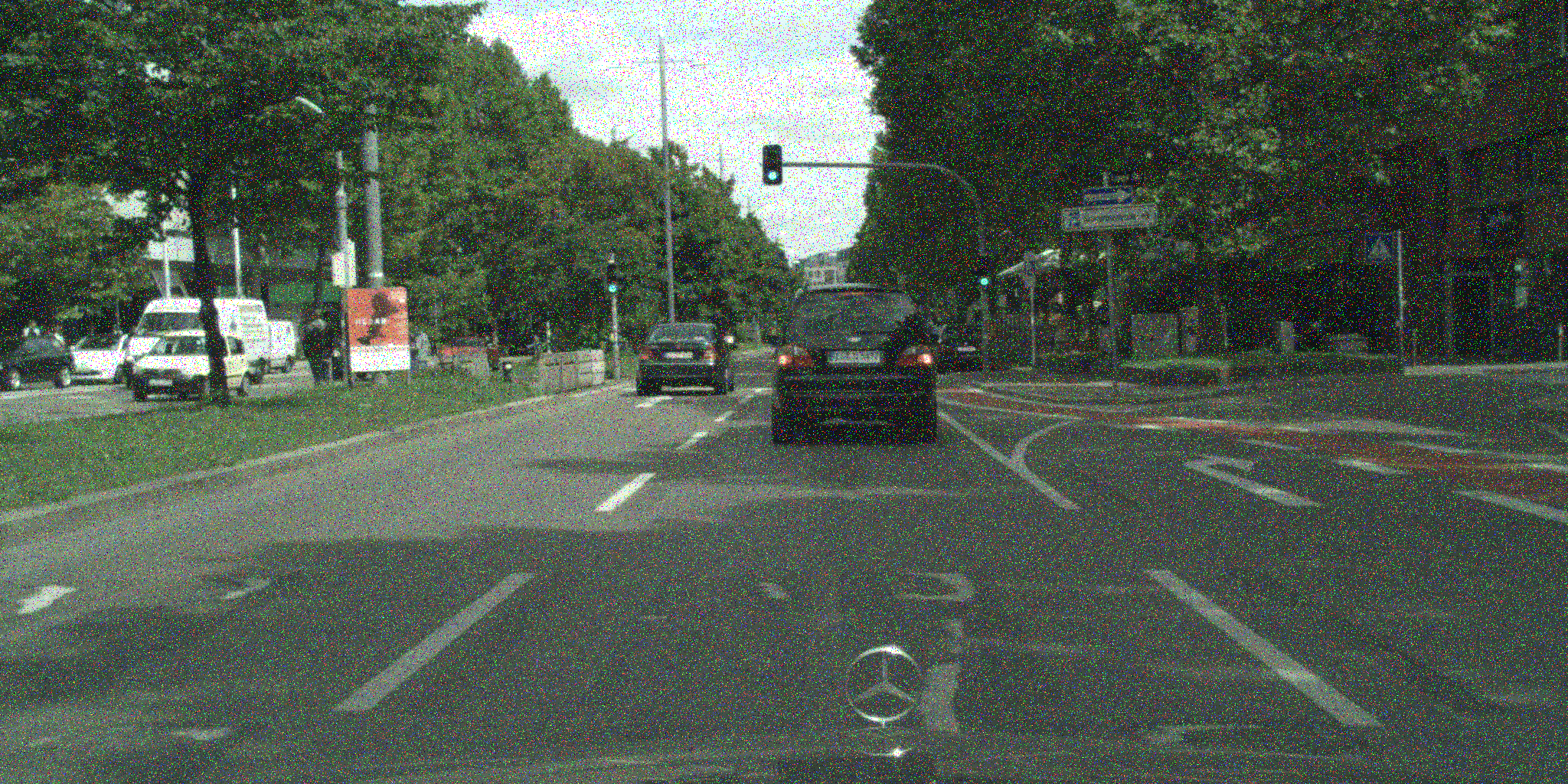}
        \caption{Impulse noise sev.3}
        \label{fig:sub1}
    \end{subfigure}    
    \begin{subfigure}[b]{0.32\textwidth}
        \centering
        \includegraphics[width=\textwidth]{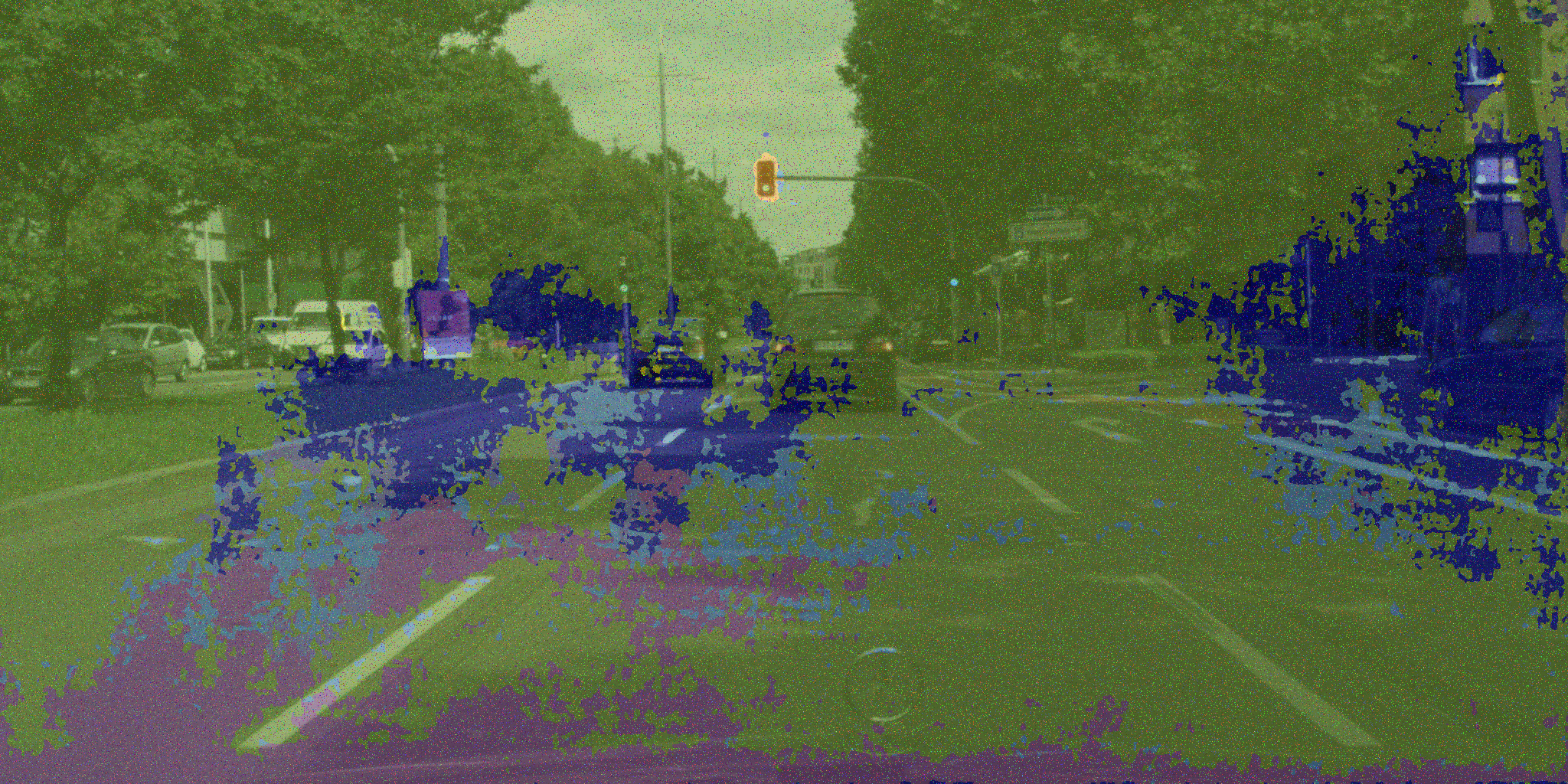}
        \caption{DeepLabV3+}
        \label{fig:sub2}
    \end{subfigure}    
    \begin{subfigure}[b]{0.32\textwidth}
        \centering
        \includegraphics[width=\textwidth]{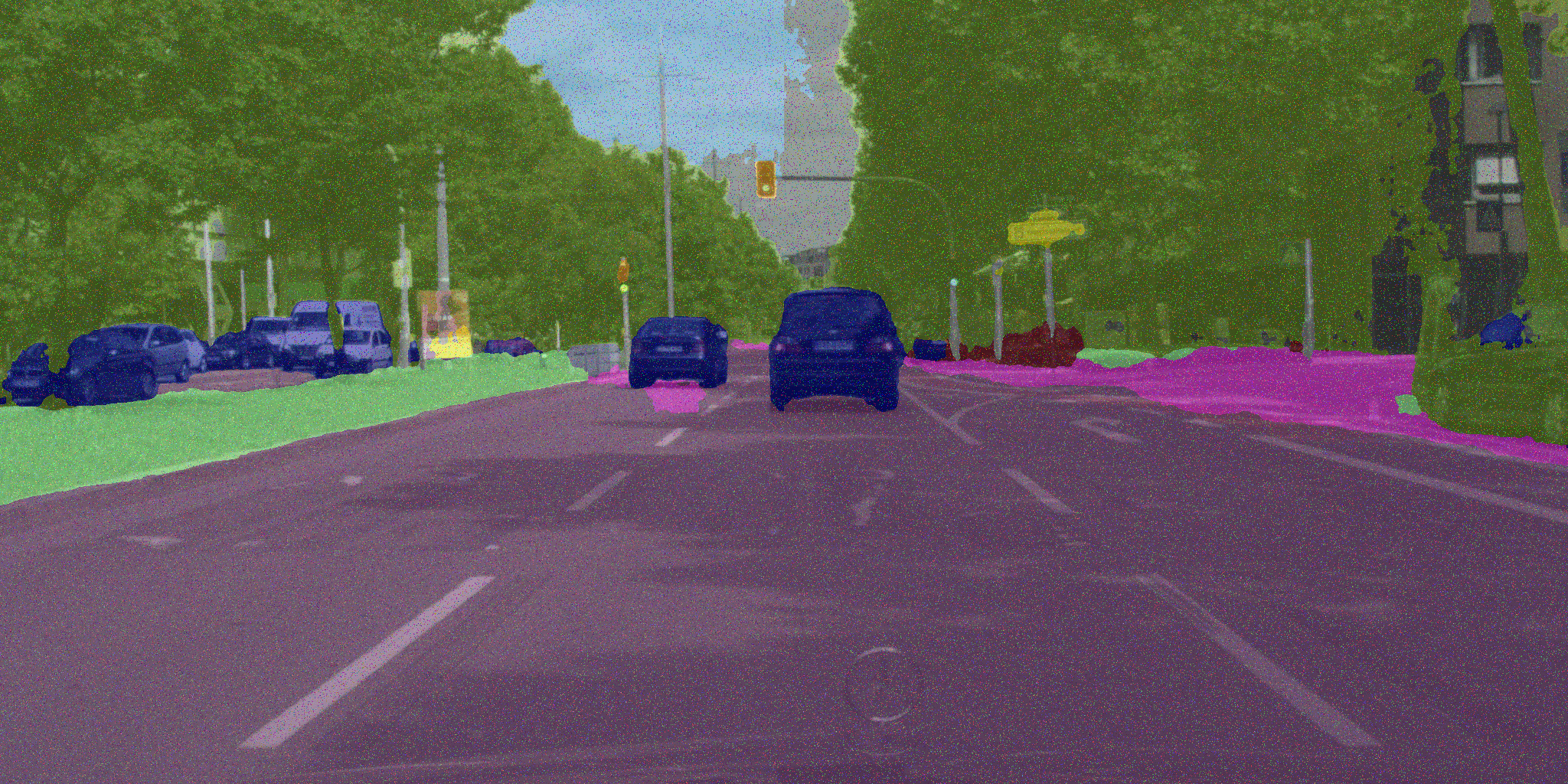}
        \caption{ConvNeXt}
        \label{fig:sub3}
    \end{subfigure}    
    
    \begin{subfigure}[b]{0.32\textwidth}
        \centering
        \includegraphics[width=\textwidth]{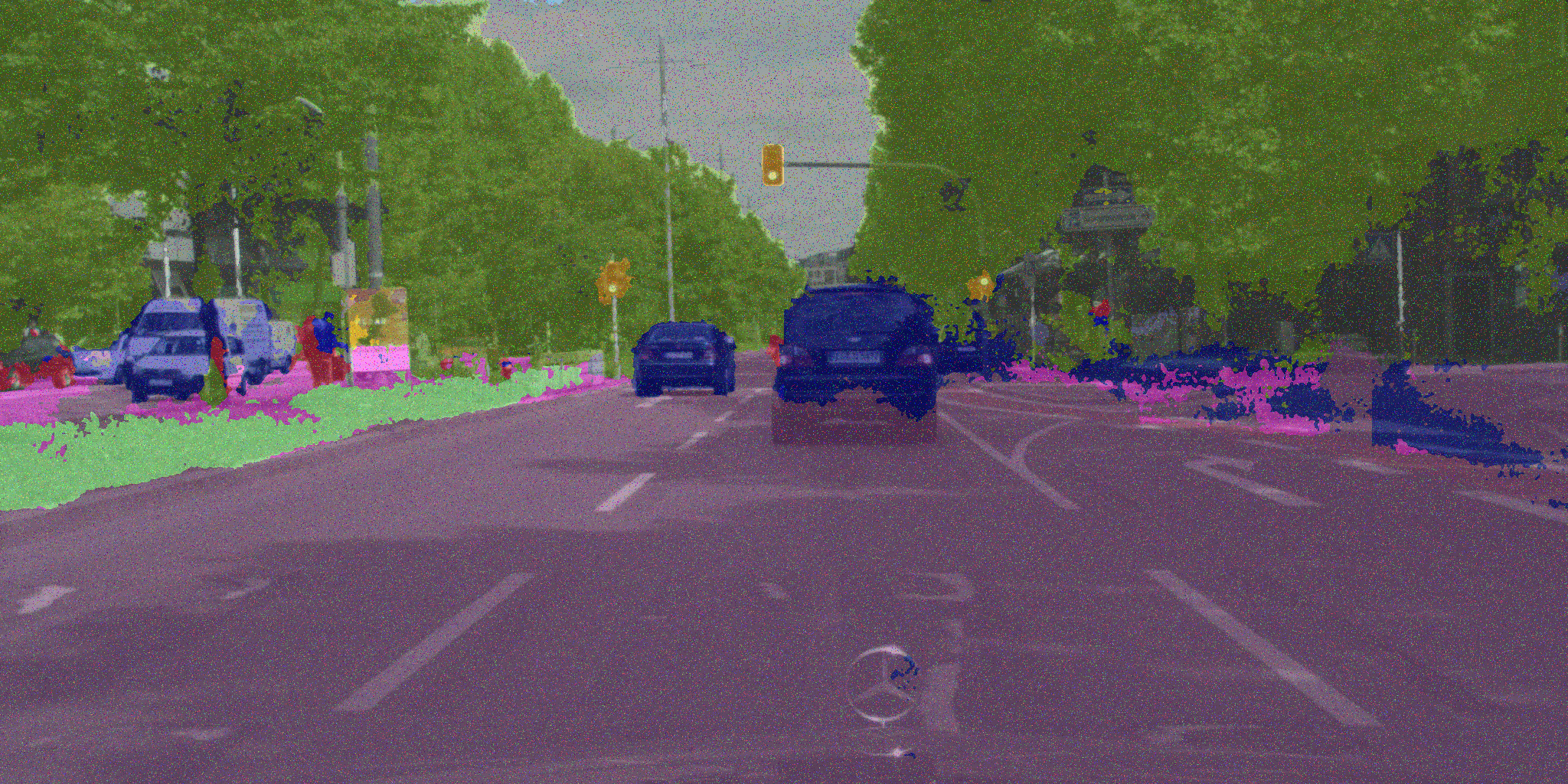}
        \caption{SegFormer-B0}
        \label{fig:sub4}
    \end{subfigure}  
    \begin{subfigure}[b]{0.32\textwidth}
        \centering
        \includegraphics[width=\textwidth]{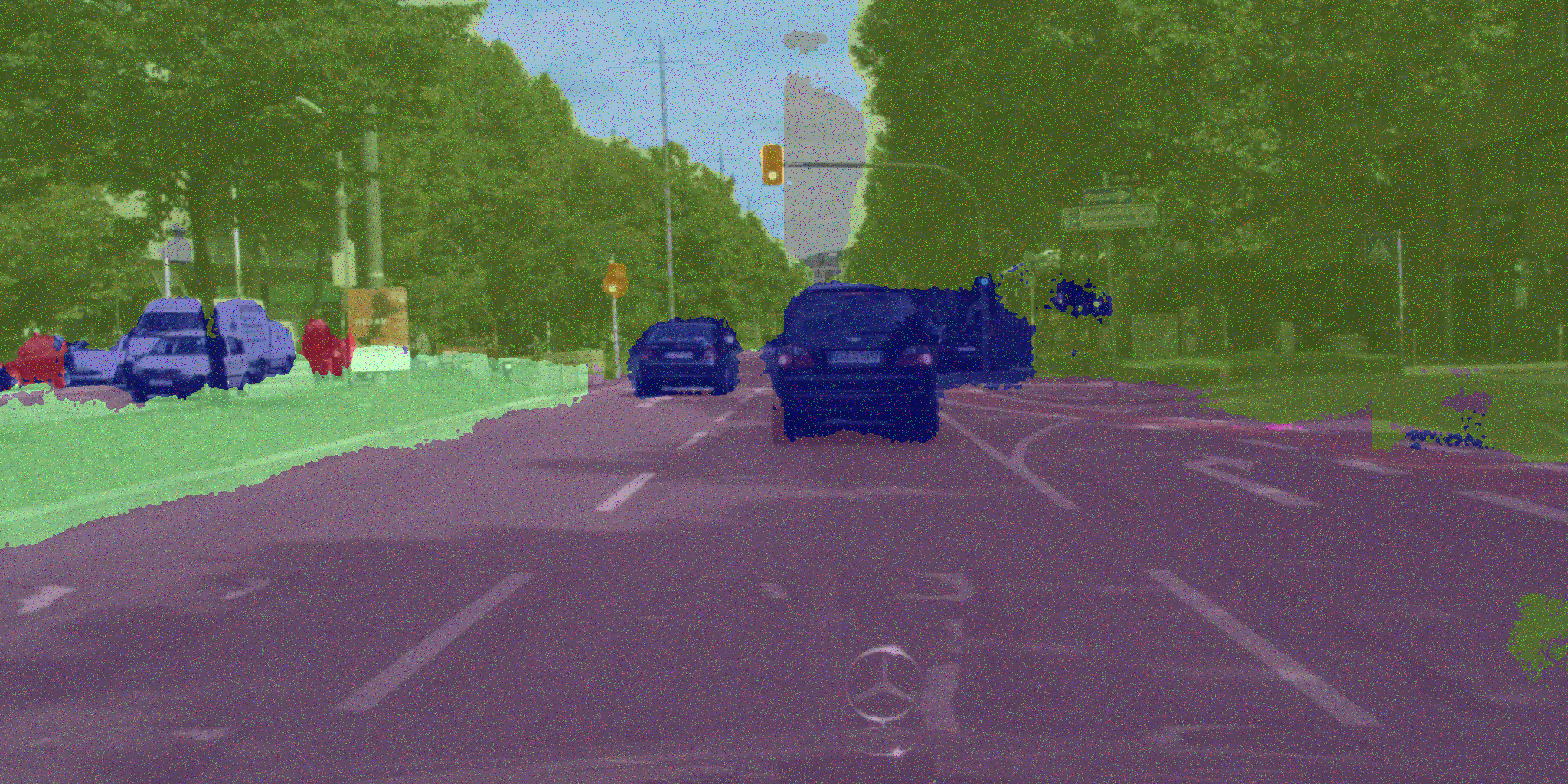}
        \caption{SegFormer-B0+AugMix}
        \label{fig:sub5}
    \end{subfigure}  
    \begin{subfigure}[b]{0.32\textwidth}
        \centering
        \includegraphics[width=\textwidth]{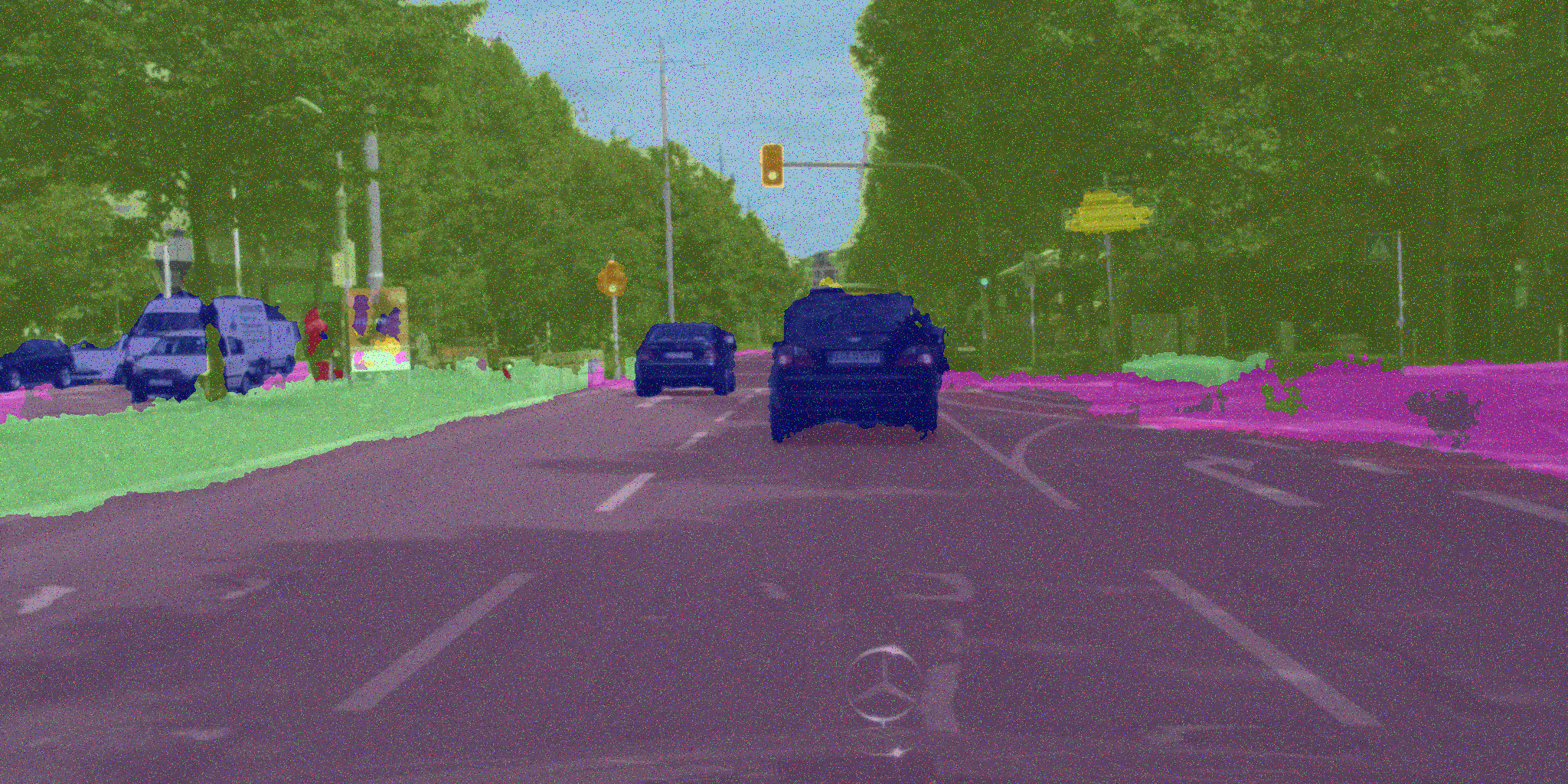}
        \caption{Ours:SegFormer-B0+CWFA}
        \label{fig:sub6}
    \end{subfigure} 
    
    \caption{\textbf{Example results.} Our method shows robustness improvements compared to existing approaches.}
    \label{fig: comparison methods to CWFA predictions}
\end{figure*}

\textbf{Comparison to state of the art methods.}
We focus now on evaluating the robustness of our approach compared to other robust algorithms in the literature. To this end, in Table~\ref{tab:sota_cityscape_c}, we compare our best performing method to related approaches on Cityscapes-C. Our approach achieves a new state-of-the-art performance in terms of robustness on the City-C dataset, as demonstrated by an average mIoU of $69.3\%$ when applied to SegFormer-B5. This outperforms newer methods such as FAN and FAN+STL, highlighting the effectiveness of our approach. In addition, even our SegFormer-B3 model with CWFA achieves similar robustness performance as the second previous state-of-the-art model, STL (FAN-B-Hybrid). Figure ~\ref{fig: comparison methods to CWFA predictions} compares multiple models' results on an impulse noise level 3 corrupted image to our CWFA-trained SegFormer-B0 model, showcasing robustness gains. Figure~\ref{fig:visual comparison cwfa B5} shows our SegFormer+CWFA model accurately segmenting a vehicle in a snow severity 5 image, where the SegFormer-B5 baseline fails.

\vspace{0.5cm}

\begin{table}[!t]
\caption{\textbf{SegFormer on Cityscapes.} As shown, our approach is able to generalize beyond the common set of corruptions in Cityscapes-C. Full table in section \ref{sec: complete tables}}
\label{table: segformer on city-c and city-c-bar}
\resizebox{\columnwidth}{!}{
\begin{tabular}{c|c|c|c|c|c|c} 
\multicolumn{2}{c|}{} & City & \multicolumn{2}{c|}{City-C} & \multicolumn{2}{c}{City-$\Bar{\text{C}}$ } \\
   \multicolumn{2}{c|}{} & mIoU  & Average & Retention & Average & Retention \\
  \hline
  \multirow{2}{*}{B0 / 3.7}& SegFormer & 76.3 & 49.2 & 64.5 & 43.1 & 56.6\\
 & Ours  & \textbf{76.4} & \textbf{57.4} & \textbf{75.1} & \textbf{49.5} & \textbf{64.8} \\
 \hline
  \multirow{2}{*}{B1 / 13.6}& SegFormer & \textbf{78.5} & 52.8 & 67.2 & 46.9 &  59.8 \\
 & Ours  & 78.2 & \textbf{61.7} & \textbf{78.9} & \textbf{53.4} & \textbf{68.3} \\
 \hline 
  \multirow{2}{*}{B2 / 27.3}& SegFormer &  81.0 & 58.3 & 72.0 & 52.8 & 65.2  \\
& Ours&  \textbf{81.2} &  \textbf{66.7} & \textbf{82.1} & \textbf{60.1} & \textbf{74.0}  \\
\hline
  \multirow{2}{*}{B3 / 47.2}& SegFormer &  80.9 & 60.7 & 75.1 & 56.1 & 68.7   \\ 
  & Ours & \textbf{81.5}  & \textbf{68.2} & \textbf{83.7} &  \textbf{61.8} &  \textbf{75.9}\\
\hline
\multirow{2}{*}{B4 / 64.0}& SegFormer &  \textbf{82.6} & 65.4 & 79.2 & 57.6 & 69.7  \\
& Ours &  82.1 & \textbf{69.0} & \textbf{84.0} & \textbf{63.9} & \textbf{77.8}  \\
\hline
 \multirow{2}{*}{B5 / 84.6}& SegFormer & \textbf{82.4} & 66.7 & 80.9 & 58.8 & 71.3  \\
& Ours &  82.2 & \textbf{69.3} & \textbf{84.3} & \textbf{63.9} & \textbf{77.8} \\
\end{tabular}
}
\vspace{-0.5cm}
\end{table}

\begin{table}[!t]
    \centering
      \caption{\textbf{Cityscapes}. Comparison to other CNNs and Transformer methods on Cityscapes and Cityscapes-C.}
         \label{tab:sota_cityscape_c}
        \resizebox{1.0\columnwidth}{!}{

        \begin{tabular}{l|cccc}
    Model  &  Encoder Size   & City       & City-C    &  Retention 
    \\
    \midrule
    DeepLabv3+ (R50)~\cite{kamann2020benchmarking}  &  25.4M  & 76.6 &  36.8  & 48.0\% 
     \\ 
    DeepLabv3+ (R101)~\cite{kamann2020benchmarking}  &  47.9M  & 77.1 &  39.4  &  51.1\%
     \\ 
    DeepLabv3+ (X65)~\cite{kamann2020benchmarking}  & 22.8M  & 78.4 &  42.7  &  54.5\%
     \\ 
    DeepLabv3+ (X71)~\cite{kamann2020benchmarking}  & -  & 78.6 &  42.5  &  54.1\%
     \\ 
     \midrule
    ICNet (\cite{zhao2018icnet})  &  -  & 65.9 &  28.0  & 42.5\% 
     \\ 
         FCN8s (\cite{long2015fully})  &  50.1M  & 66.7 &  27.4  & 41.1\% 
     \\ 
         DilatedNet (\cite{yu2015multi})  &  -  & 68.6 &  30.3  &  44.2\%
     \\ 
         ResNet38 (\cite{wu2019wider})  &  -  & 77.5 &  32.6  &  42.1\% 
     \\ 
         PSPNet (\cite{zhao2017pyramid})  &  13.7M  & 78.8 &  34.5  &  43.8\%
     \\ 
         ConvNeXt-T (\cite{liu2022convnet})  &  29.0M  & 79.0 &  54.4  &  68.9\% 
     \\ 
     \midrule
         SETR (\cite{heo2021rethinking}) &  22.1M  & 76.0 &  55.3  & 72.8\%  
     \\ 
         SWIN-T (\cite{liu2021swin})  &  28.4M  & 78.1 &  47.3  &  60.6\%
     \\ 
         SegFormer-B0 (\cite{xie2021segformer})  &  3.4M  & 76.2 &  48.8  &  64.0\%
     \\ 
         SegFormer-B1 (\cite{xie2021segformer})  & 13.1M  & 78.4 &  52.7  &  67.2\%
     \\
        SegFormer-B2 (\cite{xie2021segformer})  &  24.2M  & 81.0 &  59.6  &  73.6\%
     \\ 
         SegFormer-B5 (\cite{xie2021segformer})  &  81.4M  & 82.4 &  65.8  &  79.9\%
     \\ 
         SegFormer-B5\footnote{Our reproduced results} (\cite{xie2021segformer})  &  81.4M  & 82.4 &  66.7  &  80.9\% \\ 
      \midrule
        FAN-B-Hybrid~\cite{zhou2022understanding}   &  50.4M  & 82.2 &  66.9  &  81.5\% \\
        FAN-L-Hybrid~\cite{zhou2022understanding}   & 76.8M &  82.3  & 68.7   & {83.5\%}      \\
        \midrule
        \midrule
        STL (FAN-B-Hybrid)~\cite{fan2}   &  50.9M  & 82.5 &  68.6  & 83.2 \% \\
        STL (FAN-L-Hybrid)~\cite{fan2}   & 77.3M &  \textbf{82.8}  &  69.2& {83.6\%}      \\
        \midrule
        Ours: SegFormer-B3+CWFA   & 44.0M &  {81.5}  & 68.2  &  83.7\%   \\ 
        Ours: SegFormer-B5+CWFA   & 81.4M &  {82.2}  & \textbf{ 69.3}  & \textbf{ 84.3\% }  \\ 
    \end{tabular}
        }
\end{table}    

\textbf{SegFormer on ADE20K.} We evaluate the method's generalization to a more challenging dataset, ADE20K, containing up to 150 classes and known for its difficulty. State-of-the-art large vision models like InternImage (1300M parameters) ~\cite{wang2023internimage} achieve only 62.9 mIoU on clean validation data in ADE20K, compared to 76.4 mIoU for the smaller SegFormer-B0 on Cityscapes. Due to this inherent difficulty, when using CWFA, the absolute robustness improvements in mIoU are understandably lower for ADE20K. However, our method demonstrates remarkable relative robustness gains. The improvements in retention rate remain significant, with $7.9\%$ and $5.9\%$ gains for SegFormer-B0 and B5, respectively.Interestingly, for B5, the retention rate gain is higher than on City-C, with an additional increase of $2.5\%$. These results confirm our model's effectiveness in generalizing to new datasets with a larger number of classes, even under extreme challenges.
 
\begin{table}[!t]
\caption{\textbf{SegFormer on ADE20k-C}.Our CWFA approach yields important improvements in terms of the retention rate on this challenging dataset}
\centering
\resizebox{0.75\columnwidth}{!}{
\begin{tabular}{c|c|c|c|c} 
   \multicolumn{2}{c}{} & mIoU  & Average & Retention \\
  \hline
  \multirow{2}{*}{B0}& SegFormer& \textbf{37.1} & 22.7 & 61.1\\

 & Ours   & 36.7 & \textbf{25.3} & \textbf{69.0} \\
 \hline
  \multirow{2}{*}{B1}& SegFormer & \textbf{41.9} & 26.9 & 64.2   \\
 
 & Ours  & 40.3 & \textbf{29.4} & \textbf{73.0} \\
 \hline 
  \multirow{2}{*}{B2}& SegFormer  & \textbf{46.0} & 33.8 & 73.4 \\

& Ours & 45.5 & \textbf{36.0} &  \textbf{79.2} \\
\hline
  \multirow{2}{*}{B3}& SegFormer   & \textbf{48.1} & 37.4 & 77.6 \\ 

  & Ours  & 47.3 & \textbf{38.2} &  \textbf{80.9}\\
\hline
\multirow{2}{*}{B4}& SegFormer  & \textbf{50.1} & 38.3 & 76.3 \\

& Ours & 48.8 & \textbf{39.8} & \textbf{81.7} \\
\hline
 \multirow{2}{*}{B5}& SegFormer & \textbf{50.9} & 38.4 & 75.4 \\
 
& Ours  &  49.8 & \textbf{40.5} & \textbf{81.3} \\
\end{tabular}
}
\label{tab:cwfa on ade20k}
\end{table}

\begin{table}[!t]   
    \caption{\textbf{Comparison to SFA and PixMix.} Our approach is significantly better than SFA, the closest feature augmentation approach in the literature.}
        \label{tab: SFA experiment}
        \centering
        \resizebox{1.0\columnwidth}{!}{
        \begin{tabular}{c|c|c|c}
        \toprule
        & \multicolumn{3}{c}{Cityscapes}\\
        Model & Clean & Corrupt. & Reten. \\
        \midrule
        SegFormer B0  & \textbf{76.3} & 49.2 & 64.5 \\
        SegFormer B0 + SFA~\cite{li2021simple_compact} & 76.3 & 48.8  & 63.9 \\
        SegFormer B0 + PixMix~\cite{hendrycks2022pixmix_compact} & 75.7  & \textbf{56.7} & \textbf{74.8} \\
        SegFormer B0 + CWFA (Ours) & 75.6 & 56.3 & 74.5 \\
        \hline
        SegFormer B1  & \textbf{78.5} & 52.8 & 67.2 \\
        SegFormer B1 + SFA~\cite{li2021simple_compact} & 78.1 & 54.1 & 69.2 \\
        SegFormer B1 + PixMix~\cite{hendrycks2022pixmix_compact} & 78.4 & 61.1  & 78.0 \\
        SegFormer B1 + CWFA (Ours) & 78.2 & \textbf{61.7} & \textbf{78.9} \\
        \bottomrule
        \end{tabular}
        }    
\end{table}

\textbf{Twins and Swin Transformers on Cityscapes.}
\label{sec:extending CWFA to other Vision Transformers}
We test CWFA on the small versions of Twins, Twins-PCPVT-Small and Twins-SVT-Small. In addition, we also test our method on Swin-Transformer tiny. We proceed as for the SegFormer comparisons. Table \ref{tab:cwfa on other Vision Transformers} summarizes our results on the Cityscapes-C dataset. We observe that CWFA improves the robustness on all models. The average mIoU is improved by $5.9 \%$ and $3.7 \%$ for Twins-PCPVT and Twins-SVT respectively. For Swin-T, the average mIoU is improved by $1.2 \%$. These findings show that CWFA also works on other Vision Transformers. Moreover, it also suggests that CWFA works best for Transformers using a global attention mechanism such as SegFormer or Twins, in contrast to Swin, which makes use of local windows. The fact that using local windows might hurt robustness was already addressed in \cite{zhou2022understanding}.  

\textbf{Comparison to other augmentation techniques.}\label{sec:comparison to AugMix} AugMix is a simple yet effective image augmentation method originally designed to improve robustness of image classification networks. 
In this experiment, we instantiate AugMix for semantic segmentation following a similar approach to the one we used on CWFA. The baseline SegFormer models are fine-tuned for 160k iterations using AugMix. We adhere to the original paper's recipe and employ the same AugMix parameters. It is worth mentioning that the choice of these parameters has been shown to have minimal impact on AugMix's performance. To simplify the process, we exclude training with the Jensen-Shannon divergence (JSD) loss, as the robustness gains primarily stem from the mixing image augmentation technique \cite{hendrycks2019augmix}. This enables us to directly compare the performance of CWFA, a feature augmentation technique, with that of an image augmentation technique. We compare both methods on Cityscapes-C and ADE20K-C. Table~\ref{tab:cwfa comparison state of the art} summarizes our results. Overall, CWFA achieves better robustness performance than AugMix on both datasets. Specially, for the smallest SegFormer-B0 model, CWFA outperforms AugMix in terms of robustness by $3.7\%$ average mIoU on Cityscapes-C.
We provide additional comparisons to SFA and PixMix. The former injects Gaussian noise to the features. The latter is a similar augmentation approach to AugMix. As shown in Table~\ref{tab: SFA experiment}, our method clearly outperforms SFA in terms of robustness. We acknowledge that our method is on par with PixMix and with the larger SegFormer models trained with AugMix in terms of robustness. However, as we will show next, CWFA significantly outperforms AugMix and PixMix in terms of efficiency. 

\begin{table}[!t]    
        \caption{ \textbf{Twins and Swin Transformers on Cityscapes-C.}: Our CWFA approach also improves the robustness of other Vision Transformers models }
        \label{tab:cwfa on other Vision Transformers}
        \centering    
        \resizebox{1.0\columnwidth}{!}{
        \begin{tabular}{l|c|c|c}
        \toprule
        & \multicolumn{3}{c}{Cityscapes}\\
        Model & Clean & Corrupt. & Reten. \\
        \midrule
        Swin- T  & \textbf{80.0} & 51.6 &  64.5\\
        Swin- T + CWFA (Ours) & 79.6 & \textbf{52.8}  &  \textbf{66.3}\\
        \hline
        Twins-PCPVT & 79.5 & 53.0 & 66.6 \\
        Twins-PCPVT + CWFA (Ours) & \textbf{79.6} & \textbf{58.9} & \textbf{73.9} \\
        \hline
        Twins-SVT & \textbf{80.9} & 59.3 & 73.3 \\
        Twins-SVT+ CWFA (Ours) & 78.9 & \textbf{63.0} & \textbf{79.9} \\
        \bottomrule
        \end{tabular}
        }
\end{table}

\textbf{Computational efficiency of CWFA.}\label{sec:CWFA compute efficient} We compare the computational cost of performing CWFA with respect to not performing it, and to using image augmentations such as AugMix. Table \ref{tab:cwfa comparison state of the art} compares the efficiency in terms of sec/epoch when training on Cityscapes. Our measurement shows that while AugMix increases training time by $47\%$, CWFA only increases it by $2\%$. Additional measurements when using PixMix resulted in an increase training time of $49\%$, similar to AugMix. Our finding emphasizes the computational efficiency of CWFA and the advantages of operating in the lower dimensional global averaged feature space.

\begin{table}[t]
\caption{\textbf{Comparison to AugMix}: Comparison between SegFormer B0 to B5 baseline models and models fine-tuned with CWFA and AugMix for Cityscapes-C and ADE20K-C, together with the training efficiency comparison.}
\centering
\resizebox{0.9\columnwidth}{!}{%
\begin{tabular}{cccc|ccc}
\toprule
& \multicolumn{3}{c|}{Cityscapes-C} & \multicolumn{3}{c}{ADE20K-C}\\
  & SegFormer & AugMix & Ours &  SegFormer & AugMix & Ours    \\
\midrule
\midrule
\multicolumn{7}{c}{Average mIoU} \\
 B0 & 49.2 & 52.2 & \textbf{57.4} & 22.7 & 24.0  &  \textbf{25.3}  \\
 B1 & 52.8 & 57.4 &  \textbf{61.7} & 26.9 & 28.4 &  \textbf{29.4} \\
 B2 & 58.3 & 64.4 & \textbf{66.7} & 33.8 & 35.1 &   \textbf{36.0} \\
 B3 & 60.7 & 67.1 & \textbf{68.2} & 37.4 & 37.9 &  \textbf{38.2} \\
 B4 & 65.4 & 68.9 & \textbf{69.0} & 38.3  & 39.3 & \textbf{39.8} \\
 B5 & 66.7 & 69.0 & \textbf{69.3} & 38.4 & 38.6 & \textbf{40.5}\\
 \midrule
 \midrule
 \multicolumn{7}{c}{Training Efficiency (sec / epoch)} \\
 B0 & 443 & 654 & 452 & \\
\bottomrule
\end{tabular}
}
\label{tab:cwfa comparison state of the art}
\vspace{-0.5cm}
\end{table}
\vspace{0.5cm}

\subsection{Ablation studies}\label{sec:ablation studies}
Let us now further analyze the behaviour of our feature augmentation approach. The experiments in this section include sensitivity to $\epsilon$, sensitivity to applying augmentation to different encoders and additional insights of the benefits of CWFA for compact models. We also include an analysis of transferability of the perturbations. For these experiments, for computational reasons, we use pretrained models publicly available trained for 160k iterations and fine-tune them for 16k iterations with CWFA. Original models were trained on Cityscapes and we report results for the validation set as well as for Cityscapes-C.

\begin{figure}[!t]
\begin{center}
    \centering
    \begin{subfigure}[b]{0.45\columnwidth}
        \centering
        \includegraphics[width=\textwidth]{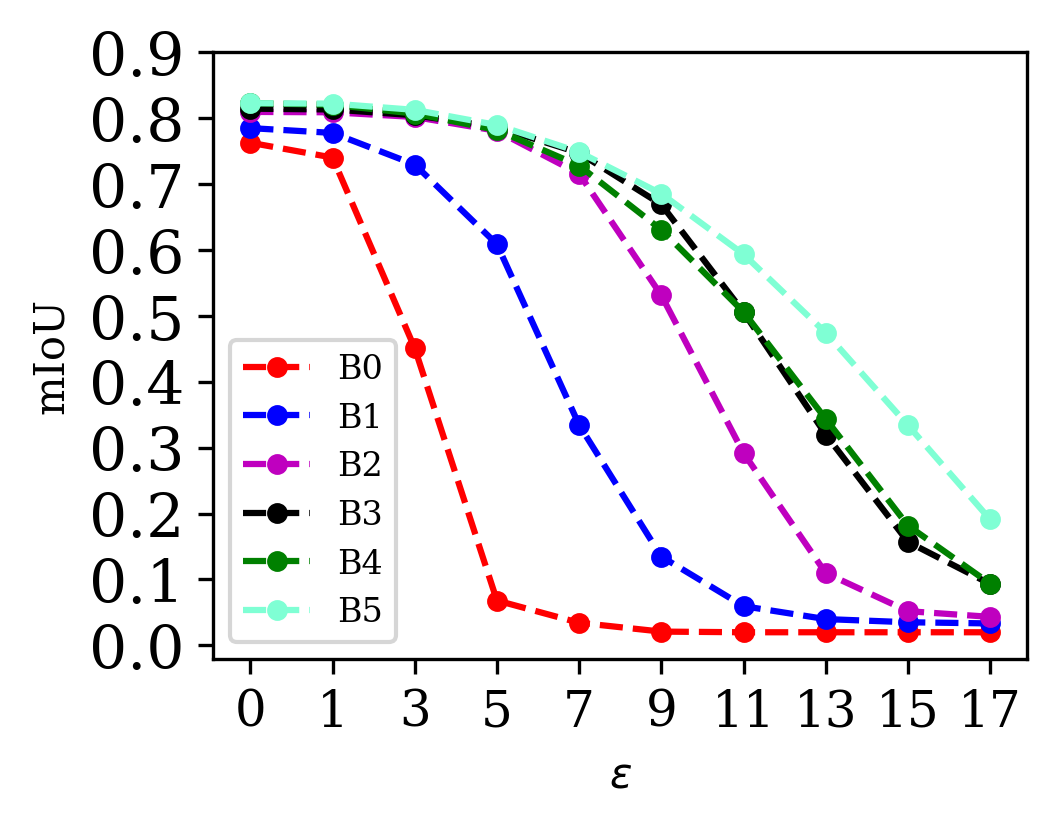}
        \caption{}
        \label{fig: choice of eps}
    \end{subfigure}    
    \begin{subfigure}[b]{0.45\columnwidth}
        \centering
        \includegraphics[width=\textwidth]{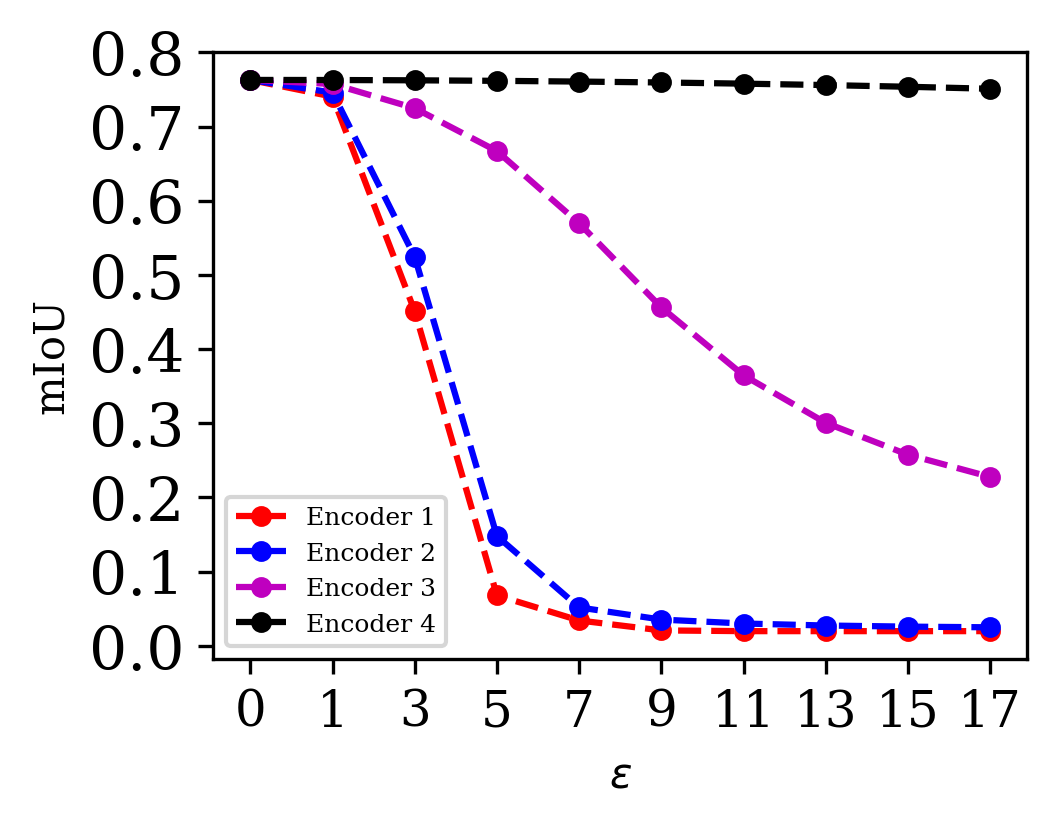}
        \caption{}
        \label{fig:segformer B0 befor and after CWFA sensitivity}
    \end{subfigure}  
    \vspace{-2mm}
    \caption{\textbf{Sensitivity of SegFormer baseline models towards CWFA and choice of $\epsilon$.} Sensitivity of SegFormer models when perturbing the feature with different $\epsilon$ values during inference on the Cityscapes validation set. a) Sensitivity as a function of the model size when applying perturbing the features from the first encoder of the original models. b) Sensitivity of SegFormer-B0 as a function of the encoder for the original model and a model fine-tuned with CWFA.}
    \label{fig:CWFAsensitivities}
\end{center}
\end{figure}

\textbf{Sensitivity to the strength of the perturbation ($\epsilon$).}\label{subsec: choice of hyperparameters} This parameter controls the strength of the perturbation. Here, we analyze the sensitivity of the approach to it as a function of the model size and as a function of the encoder where we apply the perturbation. To analyze the robustness to out of distribution as a function of the model size, given the original models (trained without additional augmentations nor CWFA), we inject CWFA perturbations to the features at each encoder using different $\epsilon$ values during inference on the Cityscapes validation set and report the corresponding mIoU score. Figure~\ref{fig:CWFAsensitivities}a shows results for different model sizes perturbed only on the first encoder and Figure~\ref{fig:CWFAsensitivities}b shows results for perturbations to the different encoders of SegFormer-B0. As we can see, the performance degradation increases as the strength of the perturbation increases. The accuracy drop is very significant for compact models, particularly B0, and smoother as the size increases. This is expected as empirically has been shown that larger models tend to be more robust to compact ones. We also observe (Figure~\ref{fig:CWFAsensitivities}b) that the network is more sensitive at the beginning and the effect of the perturbation is less prominent at deeper stages.

\begin{figure}[!t]
    \centering
    \vspace{-3mm}
    \includegraphics[width=0.5\linewidth]
    {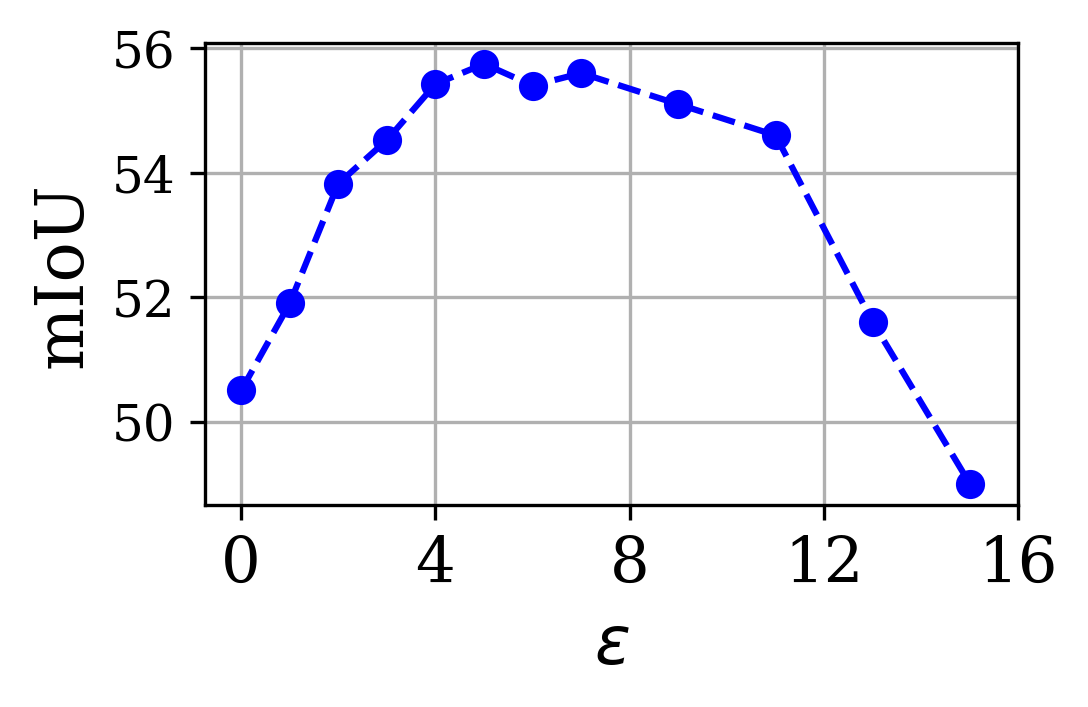}
    \vspace{-0.5cm}
    \caption{\textbf{Robustness as a function of $\epsilon$.} Sensitivity of SegFormer-B0 fine-tuned using CWFA for different perturbation strength. Evaluation on City-C. Our approach is not very sensitive to the choice of $\epsilon$ as we obtain similar robustness gains when choosing $\epsilon$ in a wide range of values}
    \vspace{-0.5cm}
    \label{fig:robustB0epsilon}
\end{figure}

Finally, we analyze the robustness benefits for SegFormer-B0 fine-tuned with CWFA as a function of the strength of the perturbation. We fine tune for 16k iterations the SegFormer-B0 baseline with CWFA and in each fine tuning, we use a different fixed $\epsilon$  value and report the corresponding averaged mIoU score on City-C. Figure~\ref{fig:robustB0epsilon} shows the results for this experiment. As expected, for lower values of $\epsilon$ there is no effective improvement as well as for extreme perturbations. For a good range of values, there improvement is quite significant with a maximum around $\epsilon=5$.

\textbf{Sensitivity to the encoder.} Our global average feature augmentation is generic and can be applied to any encoder in the architecture. Our SegFormer experiments in the main section applied CWFA to the features provided by the four encoders. To analyze the sensitivity of the results, we fine-tune four SegFormer-B0 models with CWFA applied to different encoders. Specifically, we apply CWFA to encoders $\{\text{enc}_1\}$, $\{\text{enc}_1, \text{enc}_2\}$, $\{\text{enc}_1, \text{enc}_2, \text{enc}_3\}$, and $\{\text{enc}_1, \text{enc}_2, \text{enc}_3, \text{enc}_4\}$, respectively. We use $\epsilon=4$ and $p_{\text{augm}}=0.3$ on the corresponding encoders. It is important to note that CWFA is only applied to one of the encoders to maintain the ratio between training on clean and perturbed data. Table~\ref{tab:cwfa applied on different encoders} summarizes the results for this experiment. As we can see, applying CWFA to all the encoders yields the largest robustness gains with a negligible regression on the clean accuracy. The most significant improvement is when the augmentation is applied to the first encoder. 

\textbf{The channel-wise perturbation.} To verify our intuition that corruptions affect all features within a channel block equally, we compared CWFA with a variant that removes the global average pooling, perturbing all features as $\mathbf{\Tilde{X}}= \mathbf{X} + \epsilon \dfrac{\mathbf{X}}{\|\boldsymbol{\mathbf{X}}\|_2}$ where $\mathbf{X} \in \mathbb{R}^{C \times H \times W}$ is the full feature representation. As shown in Table~\ref{tab:cwfa without average pool}, removing this pooling operation eliminates the robustness gains of CWFA, achieving performance similar to the baseline SegFormer-B0 and B1 models. This emphasizes that computing the perturbation in a channel-wise manner is key to CWFA's effectiveness

\textbf{Using the feature itself.} We further explored the significance of utilizing the channel-wise feature itself as the perturbation in our experiments. To assess this, we trained SegFormer-B0 and B1 models by applying channel-wise Gaussian noise (with zero mean and unit variance) as the perturbation method. While this approach resulted in improved robustness compared to the baseline, as shown in Table~\ref{tab:cwfa without average pool}, it did not achieve the level of robustness observed with CWFA. This finding suggests that the normalized channel-wise feature is particularly effective at capturing common corruptions along the channel dimension, which is crucial for enhancing robustness in CWFA. The superior performance of CWFA underscores the importance of this channel-wise normalization in our method, highlighting its role as a key factor in the success of our approach.

\begin{table}[!t]    
        \caption{\textbf{Performance as a function of the encoder.} Effect of applying CWFA on different encoders of SegFormer-B0 with CWFA $\epsilon=4$. Report mIoU on Cityscapes validation, average mIoU and average retention rate on Cityscapes-C. First model refers to the B0 baseline model trained without CWFA}
        \label{tab:cwfa applied on different encoders}
    \centering    
    \vspace{-0.25cm}
        \resizebox{0.65\columnwidth}{!}{
        \begin{tabular}{cccc}
        \toprule
        CWFA on & \multicolumn{3}{c}{Cityscapes}\\
        Encoder & Clean & Corrupt. & Reten. \\
        \midrule
        - & \textbf{76.4} & 50.5 &  66.1\\
        1 &  76.2 & 53.6 &  70.4\\
        1-2 &  \textbf{76.4} & 55.9 & 73.2 \\
        1-3 & 76.2 & 55.8 &  73.2\\
        1-4 & 75.9 & \textbf{56.2} & \textbf{74.0} \\
        \bottomrule
        \end{tabular}
        }
    
\end{table}
\begin{table}[!t]        
    \caption{\textbf{Ablation study: Average Pooling and Gaussian noise.} CWFA performance variations when disabling the two main blocks: Average pooling and using Gaussian Noise as perturbation. As shown, each component is essential for the overall performance.}
    \label{tab:cwfa without average pool}
    \centering    
    \vspace{-0.3cm}
        \resizebox{0.75\columnwidth}{!}{        
        \begin{tabular}{cccc}
        \toprule
        SegFormer  & \multicolumn{3}{c}{Cityscapes}\\
          & Clean & Corrupt. & Reten. \\
        \midrule
        B0 & \textbf{76.4} & 50.5 &  66.1\\
        B0 + CWFA &  75.9 & \textbf{56.2} & \textbf{74.0} \\
        B0 + CWFA without average Pooling & 75.8 & 50.6 & 66.7\\
        B0 + Channel Wise Gaussian Noise & 75.7 & 51.8 & 68.4 \\
        \bottomrule
        B1 & \textbf{78.5} & 52.8 & 67.2 \\
        B1 + CWFA & 78.2  & \textbf{61.7} & \textbf{78.9} \\
        B1 + CWFA without average Pooling & 78.3 & 52.7 & 67.3\\
        B1 + Channel Wise Gaussian Noise & 78.3 & 56.7 & 72.4\\
        \bottomrule
        
        \end{tabular}
        }  
    \vspace{-0.3cm}
\end{table}

\begin{table}[!t]
\caption{\textbf{Transferability of corruptions:} Comparison of SegFormer-B0 models fine-tuned with each corruption of Cityscapes-C to SegFormer-B0+CWFA $\epsilon=4$}
\label{table: transerability of corruptions}
\centering
\resizebox{1.0\columnwidth}{!}{
\begin{tabular}{|c|c|c|c|c|c|} 
  \hline
  SegFormer-B0 & Clean & Avg & Reten.  & B0 with CWFA  & B0 with CWFA   \\
    with & & w/o corr & w/o corr & Avg. w/o corr & Reten. w/o corr  \\
  \hline
 - (Baseline) & 76.3  & 49.2 & 64.5 &  - & -\\
\hline
  Motion Blur  & 75.8 & 47.7 & 63.0 & \textbf{55.2} & \textbf{72.9} \\
\hline
Defocus Blur   & 76.1 & 47.1 & 61.9 & \textbf{55.2} & \textbf{72.9}  \\
\hline
 Gaussian Blur  & 76.3 &  48.5 & 63.6 & \textbf{55.2} & \textbf{72.9}  \\
\hline
  Gaussian Noise  & 75.8 & 56.9 & 75.1 & \textbf{58.3} & \textbf{77.1} \\
\hline
 Impulse Noise   & 75.6 & 56.8 & 75.1 & \textbf{58.3} & \textbf{77.1}  \\
\hline
  Shot Noise & 75.5 & 56.4 & 74.7 &  \textbf{58.3} & \textbf{77.1}  \\
\hline
  Speckle Noise  & 75.9 & 55.9 & 73.7 &  \textbf{58.3} & \textbf{77.1}  \\
\hline
  Brightness   & 76.0  & 47.7 & 62.7 & \textbf{55.2} &  \textbf{73.0} \\
\hline
  Contrast  & 76.2  & 50.0 & 65.6 & \textbf{55.7} & \textbf{73.6} \\
\hline
 Saturate & 76.6 & 48.8 & 63.7 & \textbf{55.4} & \textbf{73.2} \\
\hline
 JPEG compression  &  75.3  & 51.3 & 68.1 & \textbf{57.2} & \textbf{75.6} \\
\hline
  Snow & 75.7 & 57.3 & 75.7 & \textbf{57.5} &  \textbf{76.1}\\
\hline
  Spatter  &  75.6 & 51.8 & 68.5 & \textbf{56.0} &  \textbf{74.1}\\
\hline
 Fog & 75.8 & 48.0 & 63.4 &  \textbf{55.8} & \textbf{73.7} \\
\hline
 Frost  & 75.4 &  52.9 & 70.2 & \textbf{57.4} & \textbf{75.8} \\
\hline
\end{tabular}}
\vspace{-0.7cm}
\end{table}

\textbf{Transferability of Perturbations.} We could argue that any augmentation approach breaks the zero-shot robustness assumption as the network has seen similar perturbations during training. Here, we provide insights of this effect by adding each of the perturbations to the training process and then, comparing the performance of those models with the performance of models trained only using CWFA. Precisely, for each corruption in Cityscapes-C, we train a SegFormer-B0 model where that corruption has been added to the set of training augmentations and report the accuracy on CityScapes-C. We compare those numbers to a SegFormer-B0 trained only using CWFA. We report the main numbers in~\ref{table: transerability of corruptions} without the corresponding corruption used during the fine-tuning process. We also provide a comprehensive table including the performance per corruption in the supplementary material. The first thing to note from the experiments is that, as expected, there is a clear boost for exactly the same corruption used during training. However, there is a poor transferability between corruptions. Noise seems to be the perturbation that better improves accuracy overall, which is inline with prior works~\cite{rusak2020simple}. More importantly, we can observe in Table~\ref{table: transerability of corruptions} that CWFA yields a significantly better boost in performance than all the other corruptions in terms of mIoU and retention rate. From these results, we can conclude that CWFA is a more effective method than adding augmentations at the image level to obtain robust models.

\section{Conclusion}
\label{sec:conclusion}
In this paper, we propose CWFA, a plug-in feature augmentation technique for robust semantic segmentation with Transformers. The key novelty of CWFA is to use a global feature to estimate a common perturbation for all the features in the encoder. Our extensive experiments demonstrate the method’s versatility and generalizability across various Vision Transformer architectures, including Swin and Twins, and datasets such as Cityscapes and ADE20K. For instance, CWFA improves the SegFormer-B1 performance on the impulse noise corrupted Cityscapes-C data from $23.7\%$ to $51.4\%$ mIoU. CWFA consistently outperforms the robustness of AugMix and is on par with newer methods such as PixMix for SegFormer models while being $45\%$ and $49\%$ more efficient respectively. For large models, our results set a new state of the art for robustness in semantic segmentation.

{\small
\bibliographystyle{ieee_fullname}
\bibliography{11_references}

\begin{thebibliography}{10}\itemsep=-1pt

\bibitem{bu2023towards}
Qingwen Bu, Dong Huang, and Heming Cui.
\newblock Towards building more robust models with frequency bias.
\newblock In {\em Proceedings of the IEEE/CVF International Conference on Computer Vision}, pages 4402--4411, 2023.

\bibitem{chen2018encoder}
Liang-Chieh Chen, Yukun Zhu, George Papandreou, Florian Schroff, and Hartwig Adam.
\newblock Encoder-decoder with atrous separable convolution for semantic image segmentation.
\newblock In {\em ECCV}, 2018.

\bibitem{chu2021twins}
Xiangxiang Chu, Zhi Tian, Yuqing Wang, Bo Zhang, Haibing Ren, Xiaolin Wei, Huaxia Xia, and Chunhua Shen.
\newblock Twins: Revisiting the design of spatial attention in vision transformers.
\newblock {\em NeurIPS}, 2021.

\bibitem{cordts2016cityscapes}
Marius Cordts, Mohamed Omran, Sebastian Ramos, Timo Rehfeld, Markus Enzweiler, Rodrigo Benenson, Uwe Franke, Stefan Roth, and Bernt Schiele.
\newblock The cityscapes dataset for semantic urban scene understanding.
\newblock In {\em CVPR}, 2016.

\bibitem{devries2017improved}
Terrance DeVries and Graham~W Taylor.
\newblock Improved regularization of convolutional neural networks with cutout.
\newblock {\em arXiv preprint arXiv:1708.04552}, 2017.

\bibitem{erichson2022noisymix}
N~Benjamin Erichson, Soon~Hoe Lim, Francisco Utrera, Winnie Xu, Ziang Cao, and Michael~W Mahoney.
\newblock Noisymix: Boosting robustness by combining data augmentations, stability training, and noise injections.
\newblock {\em arXiv preprint arXiv:2202.01263}, 2022.

\bibitem{hendrycks2022pixmix_compact}
Hendrycks et al.
\newblock Pixmix: Dreamlike pictures comprehensively improve safety measures.
\newblock In {\em CVPR}, 2022.

\bibitem{li2021simple_compact}
Li et al.
\newblock A simple feature augmentation for domain generalization.
\newblock In {\em ICCV}, 2021.

\bibitem{gardner2015deep}
Jacob~R Gardner, Paul Upchurch, Matt~J Kusner, Yixuan Li, Kilian~Q Weinberger, Kavita Bala, and John~E Hopcroft.
\newblock Deep manifold traversal: Changing labels with convolutional features.
\newblock {\em arXiv preprint arXiv:1511.06421}, 2015.

\bibitem{hendrycks2021many}
Dan Hendrycks, Steven Basart, Norman~Mu Mu, Saurav Kadavath, Frank Wang, Evan Dorundo, Rahul Desi, Tyler Zhu, Samyak Parajuli, Mike Guo, et~al.
\newblock The many faces of robustness: A critical analysis of out-of-distribution generalization.
\newblock In {\em ICCV}, 2021.

\bibitem{hendrycks2018benchmarking}
Dan Hendrycks and Thomas Dietterich.
\newblock Benchmarking neural network robustness to common corruptions and perturbations.
\newblock In {\em ICLR}, 2018.

\bibitem{hendrycks2019augmix}
Dan Hendrycks, Norman Mu, Ekin~Dogus Cubuk, Barret Zoph, Justin Gilmer, and Balaji Lakshminarayanan.
\newblock Augmix: A simple data processing method to improve robustness and uncertainty.
\newblock In {\em ICLR}, 2019.

\bibitem{hendrycks2022pixmix}
Dan Hendrycks, Andy Zou, Mantas Mazeika, Leonard Tang, Bo Li, Dawn Song, and Jacob Steinhardt.
\newblock Pixmix: Dreamlike pictures comprehensively improve safety measures.
\newblock In {\em Proceedings of the IEEE/CVF Conference on Computer Vision and Pattern Recognition}, pages 16783--16792, 2022.

\bibitem{heo2021rethinking}
Byeongho Heo, Sangdoo Yun, Dongyoon Han, Sanghyuk Chun, Junsuk Choe, and Seong~Joon Oh.
\newblock Rethinking spatial dimensions of vision transformers.
\newblock In {\em ICCV}, pages 11936--11945, 2021.

\bibitem{kamann2020benchmarking}
Christoph Kamann and Carsten Rother.
\newblock Benchmarking the robustness of semantic segmentation models.
\newblock In {\em CVPR}, 2020.

\bibitem{li2021simple}
Pan Li, Da Li, Wei Li, Shaogang Gong, Yanwei Fu, and Timothy~M Hospedales.
\newblock A simple feature augmentation for domain generalization.
\newblock In {\em ICCV}, 2021.

\bibitem{lim2021noisy}
Soon~Hoe Lim, N~Benjamin Erichson, Francisco Utrera, Winnie Xu, and Michael~W Mahoney.
\newblock Noisy feature mixup.
\newblock In {\em ICLR}, 2021.

\bibitem{liu2021swin}
Ze Liu, Yutong Lin, Yue Cao, Han Hu, Yixuan Wei, Zheng Zhang, Stephen Lin, and Baining Guo.
\newblock Swin transformer: Hierarchical vision transformer using shifted windows.
\newblock In {\em CVPR}, 2021.

\bibitem{liu2022convnet}
Zhuang Liu, Hanzi Mao, Chao-Yuan Wu, Christoph Feichtenhofer, Trevor Darrell, and Saining Xie.
\newblock A convnet for the 2020s.
\newblock In {\em CVPR}, 2022.

\bibitem{long2015fully}
Jonathan Long, Evan Shelhamer, and Trevor Darrell.
\newblock Fully convolutional networks for semantic segmentation.
\newblock In {\em CVPR}, 2015.

\bibitem{mao2022towards}
Xiaofeng Mao, Gege Qi, Yuefeng Chen, Xiaodan Li, Ranjie Duan, Shaokai Ye, Yuan He, and Hui Xue.
\newblock Towards robust vision transformer.
\newblock In {\em CVPR}, 2022.

\bibitem{michaelis2019benchmarking}
Claudio Michaelis, Benjamin Mitzkus, Robert Geirhos, Evgenia Rusak, Oliver Bringmann, Alexander~S Ecker, Matthias Bethge, and Wieland Brendel.
\newblock Benchmarking robustness in object detection: Autonomous driving when winter is coming.
\newblock {\em Machine Learning for Autonomous Driving Workshop, NeurIPS}, 2019.

\bibitem{mintun2021interaction}
Eric Mintun, Alexander Kirillov, and Saining Xie.
\newblock On interaction between augmentations and corruptions in natural corruption robustness.
\newblock {\em NeurIPS}, 2021.

\bibitem{modas2022prime}
Apostolos Modas, Rahul Rade, Guillermo Ortiz-Jim{\'e}nez, Seyed-Mohsen Moosavi-Dezfooli, and Pascal Frossard.
\newblock Prime: A few primitives can boost robustness to common corruptions.
\newblock In {\em ECCV}, 2022.

\bibitem{rusak2020simple}
Evgenia Rusak, Lukas Schott, Roland~S Zimmermann, Julian Bitterwolf, Oliver Bringmann, Matthias Bethge, and Wieland Brendel.
\newblock A simple way to make neural networks robust against diverse image corruptions.
\newblock In {\em ECCV}, 2020.

\bibitem{dataaugmentationsurvey}
Connor Shorten and Taghi~M Khoshgoftaar.
\newblock A survey on image data augmentation for deep learning.
\newblock In {\em J Big Data 6, 60}, 2019.

\bibitem{upchurch2017deep}
Paul Upchurch, Jacob Gardner, Geoff Pleiss, Robert Pless, Noah Snavely, Kavita Bala, and Kilian Weinberger.
\newblock Deep feature interpolation for image content changes.
\newblock In {\em Proceedings of the IEEE conference on computer vision and pattern recognition}, pages 7064--7073, 2017.

\bibitem{wang2021augmax}
Haotao Wang, Chaowei Xiao, Jean Kossaifi, Zhiding Yu, Anima Anandkumar, and Zhangyang Wang.
\newblock Augmax: Adversarial composition of random augmentations for robust training.
\newblock {\em NeurIPS}, 2021.

\bibitem{wang2023internimage}
Wenhai Wang, Jifeng Dai, Zhe Chen, Zhenhang Huang, Zhiqi Li, Xizhou Zhu, Xiaowei Hu, Tong Lu, Lewei Lu, Hongsheng Li, et~al.
\newblock Internimage: Exploring large-scale vision foundation models with deformable convolutions.
\newblock In {\em Proceedings of the IEEE/CVF Conference on Computer Vision and Pattern Recognition}, pages 14408--14419, 2023.

\bibitem{wang2021pyramid}
Wenhai Wang, Enze Xie, Xiang Li, Deng-Ping Fan, Kaitao Song, Ding Liang, Tong Lu, Ping Luo, and Ling Shao.
\newblock Pyramid vision transformer: A versatile backbone for dense prediction without convolutions.
\newblock In {\em ICCV}, 2021.

\bibitem{wu2019wider}
Zifeng Wu, Chunhua Shen, and Anton Van Den~Hengel.
\newblock Wider or deeper: Revisiting the resnet model for visual recognition.
\newblock {\em Pattern Recognition}, 90:119--133, 2019.

\bibitem{xie2021segformer}
Enze Xie, Wenhai Wang, Zhiding Yu, Anima Anandkumar, Jose~M Alvarez, and Ping Luo.
\newblock Segformer: Simple and efficient design for semantic segmentation with transformers.
\newblock In {\em NeurIPS}, 2021.

\bibitem{yu2015multi}
Fisher Yu and Vladlen Koltun.
\newblock Multi-scale context aggregation by dilated convolutions.
\newblock {\em ICLR}, 2016.

\bibitem{yun2019cutmix}
Sangdoo Yun, Dongyoon Han, Seong~Joon Oh, Sanghyuk Chun, Junsuk Choe, and Youngjoon Yoo.
\newblock Cutmix: Regularization strategy to train strong classifiers with localizable features.
\newblock In {\em ICCV}, 2019.

\bibitem{zhang2017mixup}
Hongyi Zhang, Moustapha Cisse, Yann~N Dauphin, and David Lopez-Paz.
\newblock mixup: Beyond empirical risk minimization.
\newblock {\em ICLR}, 2018.

\bibitem{fan2}
Bingyin Zhao, Zhiding Yu, Shiyi Lan, Yutao Cheng, Anima Anandkumar, Yingjie Lao, and Jose~M Alvarez.
\newblock Fully attentional networks with self-emerging token labeling.
\newblock In {\em Proceedings of the IEEE/CVF International Conference on Computer Vision}, pages 5585--5595, 2023.

\bibitem{zhao2018icnet}
Hengshuang Zhao, Xiaojuan Qi, Xiaoyong Shen, Jianping Shi, and Jiaya Jia.
\newblock Icnet for real-time semantic segmentation on high-resolution images.
\newblock In {\em ECCV}, pages 405--420, 2018.

\bibitem{zhao2017pyramid}
Hengshuang Zhao, Jianping Shi, Xiaojuan Qi, Xiaogang Wang, and Jiaya Jia.
\newblock Pyramid scene parsing network.
\newblock In {\em CVPR}, pages 2881--2890, 2017.

\bibitem{zheng2021rethinking}
Sixiao Zheng, Jiachen Lu, Hengshuang Zhao, Xiatian Zhu, Zekun Luo, Yabiao Wang, Yanwei Fu, Jianfeng Feng, Tao Xiang, Philip~HS Torr, et~al.
\newblock Rethinking semantic segmentation from a sequence-to-sequence perspective with transformers.
\newblock In {\em CVPR}, 2021.

\bibitem{zheng2016improving}
Stephan Zheng, Yang Song, Thomas Leung, and Ian Goodfellow.
\newblock Improving the robustness of deep neural networks via stability training.
\newblock In {\em CVPR}, 2016.

\bibitem{zhou2017scene}
Bolei Zhou, Hang Zhao, Xavier Puig, Sanja Fidler, Adela Barriuso, and Antonio Torralba.
\newblock Scene parsing through ade20k dataset.
\newblock In {\em CVPR}, 2017.

\bibitem{zhou2022understanding}
Daquan Zhou, Zhiding Yu, Enze Xie, Chaowei Xiao, Animashree Anandkumar, Jiashi Feng, and Jose~M Alvarez.
\newblock Understanding the robustness in vision transformers.
\newblock In {\em ICLR}, 2022.

\end{thebibliography}
}
\clearpage



\appendix
\label{sec:appendix}

\section{Choice of $\epsilon$}\label{sec: choice of epsilon}
It is only important to select an $\epsilon$ value that incites the model to learn new representations i.e, which brings decreases the baseline model performance. Otherwise, if the model is already robust against the CWFA perturbation with a certain $\epsilon$, it will not be excited by CWFA to learn new features. We follow a simple idea and divide models into small ($<30$M parameters) and mid-larger models ($>30$M parameters). By looking at the sensitivity towards the CWFA perturbation of encoder 1, see Figure \ref{fig: encoder 1 sensitivity}, we decide to set $\epsilon=9$ for the B0 to B2 models. With this, the performance of these baseline models is decreased to $\approx 5\%$, $\approx 15\%$ and $\approx 50\%$ respectively. We proceed in a similar way for the larger B3 to B5 models and set $\epsilon=15$ which decreases its performance to $\approx 5\%$, $\approx 20 \%$, $\approx 20\%$ and $ \approx 30\%$ respectively. In overall, we choose $\epsilon$ to decrease the performance of the baseline models, but realize that its choice is not critical and do not do a hyperparameter search for it. 

\section{Complete tables on corrupted datasets}\label{sec: complete tables}
We provide the full tables, including the performance on each corruption type of our corrupted datasets. Table \ref{table: segformer on ade20k-c full table} shows the results on ADE20k-C for the baseline SegFormer and the SegFormer+CWFA models. Table \ref{table: segformer on cityscapes-c-bar full table} shows the results on Cityscapes-$\Bar{C}$ for the baseline SegFormer and SegFormer+CWFA models trained on Cityscapes. Finally, table \ref{table: segformer on cityscapes-c comparison augmix full table} shows the full results on City-C of the SegFormer architectures which are trained with AugMix and CWFA respectively. In addition, in Table \ref{fig:transfFigure} we report the complete results per corruption for the transferability experiment we conduct.

\section{Sensitivity analysis to the CWFA perturbation on all encoders}\label{sec: sensitivity analysis all encoders}
\begin{figure}[!t]
\begin{center}
    \centering
    \begin{subfigure}[b]{0.49\columnwidth}
        \centering
        \includegraphics[width=\textwidth]{CWFA_CVPR_FORMAT/figs/cwfa/encoder_1_figure_comparison_models.png}
        \caption{Encoder 1}
        \label{fig: encoder 1 sensitivity}
    \end{subfigure}    
    \begin{subfigure}[b]{0.49\columnwidth}
        \centering
        \includegraphics[width=\textwidth]{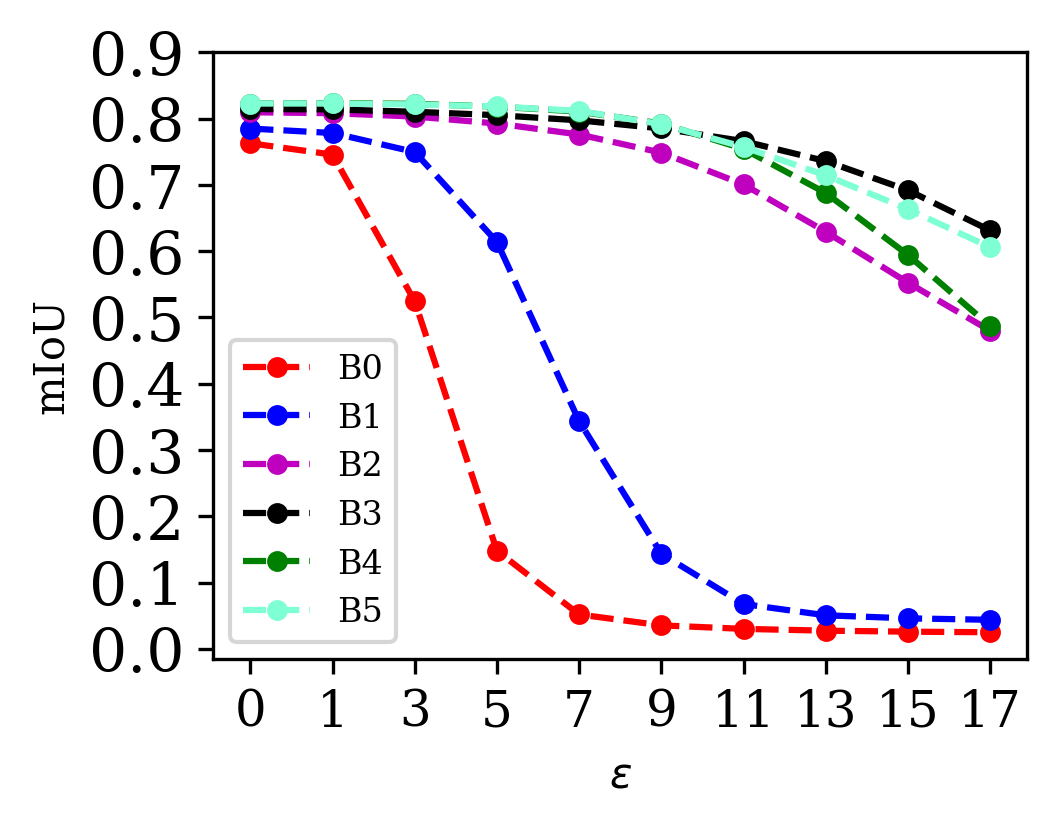}
        \caption{Encoder 2}
        \label{}
    \end{subfigure}  
    \begin{subfigure}[b]{0.49\columnwidth}
        \centering
        \includegraphics[width=\textwidth]{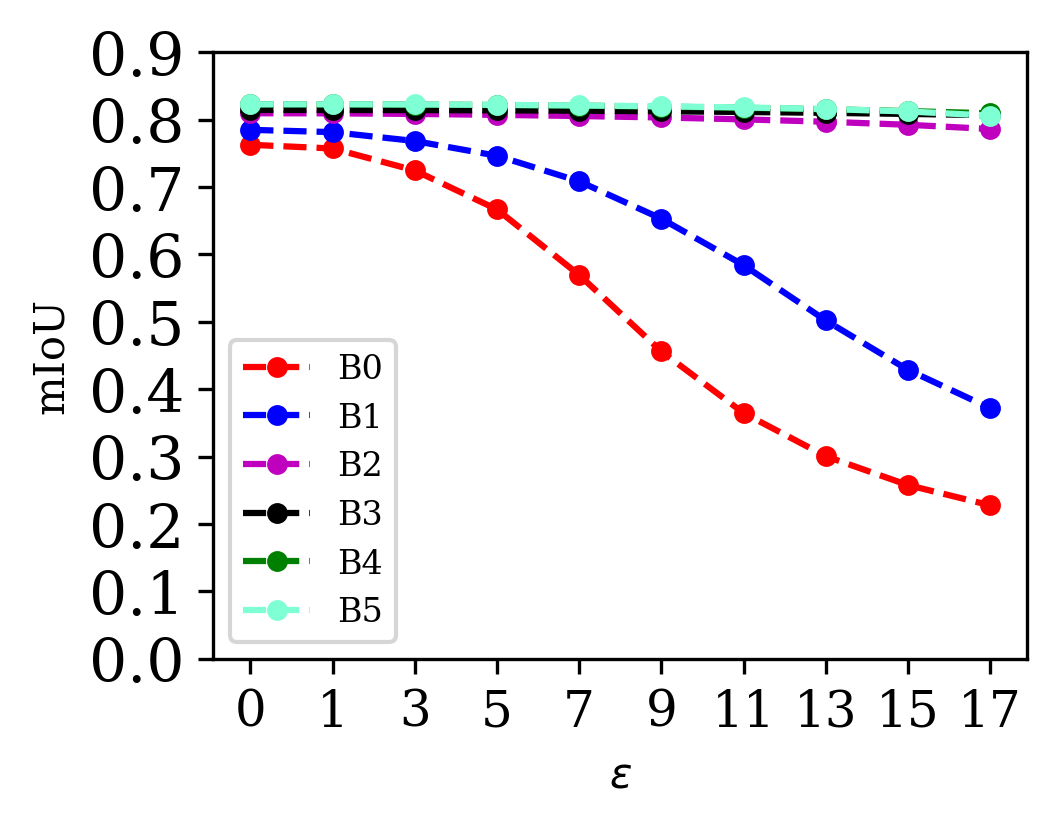}
        \caption{Encoder 3}
        \label{}
    \end{subfigure}  
    \begin{subfigure}[b]{0.49\columnwidth}
        \centering
        \includegraphics[width=\textwidth]{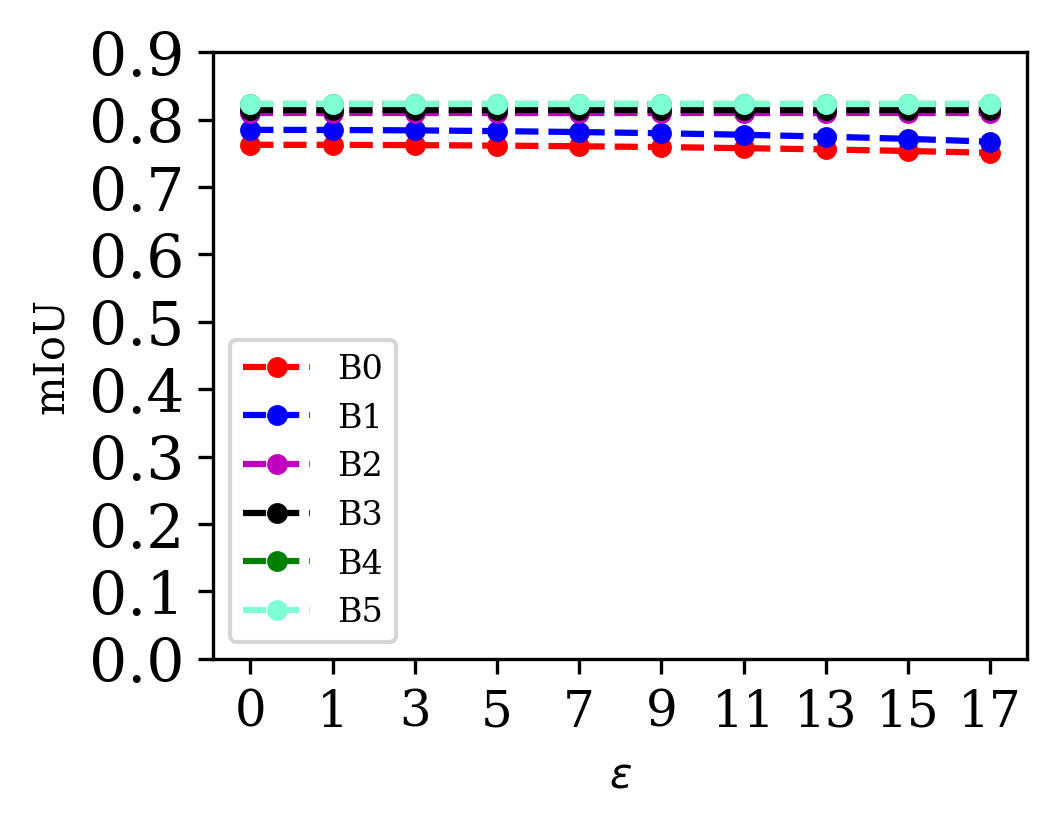}
        \caption{Encoder 4}
        \label{}
    \end{subfigure}  
    \vspace{-2mm}
    \caption{\textbf{Sensitivity of baseline SegFormer baseline models towards CWFA}}
    \label{fig:CWFAsensitivities all encoders}
\end{center}
\end{figure}

\begin{figure}[!t]
\begin{center}
    \centering
    \begin{subfigure}[b]{0.49\columnwidth}
        \centering
        \includegraphics[width=\textwidth]{CWFA_CVPR_FORMAT/figs/cwfa/segformer_b0_encoders.png}
        \caption{SegFormer B0}
        \label{}
    \end{subfigure}    
    \begin{subfigure}[b]{0.49\columnwidth}
        \centering
        \includegraphics[width=\textwidth]{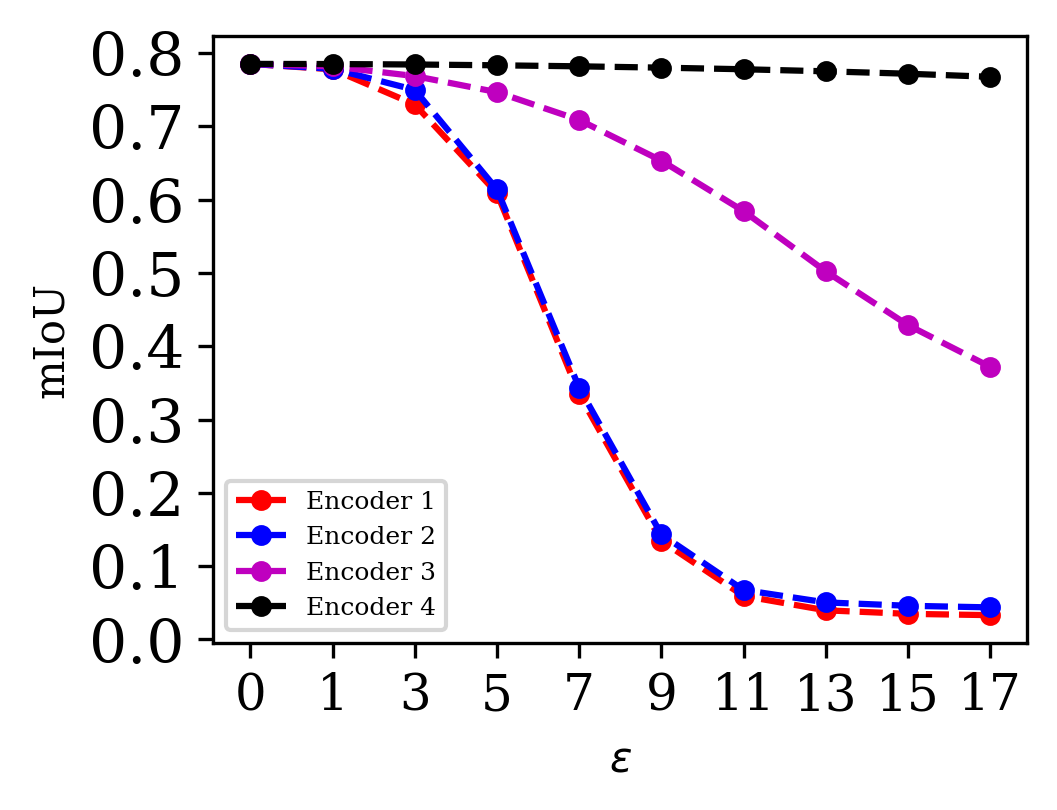}
        \caption{SegFormer B1}
        \label{}
    \end{subfigure}  
    \begin{subfigure}[b]{0.49\columnwidth}
        \centering
        \includegraphics[width=\textwidth]{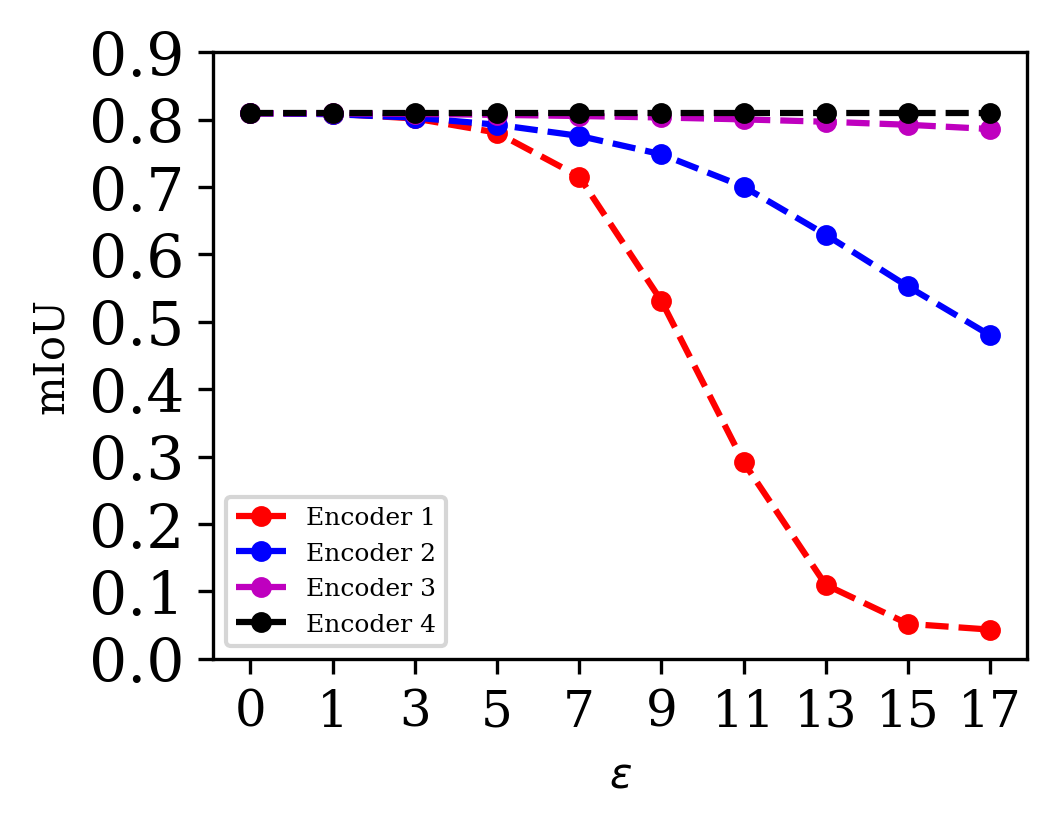}
        \caption{SegFormer B2}
        \label{}
    \end{subfigure}  
    \begin{subfigure}[b]{0.49\columnwidth}
        \centering
        \includegraphics[width=\textwidth]{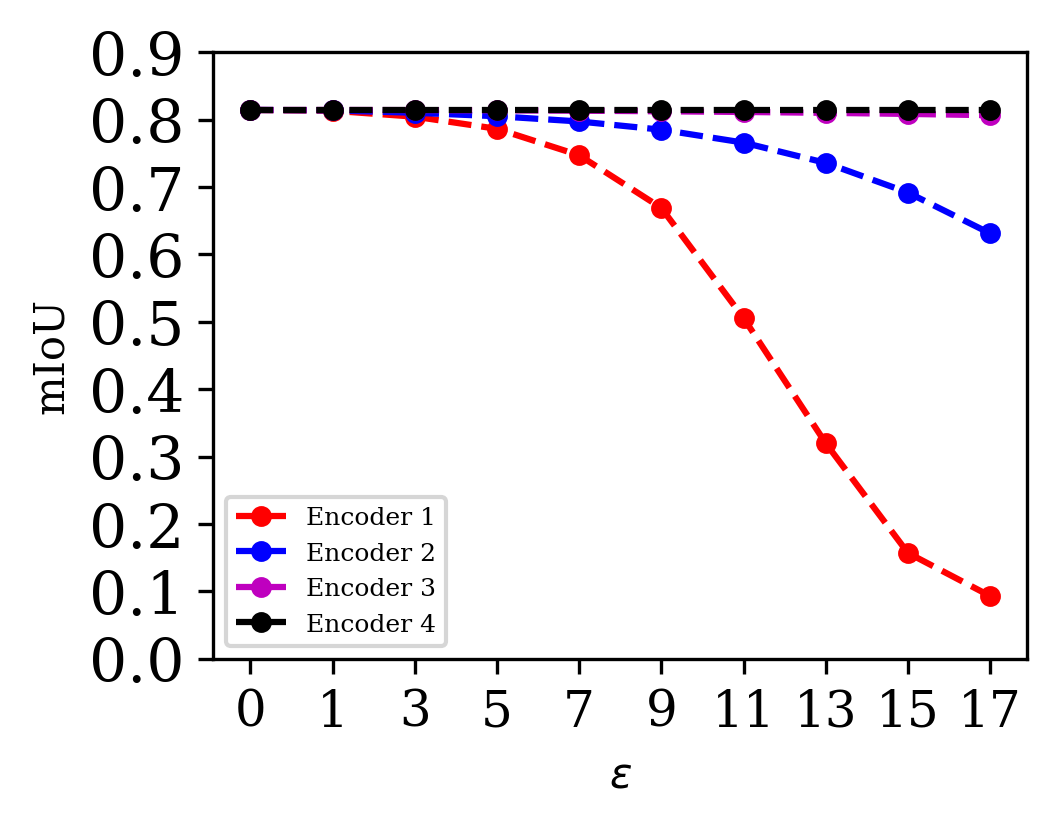}
        \caption{SegFormer B3}
        \label{}
    \end{subfigure}  
    \begin{subfigure}[b]{0.49\columnwidth}
        \centering
        \includegraphics[width=\textwidth]{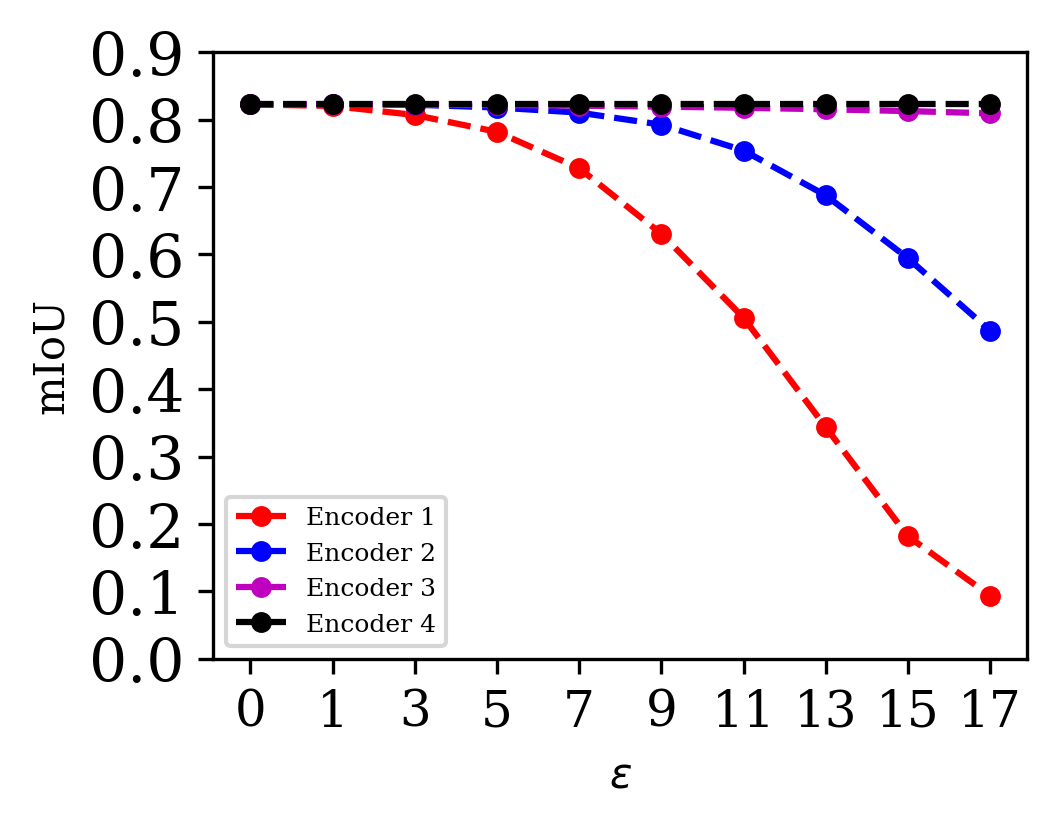}
        \caption{SegFormer B4}
        \label{}
    \end{subfigure}  
    \begin{subfigure}[b]{0.49\columnwidth}
        \centering
        \includegraphics[width=\textwidth]{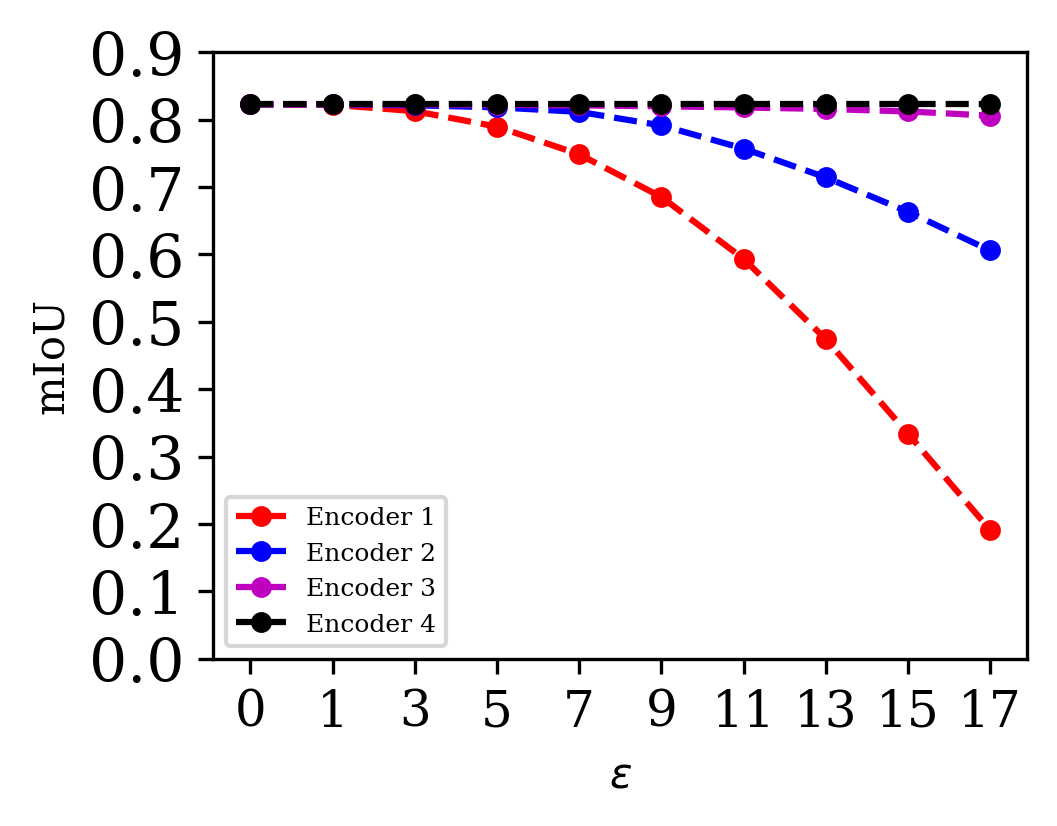}
        \caption{SegFormer B5}
        \label{}
    \end{subfigure}  
    \vspace{-2mm}
    \caption{\textbf{Sensitivity of baseline SegFormer baseline models towards CWFA at per model for all encoders}}
    \label{fig:CWFAsensitivities all models}
    
\end{center}
\end{figure}

Figure \ref{fig:CWFAsensitivities all encoders} shows the SegFormer baseline models sensitivities against the CWFA perturbation at each encoder respectively. We observe that when introducing CWFA at encoders one to three respectively, the performance drop can be significant. We recognize that the most sensible encoders are encoder one and two, while encoder four is merely affected by the perturbation. As pointed out in the main paper, the larger the model is, the less sensible the encoders are. Figure \ref{fig:CWFAsensitivities all models} makes this more evident, by comparing the sensitivities per model at each encoder. 

\section{Visual comparison baseline, AugMix and CWFA predictions}\label{sec: visual comparison images augmix cwfa baseline}
\begin{figure*}[!t]
\begin{center}
    \centering
    \begin{subfigure}[b]{0.6\columnwidth}
        \centering
        \includegraphics[width=\textwidth]{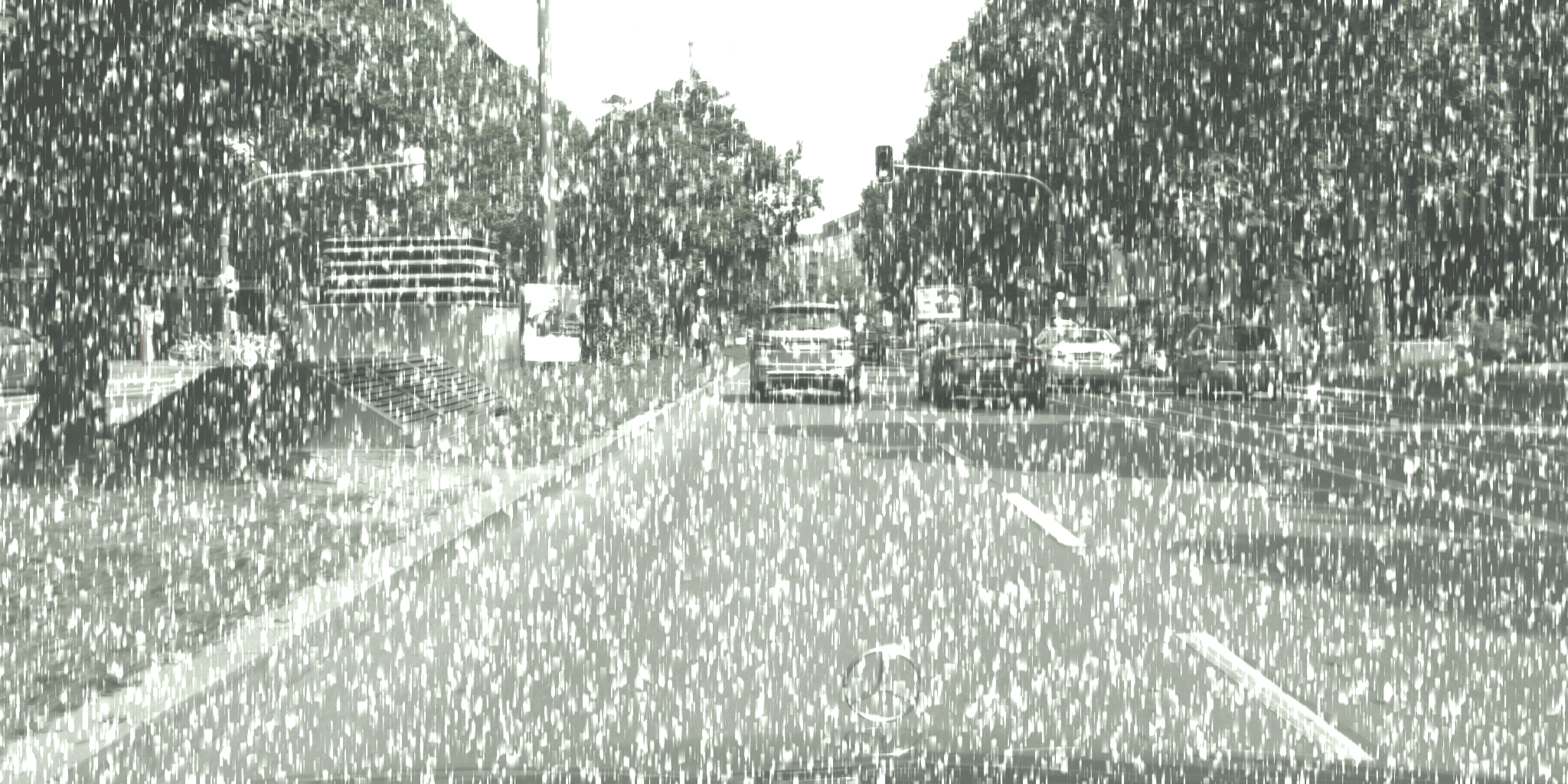}
        \caption{Snow sev. 4}
        \label{}
    \end{subfigure} 
    \begin{subfigure}[b]{0.6\columnwidth}
        \centering
        \includegraphics[width=\textwidth]{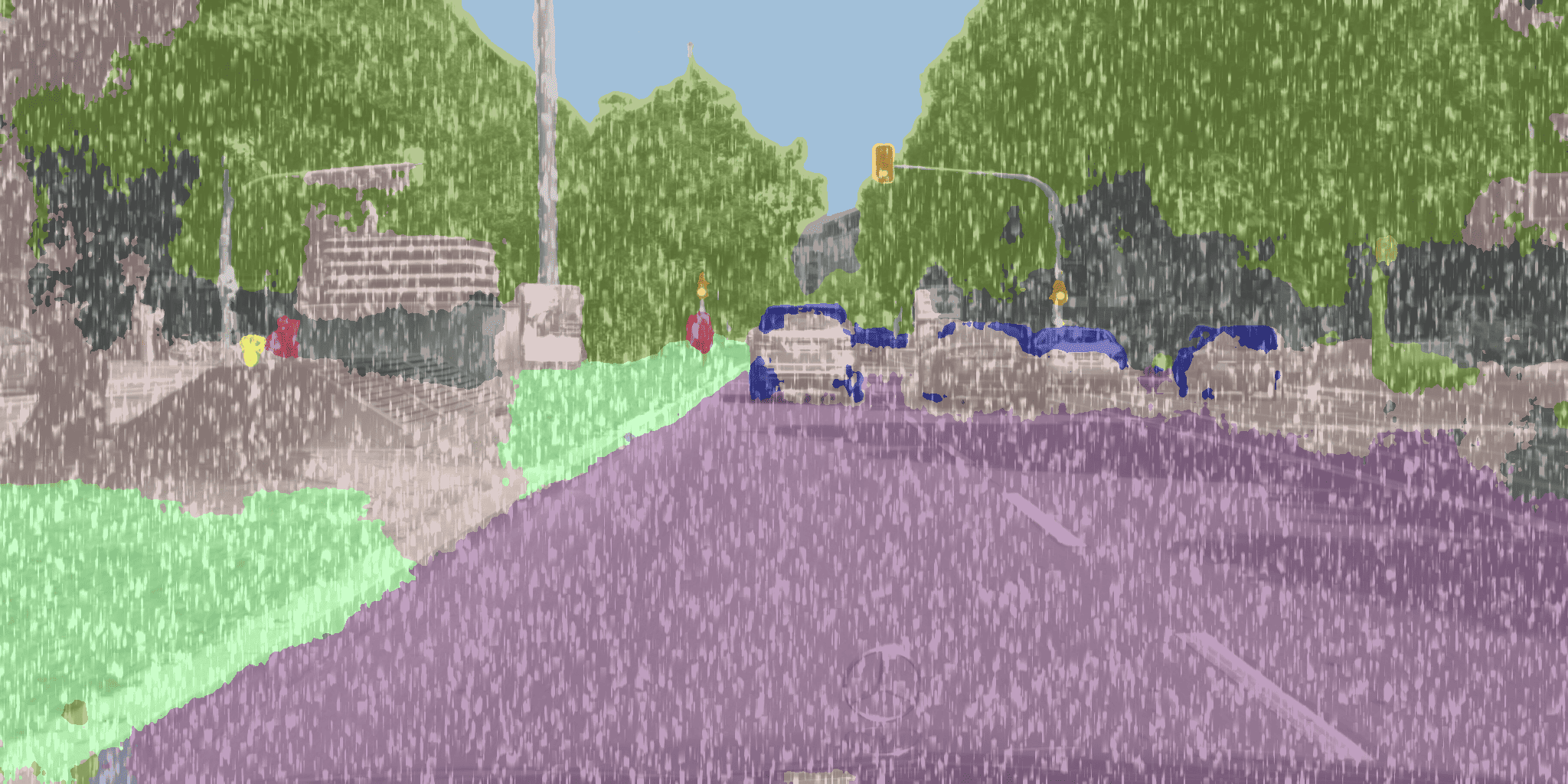}
        \caption{}
        \label{}
    \end{subfigure}    
    \begin{subfigure}[b]{0.6\columnwidth}
        \centering
        \includegraphics[width=\textwidth]{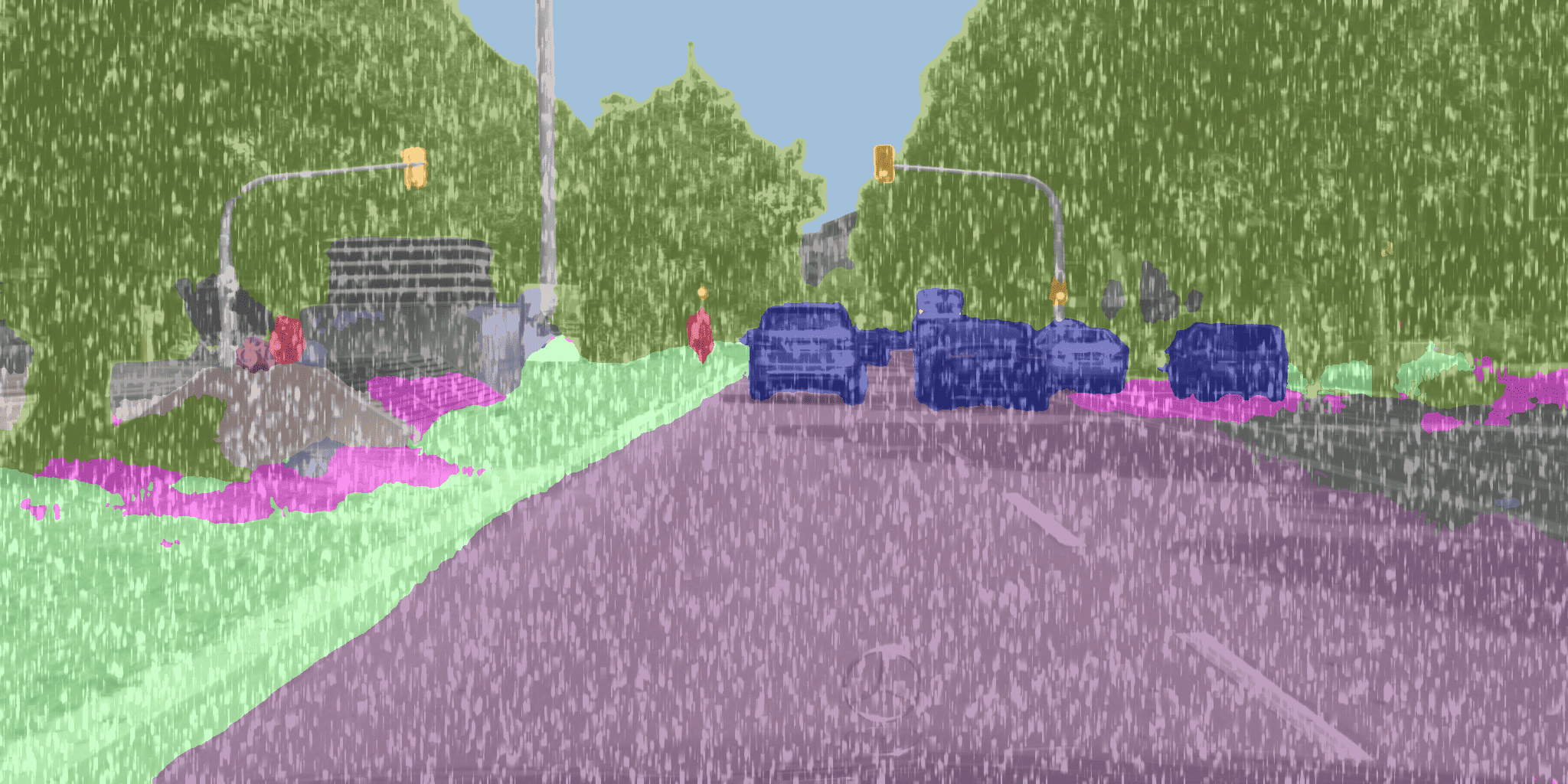}
        \caption{}
        \label{}
    \end{subfigure}  
    \begin{subfigure}[b]{0.6\columnwidth}
        \centering
        \includegraphics[width=\textwidth]{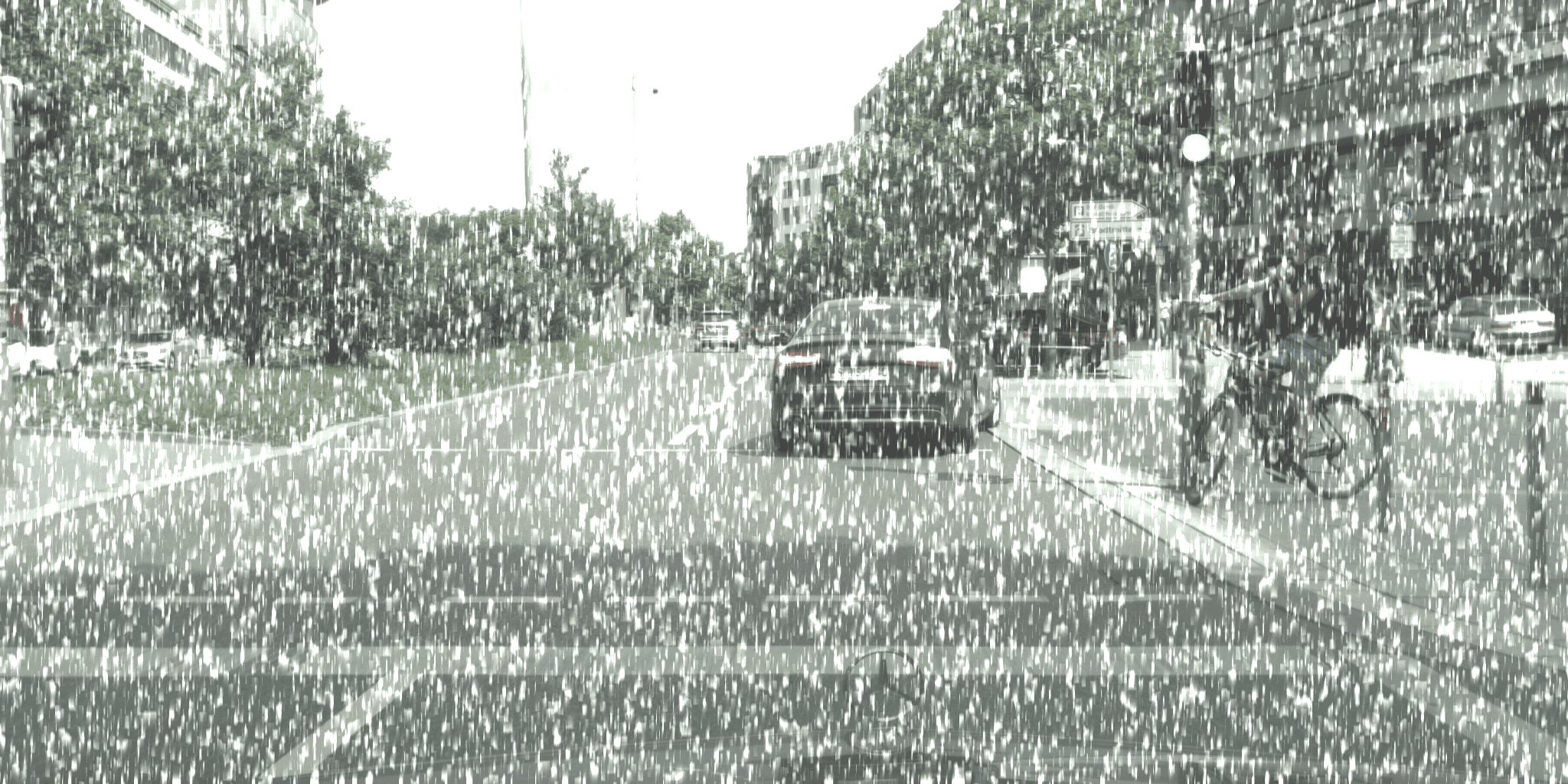}
        \caption{Snow sev. 4}
        \label{}
    \end{subfigure} 
    \begin{subfigure}[b]{0.6\columnwidth}
        \centering
        \includegraphics[width=\textwidth]{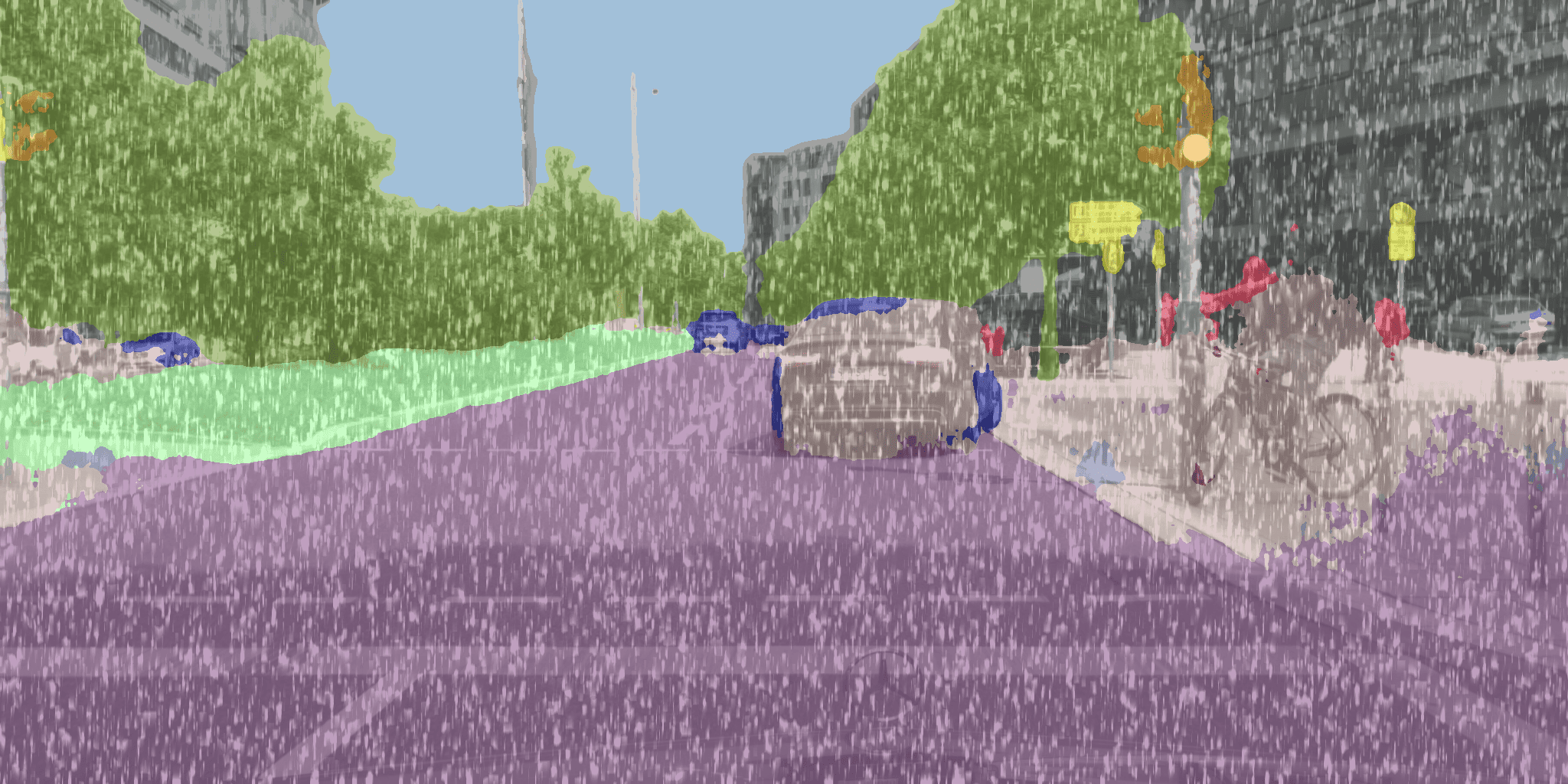}
        \caption{}
        \label{}
    \end{subfigure}  
    \begin{subfigure}[b]{0.6\columnwidth}
        \centering
        \includegraphics[width=\textwidth]{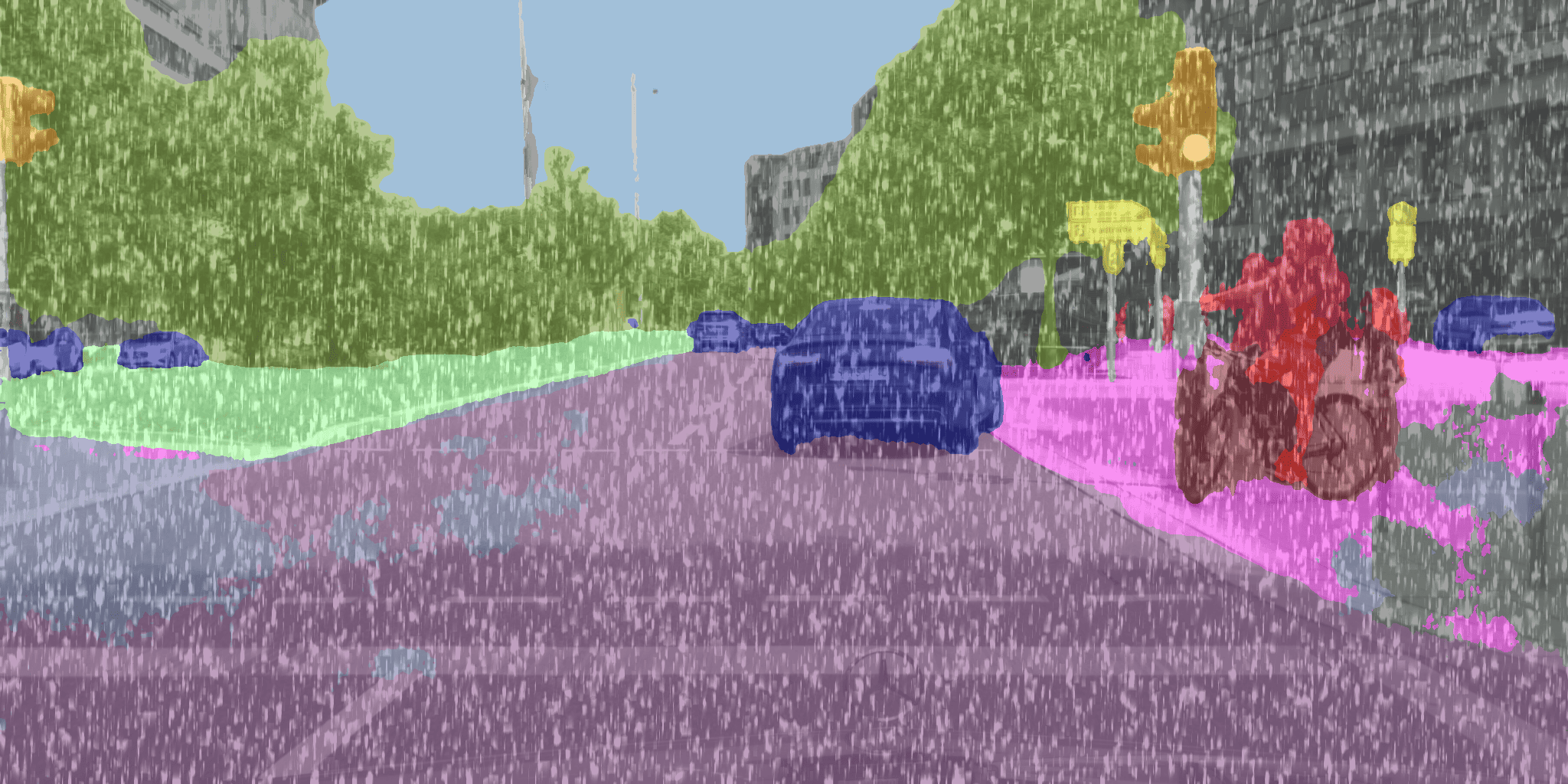}
        \caption{}
        \label{}
    \end{subfigure} 

    \begin{subfigure}[b]{0.6\columnwidth}
        \centering
        \includegraphics[width=\textwidth]{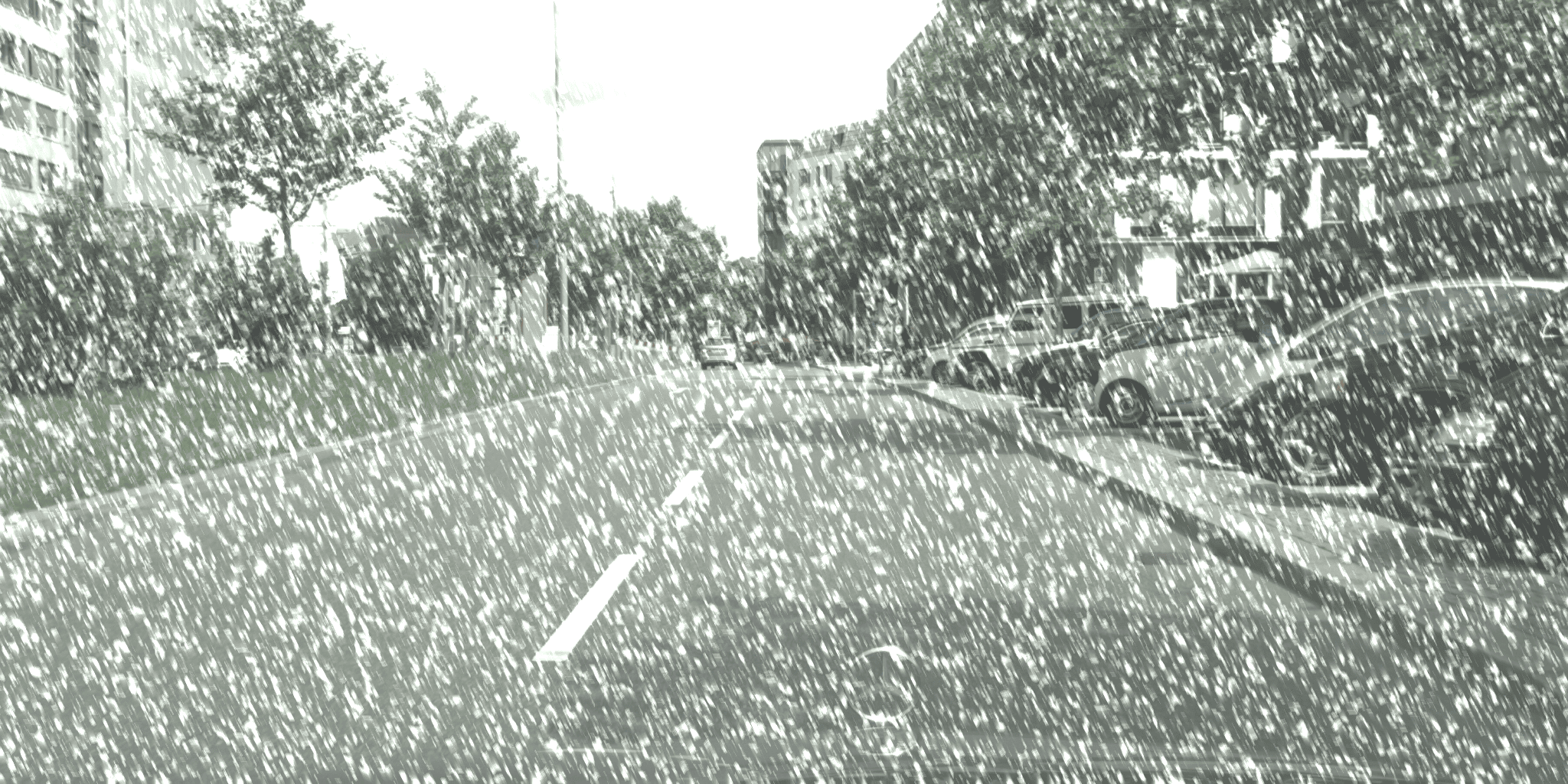}
        \caption{Snow sev. 4}
        \label{}
    \end{subfigure} 
    \begin{subfigure}[b]{0.6\columnwidth}
        \centering
        \includegraphics[width=\textwidth]{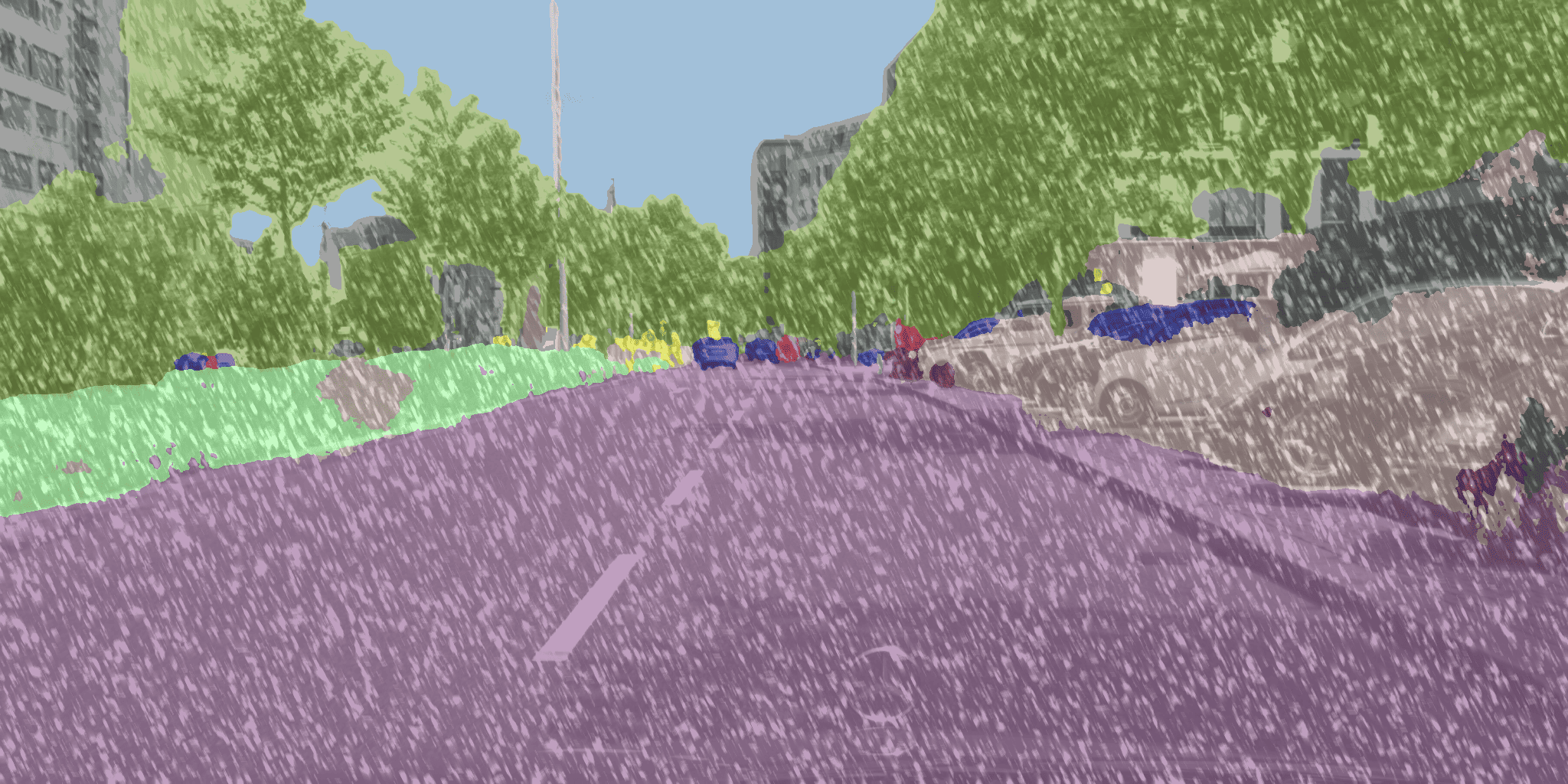}
        \caption{}
        \label{}
    \end{subfigure}  
    \begin{subfigure}[b]{0.6\columnwidth}
        \centering
        \includegraphics[width=\textwidth]{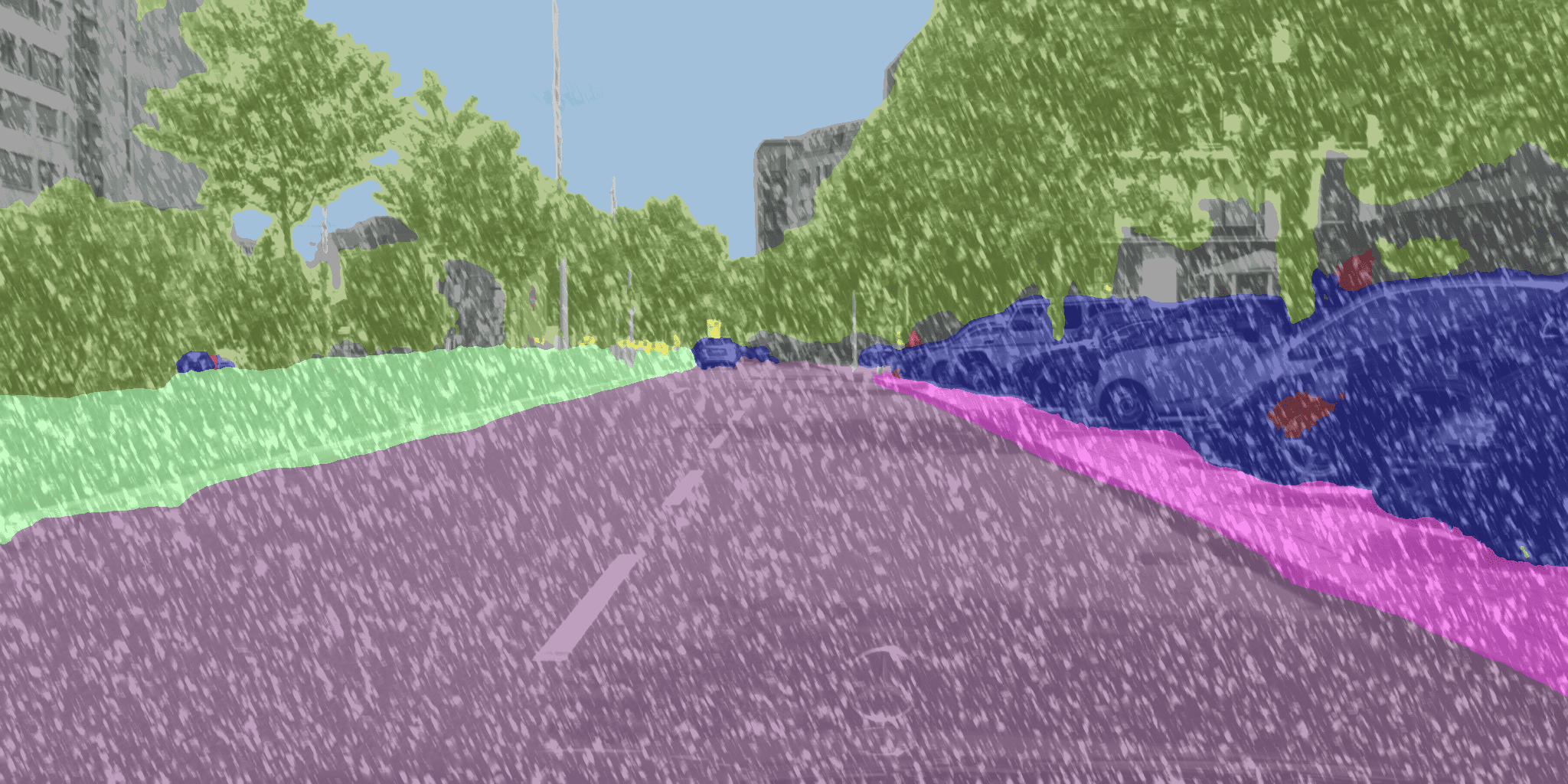}
        \caption{}
        \label{}
    \end{subfigure} 

     \begin{subfigure}[b]{0.6\columnwidth}
        \centering
        \includegraphics[width=\textwidth]{CWFA_CVPR_FORMAT/figs/predictions/b5_comparison/snow_5_example_1.png}
        \caption{Snow sev. 5}
        \label{}
    \end{subfigure} 
    \begin{subfigure}[b]{0.6\columnwidth}
        \centering
        \includegraphics[width=\textwidth]{CWFA_CVPR_FORMAT/figs/predictions/b5_comparison/snow_05_b5_baseline_example_1.png}
        \caption{}
        \label{}
    \end{subfigure}  
    \begin{subfigure}[b]{0.6\columnwidth}
        \centering
        \includegraphics[width=\textwidth]{CWFA_CVPR_FORMAT/figs/predictions/b5_comparison/snow_05_b5_cwfa_example_1.png}
        \caption{}
        \label{}
    \end{subfigure} 

    \begin{subfigure}[b]{0.6\columnwidth}
        \centering
        \includegraphics[width=\textwidth]{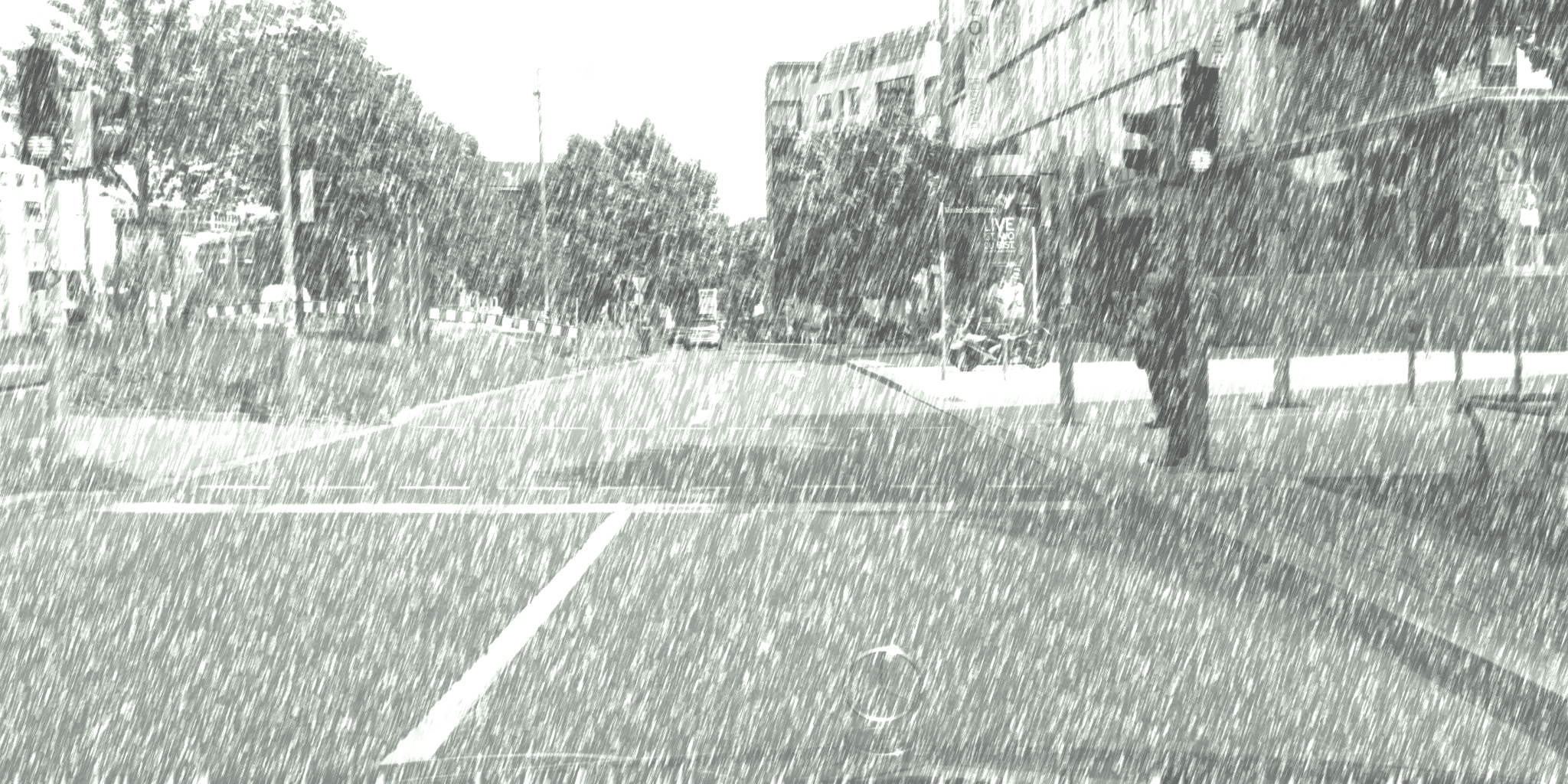}
        \caption{Snow sev. 5}
        \label{}
    \end{subfigure} 
    \begin{subfigure}[b]{0.6\columnwidth}
        \centering
        \includegraphics[width=\textwidth]{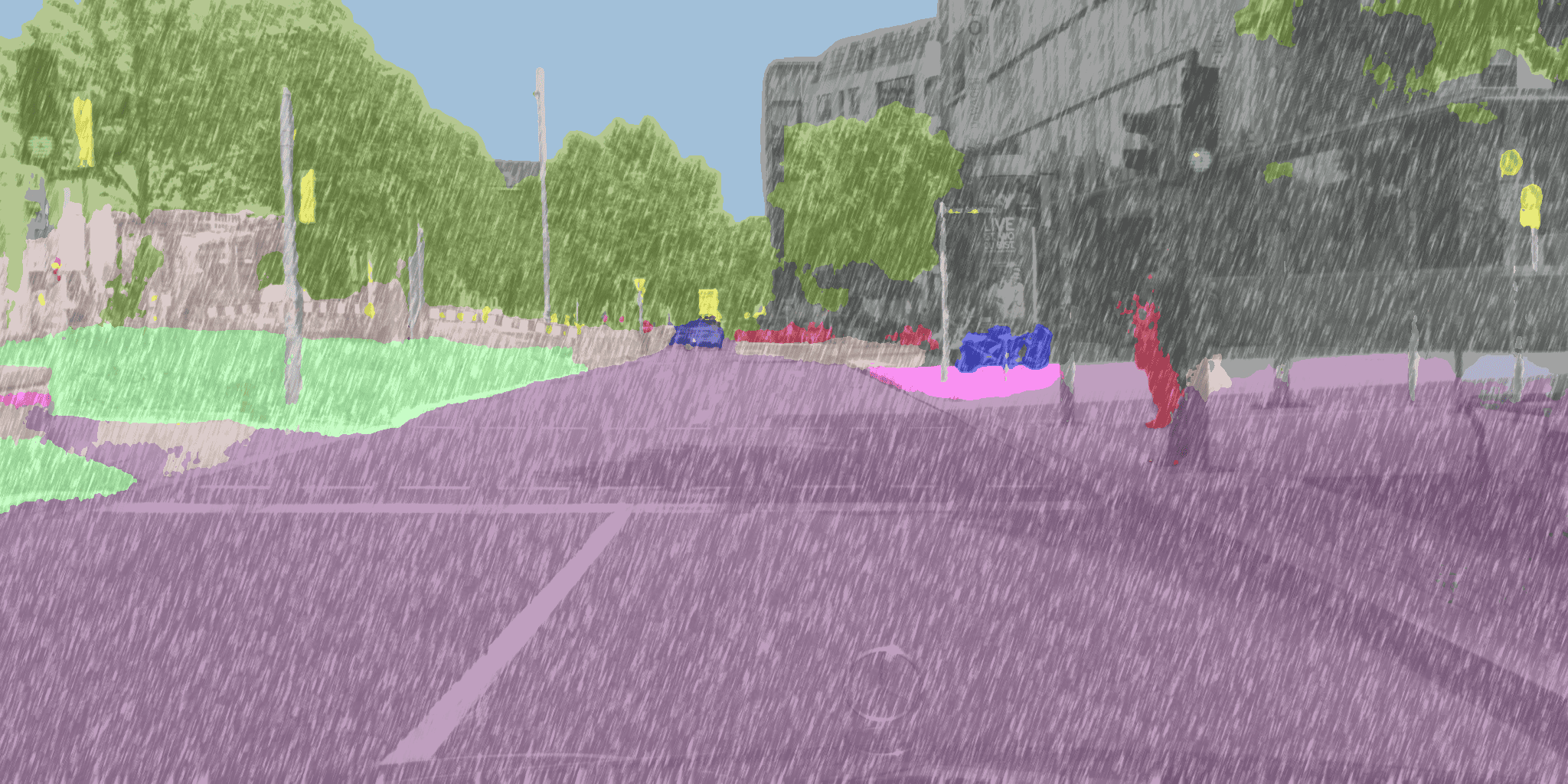}
        \caption{}
        \label{}
    \end{subfigure}  
    \begin{subfigure}[b]{0.6\columnwidth}
        \centering
        \includegraphics[width=\textwidth]{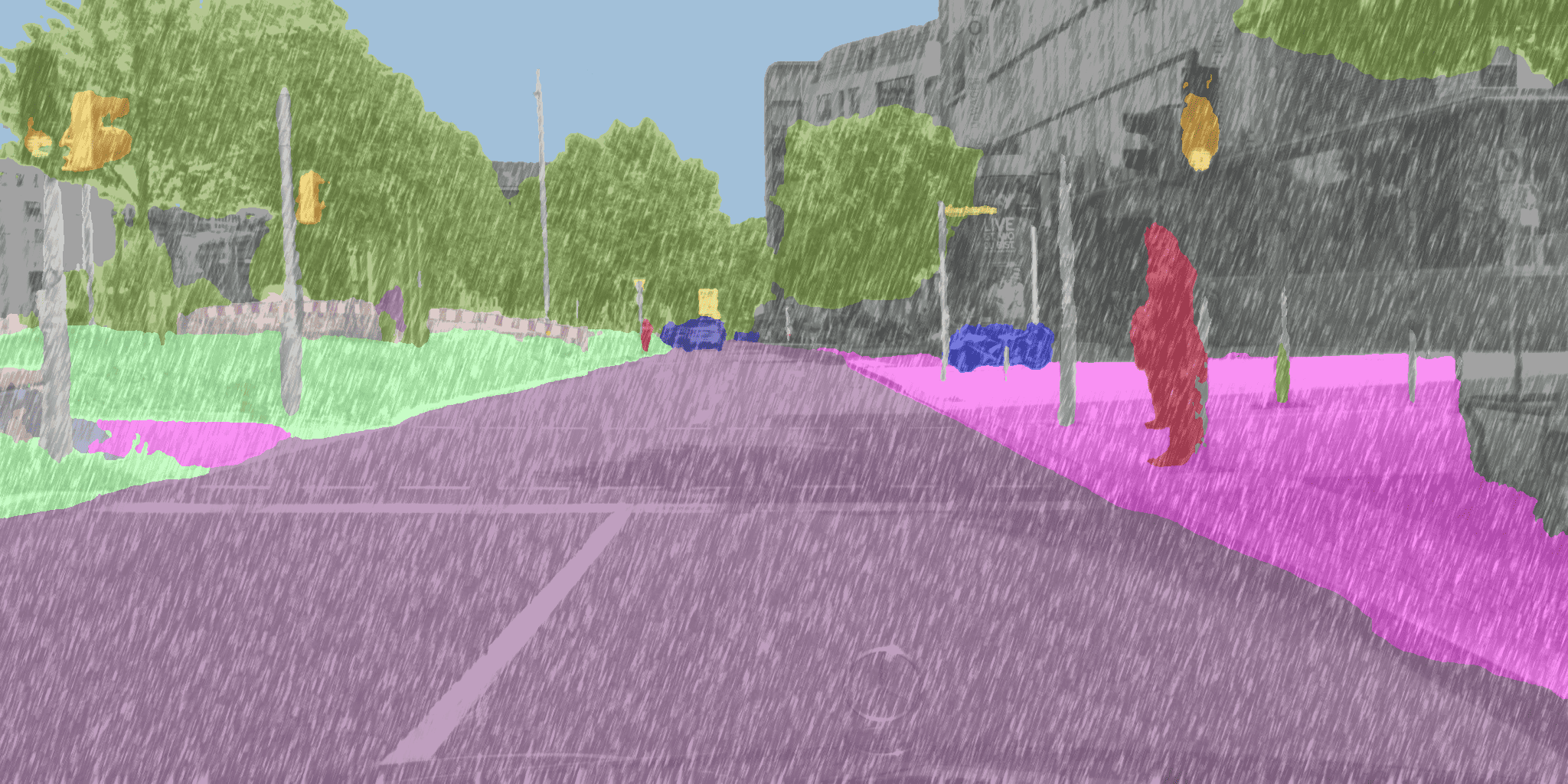}
        \caption{}
        \label{}
    \end{subfigure} 
    \vspace{-2mm}
    \caption{\textbf{B5 and B5+CWFA model comparison from left to right:} input corrupted image, B5 baseline and B5+CWFA prediction}
    \label{fig: b5  baseline and cwfa predicitons}
    
\end{center}
\end{figure*}

We provide several visual comparison between our CWFA trained SegFormer and the AugMix and baseline models. For visualisation, we use the available Cityscapes \cite{cordts2016cityscapes} sequences which do not have semantic annotations. Figure \ref{fig: b5  baseline and cwfa predicitons}  compares the baseline SegFormer B5 to the CWFA trained B5 model. We run the model inference with the same settings as the ones described in the implementation details of the paper in section~4.1. 

\section{Implementation details}\label{sec: implementation details complete}
We follow \cite{xie2021segformer} and use the same training procedure. For our main experiments and for AugMix, we train the B0 to B5 architectures for 160k iterations and initialize the models with the publicly available ImageNet-1k pre-trained weights. We use the AdamW optimizer and set the learning rate $lr=0.0002$, the weight decay to $0.0001$. We use a “poly” LR schedule with power $0.9$. For Cityscapes, we use crop size of $1024 \times 1024$ and $512 \times 512$ for ADE20K. We use standard augmentations as random cropping and flipping with probability $0.5$.

We explore the effectiveness of applying CWFA at different stages of training for B0 and B1 models. Specifically, we fine-tune pretrained B0 and B1 models with CWFA for 16k iterations, train B0 and B1 SegFormer models from scratch with CWFA for 160k iterations, and train B0 and B1 models from scratch with CWFA from iteration 16k on. Our results demonstrate that the last option leads to the best performance, with the other two options resulting in inferior results. Interestingly, when applying CWFA from the first iteration during B1 training, we observed a substantial $111.1\%$ robustness degradation, likely due to early feature perturbation hindering the model's ability to learn meaningful representations. These findings highlight the importance of allowing the model to first learn meaningful representations before applying CWFA. For further details and results, please refer to Table $\ref{tab:cwfa 160k training sup}$.

\begin{table}[!t]
\caption{\textbf{Comparison fine-tuning 16k to training 160k iterations with CWFA}: Report the mIoU on Cityscapes validation set, the average mIoU and retention rate on Cityscapes-C, * denotes the models where CWFA is applied after the 16k iteration of 160k}
\centering
\scalebox{0.65}{
\begin{tabular}{c|cc|cc|ccc}
\toprule
& & & & & \multicolumn{3}{c}{Cityscapes} \\ 
Model & Training & CWFA & Fine-Tuning & CWFA & Clean & Corrupt & Reten. \\
\midrule
SegFormer-B0 & 160k &  \ding{55} &  16k & \checkmark & 75.6 & 56.3 & 74.5\\
SegFormer-B0 & 160k &  \checkmark &  \ding{55} & \ding{55} & \textbf{76.4} & 55.9 & 73.2 \\
SegFormer-B0* & 160k & \checkmark &  \ding{55} & \ding{55} & \textbf{76.4} & \textbf{57.4} & \textbf{75.1}\\
\hline
\hline
SegFormer-B1 & 160k &  \ding{55} &  16k & \checkmark &   77.2 & 59.2 & 76.7 \\
SegFormer-B1 & 160k &  \checkmark &  \ding{55} & \ding{55} & 64.4 & 45.4 & 70.0\\
SegFormer-B1* & 160k &  \checkmark &  \ding{55} & \ding{55} & \textbf{78.2} & \textbf{61.7} & \textbf{78.9}\\

\bottomrule
\end{tabular}}
\label{tab:cwfa 160k training sup}
\end{table}

\begin{table*}[!t]
\caption{\textbf{Segformer on Cityscapes comparison CWFA with AugMix.} Our results compared to the AugMix Segformer models. Our approach outpeforms AugMix }
\label{table: segformer on cityscapes-c comparison augmix full table}
\scalebox{0.6}{
\begin{tabular}{c|c|c|c|c|c|c|c|c|c|c|c|c|c|c|c|c|c|c|c|c|} 
   \multicolumn{1}{c|}{Model /}& \multirow{2}{*}{Method} & City &\multicolumn{4}{|c|}{Blur}  & \multicolumn{4}{|c|}{Noise} & \multicolumn{4}{|c|}{Digital} & \multicolumn{4}{|c|}{Weather} & \multirow{2}{*}{Average} & \multirow{2}{*}{Reten.}  \\
  \cline{4-19}
    \multicolumn{1}{c|}{Params. [M.]} & &(Clean) & Motion & Defoc & Glass & Gauss & Gauss & Impul & Shot & Speck & Bright & Contr & Satur & JPEG & Snow & Spatt & Fog & Frost & & \\
     \hline
    \multirow{2}{*}{B0 / 3.4}  & AugMix & 75.7 & 61.3 & 58.9 & 50.1 & 58.9 & 29.7 & 33.3 & 35.4 & 52.8 & 73.0 & 69.2 & 71.7 & 44.2 & 31.4 & 59.7 & 66.1 & 39.5 & 52.2 & 69.0\\
   
 & CWFA (Ours) & \textbf{76.4} & \textbf{64.3} & \textbf{62.8} & \textbf{59.0} & \textbf{63.3} & \textbf{43.1} & \textbf{43.9} & \textbf{49.9} & \textbf{62.4} & 72.9 & 66.6 & 71.7 & \textbf{47.3} & \textbf{41.6} & \textbf{63.0} & 64.5 & \textbf{42.6} & \textbf{57.4} & \textbf{75.1} \\
 \hline
    \multirow{2}{*}{B1 / 13.1}  & AugMix & 77.6 & 64.5 & 47.1 & 51.0 & 47.6 & 47.1 & 48.2 & 52.3 & 63.9 & 76.0 & 73.2 & 75.0 & 45.8 & 37.5 & 67.6 & 72.0 & 49.4 & 57.4 & 74.0 \\

& CWFA (Ours) & \textbf{78.2} & \textbf{66.9} & \textbf{65.4} & \textbf{60.9} & \textbf{65.7} & \textbf{50.1} & \textbf{51.4} & \textbf{56.0} & \textbf{67.1} & \textbf{76.1} & 70.6 & 74.8 & \textbf{47.0} & \textbf{49.3} & \textbf{68.2} & 70.0 & 47.2 & \textbf{61.7} & \textbf{78.9}\\
  \hline
    \multirow{2}{*}{B2 / 24.2}  & AugMix & 81.0 & 69.9 & 56.2 & 53.7 & 55.7 & 56.2 & 54.0 & 60.7 & 70.0 & 80.3 & 78.4 & 78.8 & 54.3 & 52.9 & 74.5 & 74.7 & 59.2 & 64.4 & 79.4 \\

&  CWFA (Ours)  & \textbf{81.2} & \textbf{70.5} & \textbf{69.7} & \textbf{66.3} & \textbf{70.4} & \textbf{57.7} & \textbf{58.0} & \textbf{63.1} & \textbf{71.9} & 79.8 & 76.1 & 78.4 & 50.4 & \textbf{55.3} & 72.1 & \textbf{75.7} & 51.1 & \textbf{66.7} & \textbf{82.1} \\
\hline
  \multirow{2}{*}{B3 / 44.0}  & AugMix  & 80.8 & 70.6 & 63.2 & 60.2 & 61.1 & 59.0 & 57.2 & 64.3 & 72.7 & 80.2 & 78.6 & 79.0 & 62.5 & 56.8 & 75.0 & 72.4 & 60.5 & 67.1 & 83.1 \\ 

&  CWFA (Ours) &\textbf{81.5} & \textbf{70.9} & \textbf{70.5} & \textbf{66.9} & \textbf{71.3} & \textbf{62.5} & \textbf{63.0} & \textbf{67.1} & \textbf{74.0} & 80.2 & 77.0 & \textbf{79.1} & 52.2 & 54.5 & 73.6 & \textbf{76.9} & 51.7 & \textbf{68.2} & \textbf{83.7}  \\
\hline
  \multirow{2}{*}{B4 / 60.8}  & AugMix & 81.7 & 68.9 & 63.1 & 57.0 & 62.1 & 64.8 & 64.6 & 69.0 & 75.1 & 80.9 & 78.8 & 80.0 & 64.2 & 58.3 & 76.1 & 74.9 & 64.0 & 68.9 & 84.3 \\

& CWFA (Ours) & \textbf{82.1} & \textbf{71.0} & \textbf{70.8} & \textbf{68.0} & \textbf{71.7} & 61.8 & 64.0 & 67.3 & 74.7 & 80.9 & 77.5 & 79.7 & 53.9 & 58.2 & 72.7 & \textbf{77.9} & 54.5 & \textbf{69.0} & 84.0  \\
\hline
  \multirow{2}{*}{B5 / 81.4}  & AugMix & 82.1 & 70.2 & 64.7 & 61.0 & 63.4 & 60.3 & 62.7 & 66.7 & 74.4 & 81.3 & 79.6 & 80.4 & 64.4 & 60.2 & 76.5 & 73.1 & 65.7 & 69.0 & 84.1\\

&  CWFA (Ours)  & \textbf{82.2} & 70.1 & \textbf{69.9} & \textbf{66.5} & \textbf{71.2} & \textbf{64.5} & \textbf{66.6} & \textbf{68.6} & \textbf{74.7} & 80.8 & 77.1 & 79.9 & 54.7 & 58.2 & 73.6 & \textbf{78.1} & 53.6 & \textbf{69.3} & \textbf{84.3}\\
\end{tabular}}
\end{table*}

\begin{table*}[!t]
\caption{\textbf{Segformer on City-$\Bar{\text{C}}$}  Our results compared to the Segformer baseline. Our results show that CWFA improves robustness against other corruptions than the City-C corruptions.}
\label{table: segformer on cityscapes-c-bar full table}
\scalebox{0.65}{
\begin{tabular}{c|c|c|c|c|c|c|c|c|c|c|c|c|c|} 
   \multicolumn{1}{c|}{Model /}& \multirow{2}{*}{Method} & City &\multicolumn{5}{|c|}{Noise}  & Inverse & & Checkerboard & Cocentric &  \multirow{2}{*}{Average} &  \multirow{2}{*}{Reten.}  \\
  
    \multicolumn{1}{c|}{Params. [M.]} & &(Clean) & Single frequency & Brown & Plasma & Perlin & Blue Sample & Sparkles & Sparkles & Cutout & Sine WAVES & &  \\
     \hline
    \multirow{2}{*}{B0 / 3.7}  & Segformer & 76.3 & 9.8 & 54.7 & 25.1 & 67.6 & 40.7 & 31.9 & 64.1 & 59.0 & 35.5 & 43.1 & 56.6  \\

 &  CWFA (Ours) & \textbf{76.4} & \textbf{30.2} & \textbf{58.3} & \textbf{31.4} & \textbf{69.3} & \textbf{56.0} & \textbf{37.5} & \textbf{65.0} & 59.0 & \textbf{39.2} & \textbf{49.5} & \textbf{64.8} \\
  \cline{1-14}
    \multirow{2}{*}{B1 / 13.6}  & Segformer & 78.5 & 12.1 & 59.4 & 30.0 & 71.3 & 45.5 & 35.9 & 67.2 & 61.0 & 39.7 & 46.9 & 59.8 \\

& CWFA (Ours) &  78.2 & \textbf{34.1} & \textbf{62.9} & \textbf{36.0} & \textbf{72.5} & \textbf{59.9} & \textbf{41.2} & \textbf{68.1} & \textbf{61.9} & \textbf{44.0} & \textbf{53.4} & \textbf{68.3}\\
\cline{1-14}
    \multirow{2}{*}{B2 / 27.3}  & Segformer &  81.0 & 16.5 & 68.1 & 37.0 & 75.8 & 49.8 & 42.2 & 71.5 & 64.2 & 49.9 & 52.8 & 65.2 \\

&  CWFA (Ours)  &  \textbf{81.2} & \textbf{49.7} & \textbf{70.9} & \textbf{43.1} & \textbf{76.2} & \textbf{64.2} & \textbf{47.2} & 71.4 & \textbf{64.8} & \textbf{53.0} & \textbf{60.1} & \textbf{74.0}  \\
\cline{1-14}
  \multirow{2}{*}{B3 / 47.2}  & Segformer  & 81.6 & 25.4 & 72.0 & 42.5 & 77.0 & 52.6 & 44.5 & 72.6 & 66.4 & 51.6 & 56.1 & 68.7 \\ 

&  CWFA (Ours) &  81.5 & \textbf{55.4} & \textbf{72.9} & \textbf{48.9} & \textbf{77.3} & \textbf{63.2} & \textbf{46.1} & \textbf{72.7} & 66.1 & \textbf{53.9} & \textbf{61.8} & \textbf{75.9} \\
\cline{1-14}
  \multirow{2}{*}{B4 / 64.0} & Segformer & 82.6 & 27.4 & 73.4 & 45.2 & 78.2 & 56.5 & 45.8 & 73.1 & 64.9 & 53.8 & 57.6 & 69.7 \\

&  CWFA (Ours) &   82.1 & \textbf{58.9} & \textbf{74.2} & \textbf{47.4} & 78.1 & \textbf{69.0} & \textbf{50.5} & \textbf{73.7} & \textbf{65.7} & \textbf{57.8} & \textbf{63.9} & \textbf{77.8}\\
\cline{1-14}
  \multirow{2}{*}{B5 / 84.6}  & Segformer & 82.4 & 31.4 & 74.4 & 46.9 & 78.3 & 58.5 & 46.4 & 73.2 & 65.3 & 54.5 & 58.8 & 71.3 \\

 &  CWFA (Ours)  & 82.2 & \textbf{56.9} & 73.7 & \textbf{51.8} & 77.7 & \textbf{69.0} & \textbf{49.4} & \textbf{73.3} & \textbf{65.9} & \textbf{57.6} & \textbf{63.9} & \textbf{77.8}  \\

\end{tabular}}
\end{table*}

\begin{table*}[!t]
\caption{\textbf{Segformer on Cityscapes.} Our results compared to the Segformer baseline when fine-tuning the baseline with and w/o CWFA. Our results show that in only 16k iterations CFWA improves the robustness of the model substantially}
\label{table: segformer on cityscapes-c 16k-finetuning}
\scalebox{0.6}{
\begin{tabular}{c|c|c|c|c|c|c|c|c|c|c|c|c|c|c|c|c|c|c|c|c|} 
   \multicolumn{1}{c|}{Model /}& \multirow{2}{*}{Method} & City &\multicolumn{4}{|c|}{Blur}  & \multicolumn{4}{|c|}{Noise} & \multicolumn{4}{|c|}{Digital} & \multicolumn{4}{|c|}{Weather} & \multirow{2}{*}{mCE} & \multirow{2}{*}{Reten.}  \\
  \cline{4-19}
    \multicolumn{1}{c|}{Params. [M.]} & &(Clean) & Motion & Defoc & Glass & Gauss & Gauss & Impul & Shot & Speck & Bright & Contr & Satur & JPEG & Snow & Spatt & Fog & Frost & & \\
     \hline
    \multirow{2}{*}{B0 / 3.7}  & Segformer &  76.4 & 60.4 & 60.1 & 51.6 & 60.2 & 29.5 & 31.2 & 35.1 & 53.1 & 73.5 & 67.1 & 71.8 & 39.2 & 23.0 & 54.2 & 65.4 & 32.4 & 50.5 & 66.1\\

 &  CWFA (Ours) &  75.6 & \textbf{61.6} & 60.0 & \textbf{57.2} & \textbf{60.7} & \textbf{44.2} & \textbf{46.2} & \textbf{49.8} & \textbf{61.6} & 73.2 & 66.8 & 70.9 & \textbf{44.2} & \textbf{38.4} & \textbf{60.9} & 64.9 & \textbf{41.1} & \textbf{56.3} & \textbf{74.5}\\
 \hline
    \multirow{2}{*}{B1 / 13.6}  & Segformer & 77.8 & 63.6 & 63.0 & 52.0 & 62.9 & 34.6 & 25.9 & 39.6 & 58.4 & 75.4 & 71.5 & 74.4 & 37.3 & 28.0 & 59.3 & 71.4 & 38.5 & 53.5 & 68.7   \\

&  CWFA (Ours) & 77.2 & \textbf{64.5} & \textbf{63.3} & \textbf{59.8} & \textbf{63.8} & \textbf{49.0} & \textbf{46.4} & \textbf{54.4} & \textbf{63.3} & 74.2 & 68.6 & 73.3 & \textbf{46.0} & \textbf{45.5} & \textbf{65.3} & 67.0 & \textbf{43.0} & \textbf{59.2} & \textbf{76.7} \\
  \hline
    \multirow{2}{*}{B2 / 27.3}  & Segformer & 81.0 & 68.2 & 67.6 & 59.6 & 67.9 & 35.3 & 34.2 & 40.9 & 58.9 & 79.9 & 75.9 & 78.5 & 44.3 & 36.6 & 66.1 & 75.7 & 41.7 & 58.2 & 71.9 \\

&  CWFA (Ours)  & 80.8 & \textbf{69.4} & \textbf{68.5} & \textbf{63.8} & \textbf{69.0} & \textbf{55.6} & \textbf{55.0} & \textbf{61.6} & \textbf{71.1} & 79.8 & 74.6 & 78.3 & \textbf{50.1} & \textbf{52.0} & \textbf{70.0} & 74.4 & \textbf{50.4} & \textbf{65.2} & \textbf{80.7}\\
\hline
  \multirow{2}{*}{B3 / 47.2}  & Segformer  &  81.4 & 69.5 & 68.9 & 61.4 & 70.0 & 51.4 & 47.9 & 56.9 & 68.4 & 80.5 & 76.8 & 79.6 & 52.2 & 35.0 & 66.5 & 77.8 & 45.5 & 63.0 & 77.4 \\ 

&  CWFA (Ours) &  \textbf{81.6} & \textbf{71.2} & \textbf{70.5} & \textbf{64.8} & \textbf{71.7} & \textbf{61.7} & \textbf{61.3} & \textbf{66.8} & \textbf{74.3} & 80.4 & 76.8 & 79.3 & \textbf{53.9} & \textbf{52.0} & \textbf{70.2} & 76.4 & \textbf{54.0} & \textbf{67.8} & \textbf{83.1}\\
\hline
  \multirow{2}{*}{B4 / 64.0}  & Segformer & 82.3 & 69.6 & 69.2 & 61.7 & 70.1 & 58.4 & 59.7 & 62.6 & 71.4 & 81.1 & 77.6 & 80.2 & 55.2 & 39.3 & 68.1 & 78.2 & 45.1 & 65.5 & 79.6 \\

&  CWFA (Ours) &  81.9 & \textbf{70.8} & \textbf{70.2} & \textbf{64.8} & \textbf{71.2} & \textbf{61.8} & \textbf{60.8} & \textbf{67.4} & \textbf{74.3} & 80.8 & 77.4 & 79.9 & \textbf{56.2} & \textbf{50.2} & \textbf{70.3} & 77.4 & \textbf{52.8} & \textbf{67.9} & \textbf{82.9} \\
\hline
  \multirow{2}{*}{B5 / 84.6}  & Segformer &  82.3 & 69.6 & 68.9 & 64.4 & 69.7 & 59.3 & 55.1 & 64.5 & 73.4 & 81.0 & 77.9 & 80.2 & 57.2 & 43.5 & 69.2 & 78.5 & 47.9 & 66.3 & 80.5  \\

 &  CWFA (Ours)  & 82.2 & \textbf{72.0} & \textbf{71.6} & \textbf{66.5} & \textbf{72.5} & \textbf{65.8} & \textbf{64.4} & \textbf{70.4} & \textbf{75.8} & 80.0 & 77.8 & 80.0 & \textbf{59.0} & \textbf{54.7} & \textbf{70.9} & 78.0 & \textbf{54.8} & \textbf{69.7} & \textbf{84.8} \\
\end{tabular}}
\end{table*}
\begin{table*}[!t]
\caption{\textbf{Segformer on ADE20K-C.} Our results compared to the Segformer baseline. Our approach yields impressive improvements for most corruptions independently of the model size.}
\label{table: segformer on ade20k-c full table}
\scalebox{0.6}{
\begin{tabular}{c|c|c|c|c|c|c|c|c|c|c|c|c|c|c|c|c|c|c|c|c|} 
   \multicolumn{1}{c|}{Model /}& \multirow{2}{*}{Method} & ADE20K &\multicolumn{4}{|c|}{Blur}  & \multicolumn{4}{|c|}{Noise} & \multicolumn{4}{|c|}{Digital} & \multicolumn{4}{|c|}{Weather} & \multirow{2}{*}{mCE} & \multirow{2}{*}{Reten.}  \\
  \cline{4-19}
    \multicolumn{1}{c|}{Params. [M.]} & &(Clean) & Motion & Defoc & Glass & Gauss & Gauss & Impul & Shot & Speck & Bright & Contr & Satur & JPEG & Snow & Spatt & Fog & Frost & & \\
     \hline
    \multirow{2}{*}{B0 / 3.7}  & Segformer & 37.2 & 21.5 & 22.4 & 18.6 & 23.3 & 18.3 & 15.9 & 18.0 & 22.9 & 33.8 & 27.1 & 32.5 & 29.8 & 13.4 & 21.7 & 28.2 & 15.5 & 22.7 & 61.1 \\

 &  CWFA (Ours) & 36.7 & \textbf{23.8} & \textbf{23.1} & \textbf{20.8} & \textbf{24.0} & \textbf{23.1} & \textbf{21.3} & \textbf{23.1} & \textbf{26.5} & 33.8 & \textbf{29.0} & 32.4 & \textbf{30.4} & \textbf{18.9} & \textbf{25.5} & \textbf{29.9} & \textbf{19.8} & \textbf{25.3} & \textbf{69.0}\\
 \hline
    \multirow{2}{*}{B1 / 13.6}  & Segformer & 42.0 & 26.5 & 26.5 & 21.2 & 27.2 & 22.3 & 19.2 & 21.6 & 25.9 & 38.4 & 32.1 & 37.1 & 32.9 & 18.0 & 26.3 & 35.4 & 20.1 & 26.9 & 64.2  \\
&  CWFA (Ours) & 40.3 & \textbf{28.7} & \textbf{26.8} & \textbf{23.5} & \textbf{27.7} & \textbf{28.6} & \textbf{27.2} & \textbf{27.9} & \textbf{30.1} & 37.2 & \textbf{32.9} & 36.0 & \textbf{33.7} & \textbf{23.1} & \textbf{29.5} & 34.3 & \textbf{23.5} & \textbf{29.4} & \textbf{73.0}  \\
  \hline
    \multirow{2}{*}{B2 / 27.3}  & Segformer & 46.0 & 31.0 & 32.3 & 24.1 & 33.0 & 32.4 & 29.6 & 31.8 & 35.2 & 43.0 & 40.1 & 42.3 & 38.6 & 25.9 & 33.0 & 41.3 & 27.0 & 33.8 & 73.4  \\
&  CWFA (Ours)  & 45.5 & \textbf{34.0} & 32.0 & \textbf{27.6} & 32.8 & \textbf{36.9} & \textbf{35.7} & \textbf{36.5} & \textbf{38.1} & 42.7 & \textbf{41.4} & 42.0 & \textbf{39.5} & \textbf{30.7} & \textbf{36.6} & 40.6 & \textbf{29.3} & \textbf{36.0} & \textbf{79.2}\\
\hline
  \multirow{2}{*}{B3 / 47.2}  & Segformer  & 48.1 & 35.0 & 33.9 & 27.8 & 34.7 & 37.1 & 37.0 & 37.0 & 39.3 & 45.2 & 41.7 & 44.8 & 42.3 & 29.4 & 38.3 & 43.8 & 30.5 & 37.4 & 77.6 \\ 
&  CWFA (Ours) & 47.3 & \textbf{36.4} & \textbf{34.1} & \textbf{29.3} & \textbf{34.9} & \textbf{39.5} & \textbf{39.5} & \textbf{39.4} & \textbf{40.3} & 44.4 & \textbf{42.9} & 43.9 & 41.5 & \textbf{32.6} & \textbf{39.1} & 42.5 & \textbf{31.5} & \textbf{38.2} & \textbf{80.9} \\
\hline
  \multirow{2}{*}{B4 / 64.0}  & Segformer & 50.1 & 35.2 & 35.9 & 27.7 & 36.8 & 38.7 & 36.3 & 38.4 & 40.1 & 47.2 & 44.0 & 46.2 & 43.6 & 30.0 & 36.7 & 45.7 & 30.1 & 38.3 & 76.4\\
&  CWFA (Ours) &  48.8 & \textbf{38.0} & 35.5 & \textbf{31.1} & 36.5 & \textbf{42.0} & \textbf{40.1} & \textbf{41.5} & \textbf{42.2} & 46.4 & \textbf{45.0} & 45.4 & 43.4 & \textbf{34.1} & \textbf{39.5} & 44.7 & \textbf{32.3} & \textbf{39.8} & \textbf{81.7} \\
\hline
  \multirow{2}{*}{B5 / 84.6}  & Segformer & 51.0 & 35.1 & 35.3 & 28.1 & 36.7 & 34.5 & 31.7 & 39.1 & 41.3 & 47.7 & 45.0 & 47.0 & 43.6 & 31.8 & 39.7 & 46.6 & 31.2 & 38.4 & 75.3  \\
 &  CWFA (Ours)  & 49.8 & \textbf{38.2} & \textbf{36.1} & \textbf{31.6} & \textbf{37.3} & \textbf{42.0} & \textbf{41.2} & \textbf{41.6} & \textbf{42.6} & 47.1 & \textbf{46.5} & 46.5 & \textbf{43.7} & \textbf{34.2} & \textbf{40.9} & 45.7 & \textbf{32.3} & \textbf{40.5} & \textbf{81.3} \\
\end{tabular}}
\end{table*}
\begin{table*}[!t]
\caption{\textbf{Transferability of the corruptions}}
\label{fig:transfFigure}
\scalebox{0.5}{
\begin{tabular}{|c|c|c|c|c|c|c|c|c|c|c|c|c|c|c|c|c|c|c|c|c|c|c|c|} 
  \hline
  Model & Clean &\multicolumn{4}{|c|}{Blur}  & \multicolumn{4}{|c|}{Noise} & \multicolumn{4}{|c|}{Digital} & \multicolumn{4}{|c|}{Weather} & Average  & Reten. & Avg & Ret  & CWFA model & CWFA model Reten. \\
  \cline{2-5}
  \cline{5-9}
  \cline{9-13}
  \cline{13-19}
   & & Motion & Defoc & Glass & Gauss & Gauss & Impul & Shot & Speck & Bright & Contr & Satur & JPEG & Snow & Spatt & Fog & Frost & & & w/o corr & w/o corr & w/o corr & w/o corr  \\
  \hline
B0 (Baseline)  & 76.3 & 59.8 & 59.1 & 51.0 & 59.2 & 25.3 & 27.0 & 30.7 & 51.5 & 73.4 & 66.5 & 72.0 & 38.3 & 22.4 & 53.5 & 65.6 & 31.6 & 49.2 & 64.5 & 49,2 & 64.5 & - & - \\
 \hline
CWFA & 76.2 & 62.2 & 60.9 & 57.1 & 61.7 & 41.7 & 42.0 & 47.3 & 60.4 & 73.1 & 68.4 & 71.4 & 43.7 & 37.0 & 60.7 & 65.8 & 39.4 & 55.8 &  73.2 &  55.8 & 73.2 & - & - \\
\hline
Motion Blur & 75.8 & 68.5& 65.6& 58.8& 65.7& 27.8& 28.2& 33.6& 52.5& 72.6& 68.6& 68.6& 39.9& 26.2& 55.2& 65.4& 34.2& 52.0& 68.6 & 47.7 & 63.0 & 54.3 & 71.2\\
\hline

Defocus Blur  & 76.1& 63.5& 69.6& 54.5& 69.4& 28.8& 28.4& 33.7& 53.6& 71.8& 68.3& 67.8& 37.9& 22.4& 54.3& 65.2& 32.7& 51.4& 67.5 & 47.1 & 61.9 & 54.2 & 71.2 \\
\hline

Gaussian Blur  & 76.3& 64.2& 69.6& 55.8& 70.5& 31.8& 32.1& 37.8& 56.5& 72.5& 67.8& 68.4& 38.5& 22.5& 55.2& 65.3& 33.8& 52.6& 69.0 &  48.5 & 63.6 & 54.2 & 71.2\\
\hline

Gaussian Noise  & 75.8& 60.9& 61.1& 59.4& 61.3& 66.2& 64.6& 68.9& 72.2& 72.6& 65.7& 69.8& 47.8& 32.2& 60.4 & 59.7& 32.3& w/o 59.7& 78.7 & 56.9 & 75.1 & 58.4 &  76.7\\
\hline

 Impulse Noise   & 75.6& 60.8& 60.4& 58.3& 60.7& 64.1& 67.6& 67.6& 71.6& 71.7& 64.9& 69.1& 46.9& 33.9& 61.2& 59.7& 33.9& 59.5& 78.7 & 56.8 & 75.1 & 58.4 & 76.7 \\
\hline

Shot Noise   & 75.5& 61.0& 60.5& 59.0& 60.8& 65.3& 63.7& 68.7& 72.4& 71.7& 65.8& 69.2& 46.7& 29.1& 58.9& 60.7& 33.4& 59.2& 78.4 & 56.4 & 74.7 & 58.4 & 76.7\\
\hline

Speckle Noise  & 75.9& 61.2& 60.9& 58.3& 61.5& 63.1& 60.0& 67.7& 72.2& 72.0& 64.9& 69.1& 44.8& 26.4& 58.6& 61.9& 31.7& 58.4& 76.9 & 55.9 & 73.7 & 58.4 & 76.7\\
\hline

Brightness   & 76.0& 60.0& 59.2& 52.1& 59.5& 25.3& 28.0& 29.3& 48.5& 75.3& 67.0& 70.3& 38.1& 23.7& 55.8& 63.1& 34.9& 49.4& 65.0 & 47.7 & 62.7 & 54.6 & 71.7 \\
\hline

Contrast  & 76.2& 61.2& 60.1& 52.4& 60.1& 32.1& 32.0& 37.0& 53.8& 73.1& 74.7& 70.6& 39.9& 22.6& 54.4& 65.7& 34.3& 51.5& 67.6 & 50.0 & 65.6 & 55.0 & 72.1\\
\hline

Saturate  & 76.6& 60.7& 59.2& 52.1& 59.7& 25.2& 27.8& 30.9& 51.7& 73.6& 68.1& 75.0& 40.5& 25.3& 56.0& 65.7& 35.0& 50.4& 65.9 & 48.8 & 63.7 & 54.8 & 71.9\\
\hline

JPEG compression  &  75.3 & 60.6& 60.1& 57.2& 60.3& 34.9& 35.8& 39.2& 55.0& 71.7& 62.4& 65.3& 66.5& 23.4& 54.7& 58.2& 31.0& 52.3 & 69.4  & 51.3 & 68.1 & 56.6 & 74.3\\
\hline

Snow  & 75.7& 61.5& 59.3& 61.2& 59.6& 41.9& 44.5& 48.0& 62.2& 72.1& 65.0& 68.9& 44.4& 67.1& 67.4& 63.4& 40.8& 58.0& 76.5  & 57.3 & 75.7 & 57.1 & 74.3\\
\hline

Spatter  &  75.6& 60.2& 59.7& 58.1& 59.6& 28.5& 35.5& 35.0& 55.7& 72.3& 65.6& 68.9& 42.6& 40.4& 72.6& 61.2& 33.9& 53.1& 70.2 & 51.8 & 68.5 & 55.5 & 72.8\\
\hline

Fog  & 75.8& 61.0& 59.9& 53.8& 60.2& 22.9& 22.5& 27.9& 50.1& 72.8& 67.7& 69.8& 36.7& 23.0& 55.7& 73.6& 36.0& 49.6& 65.5 & 48.0 & 63.4 &  55.1 & 72.4\\
\hline

Frost  & 75.4& 60.3& 59.9& 53.3& 59.3& 29.5& 30.4& 33.6& 51.8& 73.6& 68.7& 69.0& 36.1& 37.4& 61.9& 69.5& 66.5& 53.8& 71.3 &  52.9 & 70.2 &  56.9 & 74.7\\
\hline

\end{tabular}}
\end{table*}


\end{document}